\documentclass[10pt,journal,compsoc]{IEEEtran}

\ifCLASSOPTIONcompsoc
  \usepackage[nocompress]{cite}
\else
  \usepackage{cite}
\fi

\usepackage{setspace}
\usepackage{booktabs}
\usepackage{float}
\usepackage{graphicx}
\usepackage{epsfig}
\usepackage{amsmath}
\usepackage{subfig}
\usepackage{bm}
\usepackage{verbatim}
\usepackage{amsfonts}
\usepackage{amssymb}
\usepackage{makecell}
\usepackage{multirow}
\usepackage{booktabs}
\usepackage{ragged2e}
\usepackage{enumerate}
\usepackage[normalem]{ulem}

\ifCLASSINFOpdf
\else
\fi

\usepackage{xcolor}

\newcommand{\zxr}[1]{{\color{red} #1}}

\newcommand{\eat}[1]{{}}

\begin{document}
%\title{Progresses, Opportunities and Challenges: A Survey on Multimodal Knowledge Graph Construction and Application}
\title{Multi-Modal Knowledge Graph Construction and Application: A Survey}

\author{Xiangru~Zhu,
        Zhixu~Li~\IEEEmembership{Member,~IEEE,}
        Xiaodan~Wang,
        Xueyao~Jiang,
        Penglei~Sun,
        Xuwu~Wang,
        Yanghua~Xiao~\IEEEmembership{Member,~IEEE,}
        Nicholas~Jing~Yuan~\IEEEmembership{Member,~IEEE,}
\IEEEcompsocitemizethanks{
\IEEEcompsocthanksitem X. Zhu, Z. Li, X. Wang, X. Jiang, P. Sun, X. Wang and Y. Xiao are with the School of Computer Science, Fudan University.\protect\\
E-mail: \{xrzhu19, zhixuli, xiaodanwang20, xueyaojiang19, plsun20, xwwang18, shawyh\}@fudan.edu.cn. Z. Li and Y. Xiao are the corresponding authors.
\IEEEcompsocthanksitem N.J. Yuan is with Huawei Cloud \& AI, Hangzhou, Zhejiang, China. \protect\\
E-mail: nicholas.yuan@huawei.com}
}

% if arXiv, comment it
% if TKDE, uncomment it 
\markboth{Journal of \LaTeX\ Class Files,~Vol.~14, No.~8, August~2015}%
{Shell \MakeLowercase{\textit{et al.}}: Bare Demo of IEEEtran.cls for Computer Society Journals}

\IEEEtitleabstractindextext{%
\begin{abstract}
Recent years have witnessed the resurgence of knowledge engineering which is featured by the fast growth of knowledge graphs. However, most of existing knowledge graphs are represented with pure symbols, which hurts the machine's capability to understand the real world. The multi-modalization of knowledge graphs is an inevitable key step towards the realization of human-level machine intelligence. The results of this endeavor are Multi-modal Knowledge Graphs (MMKGs).
  In this survey on MMKGs constructed by texts and images, we first give definitions of MMKGs, followed with the preliminaries on multi-modal tasks and techniques. We then systematically review the challenges, progresses and opportunities on the construction and application of MMKGs respectively, with detailed analyses of the strengths and weaknesses of different solutions. 
We finalize this survey with open research problems relevant to MMKGs.
\end{abstract}

}

\maketitle

% not be used in maketitle, but will appear (i.e., to be ``transported'')
\IEEEdisplaynontitleabstractindextext

% \fi
\IEEEpeerreviewmaketitle

\IEEEraisesectionheading{\section{Introduction}\label{sec:introduction}}
% with the very first section heading (almost always called ``Introduction'').

% Here we have the typical use of a ``T'' for an initial drop letter
% and ``HIS'' in caps to complete the first word.
\IEEEPARstart{R}{e}cent years have witnessed the resurgence of knowledge engineering featured by the fast growth of knowledge graphs. A knowledge graph (KG) is essentially a large-scale semantic network that contains entities, concepts as nodes and various semantic relationships among them as edges. %Knowledge graphs have found their great value in a wide range of real-world applications
The great value of knowledge graphs has been found in a wide range of real-world applications, including text understanding, recommendation systems and natural language question answering. More and more knowledge graphs have been created, covering common sense knowledge (e.g., Cyc~\cite{matuszek2006introduction}, ConceptNet~\cite{liu2004conceptnet}), lexical knowledge (e.g., WordNet~\cite{miller1995wordnet}, BabelNet~\cite{navigli2010babelnet}), encyclopedia knowledge (e.g., Freebase~\cite{bollacker2008freebase}, DBpedia~\cite{auer2007dbpedia}, YAGO~\cite{suchanek2007yago}, WikiData~\cite{vrandevcic2014wikidata}, CN-DBpedia~\cite{xu2017cn}), taxonomic knowledge (e.g., Probase~\cite{wu2012probase}) and geographic knowledge (e.g., GeoNames~\cite{wick2012geonames}).

However, most of the existing knowledge graphs are represented with pure symbols denoted in the form of text, which weakens the capability of machines to describe and understand the real world. 
A human being cannot understand what a \texttt{dog} is without the experience of living with a dog, which enlightens researchers to establish the connection between the symbol \texttt{Dog} and the experience of dogs, that is, grounding a symbol to its physical world meaning~\cite{harnad1990symbol,harnad2003symbol,steels2008symbol}.
Similarly, grounding symbolic forms to non-symbolic experiences benefits receiving real communicative intents~\cite{bender2020climbing}. For example, the customers cannot understand the meaning of \texttt{Hand-in-waistcoat} as a particular pose (hand inside coat flap) without the \emph{experience} of \texttt{Hand-in-waistcoat} so that the customer would respond incorrectly to the request of photographers.
Thus, it is necessary to ground symbols to corresponding images, sound and video data and map symbols to their corresponding referents with meanings in the physical world, enabling machines to generate similar ``\emph{experiences}'' like a real human~\cite{harnad1990symbol} when they are confronted with a specific entity \texttt{Hand-in-waistcoat} or an abstract concept \texttt{Dog}.
On the other hand, there is an increasing demand for the multi-modality of knowledge to break through the bottleneck of real-world applications~\cite{shi2019knowledge, niu2021counterfactual, zhao2021boosting}.  
For instance, in relation extraction tasks, an additional image usually greatly improves the performance in the extraction of the attributes and relationships that are visually obvious but difficult to be recognized in symbols and text, such as \emph{partOf} (e.g., \textit{The keyboard and the screen are parts of a laptop.}) and \emph{colorOf} (e.g., \emph{A banana is usually yellow or yellowish-green but not blue }).
In text generation tasks, if the machine has been empowered with the ability to recognize a specific entity in an image by the reference to a Multi-Modal KG (MMKG), the machine is possible to generate a more informative entity-level sentence (e.g., \emph{Donald Trump is making a speech}) instead of a vague concept-level description (e.g., \emph{A tall man with blond hair is making a speech}). 

\begin{figure*}[t]
\centering
\subfloat[MMKG with multi-modal data as attribute values]
{
\begin{minipage}[t]{0.46\textwidth}
\centering
\scalebox{1}{\includegraphics[width=1\linewidth]{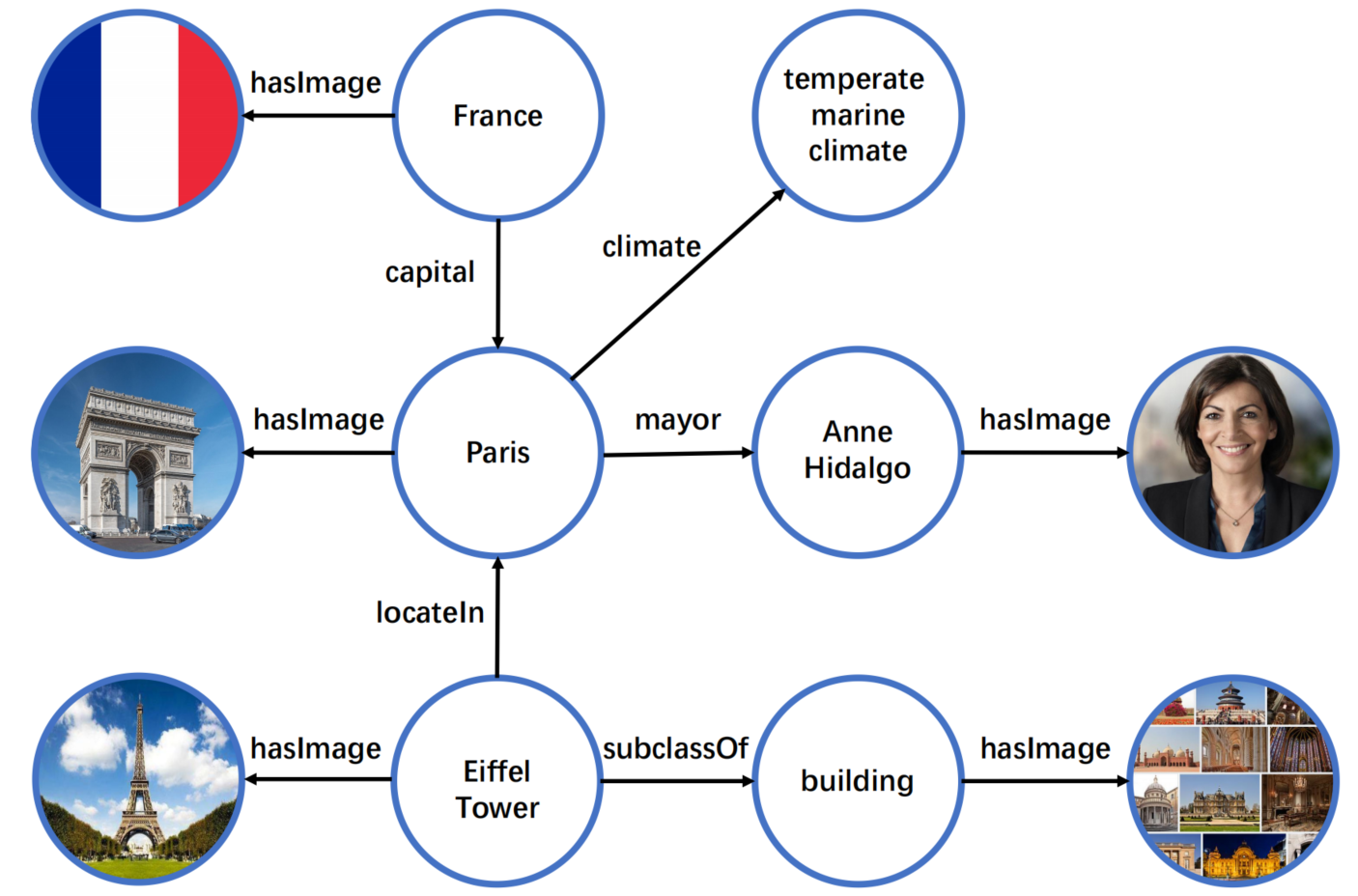}}
\end{minipage}
}
%\hspace{3mm}
\subfloat[MMKG with multi-modal data as entities]
{
\begin{minipage}[t]{0.46\textwidth}
\centering
\scalebox{1}{\includegraphics[width=1\linewidth]{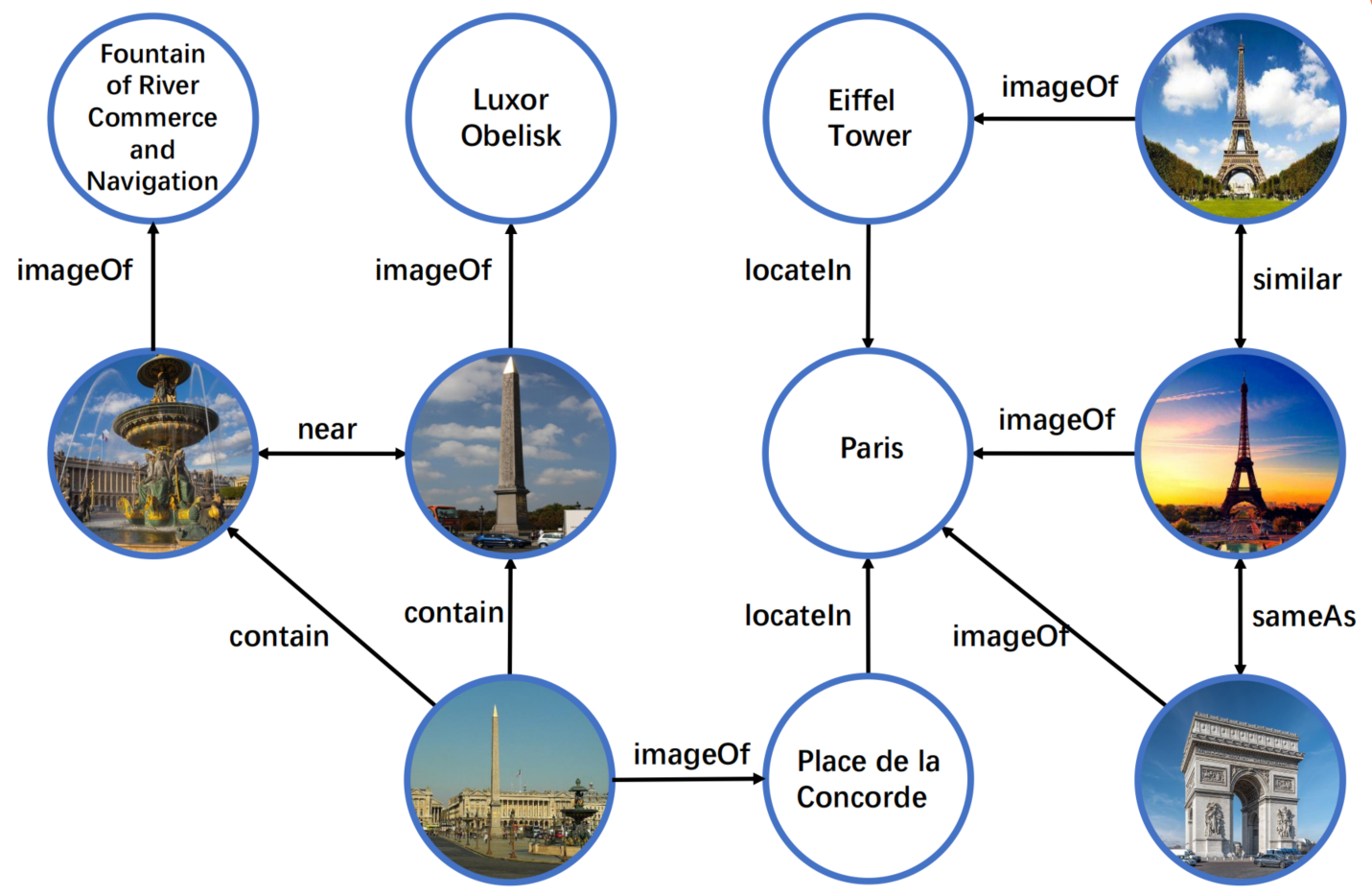}}
\end{minipage}
}
\caption{Example MMKGs of two different types: A-MMKG and N-MMKG}%: (a) A-MMKG: MMKG with multi-modal data as attribute values; (b) N-MMKG: MMKG with multi-modal data as entities.}
\label{fig:definitions_2mmkg}
\vspace{-1.0em}
\end{figure*}

Due to the rapid growth of applications' demand for multi-modal knowledge guidance, the multi-modalization of KGs and their applications 
%\cite{zhifeili2022learning, zhaolizhang2022multiscale, zhifeili2021recalibration}
has been booming in recent years\cite{sun2020multi,xu2021alime,li2020gaia}. 
Nevertheless, a systematic review of the recent research progresses, challenges and opportunities in this emerging area are still lacking. In this paper, we hope to fill the gap and systematically survey the recent research progresses relevant to MMKG as follows:
1) {\bf Construction.} The construction of MMKGs could be conducted in two opposite directions. One is from images to symbols, i.e., labeling images with symbols in KG; the other is from symbols to images, i.e., grounding symbols in KG to images. In the Construction section, we will systematically cover the challenges, progresses as well as opportunities to correlate various symbol knowledge (e.g., entities, concepts, relations and events) to their corresponding images in the two opposite directions.
2) {\bf Application.} The application of MMKGs could be roughly divided into two categories: In-MMKG applications aiming at addressing the quality or integration issues of MMKGs themselves, and Out-of-MMKG applications which are general multi-modal tasks that MMKGs can help.
The Application section will present how MMKGs are applied in several well-studied multi-modal tasks.

To summarize, we are the first to thoroughly survey the existing work on MMKGs consisting of texts and images. To enhance the value of this survey, we pay attention to the following features:
1) {\bf Comprehensive Survey.} We systematically and comprehensively review the existing work on MMKG construction and application.
2) {\bf Insightful Analysis.} We analyze the strengths and weaknesses of different solutions in MMKG construction and discuss how MMKGs can help in various downstream applications.
3) {\bf Revealed Opportunities.} We not only point out some potential opportunities with the studied tasks relevant to MMKG construction, but also list some promising future directions with MMKG.

The rest of the survey is organized as follows: 
Sec.~\ref{sec:background} gives definitions and preliminaries on MMKGs.
Sec.~\ref{sec:construction} conducts a comprehensive review of the challenges, progresses and opportunities of the construction of MMKGs, while Sec.~\ref{sec:application} presents how MMKGs are applied in several well-studied multi-modal applications.
Sec.~\ref{sec:opportunities} reviews some open problems of MMKG and highlights promising future directions. 
Sec.~\ref{sec:conclusion} finally concludes the paper.

\section{Definitions and Preliminaries}
\label{sec:background}

This section first defines two representation ways for KGs and then reviews some preliminaries on multi-modal techniques, followed by a discussion on the connections between MMKGs and the existing multi-modal techniques.

\eat{
This section first defines two representation ways for KGs and then reviews some preliminaries on multi-modal tasks and techniques, followed by a discussion on the connections between MMKGs and the existing multi-modal tasks and techniques.
}

%multi-modalization ways for KGs
\subsection{Definition \& Representation of MMKGs}
%Symbolic Representation and Design}
\label{sec:defMMKG}

A traditional Knowledge Graph (KG) is defined as a directed graph $\mathcal{G=\{E,R,A,V,T_R,T_A\}}$, where $\mathcal{E}$, $\mathcal{R}$, $\mathcal{A}$, $\mathcal{V}$ are sets of entities, relations, attributes and literal attribute values, and $\mathcal{T_R}=\mathcal{E}\times\mathcal{R}\times\mathcal{E}$ and $\mathcal{T_A}=\mathcal{E}\times\mathcal{A}\times\mathcal{V}$ are sets of relation triples and attribute triples respectively.
A triple $\mathcal(s,p,o)\in\mathcal{T_R}$ denotes that \emph{entity} $s\in\mathcal{E}$ has a \emph{relation} $p\in\mathcal{R}$ with \emph{entity} $o\in\mathcal{E}$.
A triple $\mathcal(s,p,o)\in\mathcal{T_A}$ denotes that \emph{entity} $s\in\mathcal{E}$ has an \emph{attribute} $p\in\mathcal{A}$ with the \emph{attribute value} $o\in\mathcal{V}$.

A Multi-modal Knowledge Graph (MMKG) can be seen as a multi-modalized KG, which has part of its knowledge in $\mathcal{\{E,R,A,V,T_R,T_A\}}$ multi-modalized.
We say a particular knowledge symbol   is multi-modalized if it is associated with its corresponding data items in modalities other than text, such as image, sound or video, that could embody the knowledge.
For instance, a relation triple (\emph{s}, \emph{p}, \emph{o}) can be multi-modalized with an image describing the \emph{relation} $p$ between $s$ and $o$.
%
%, or the \emph{attribute} $p$ of \emph{entity} $s$.
%
%Note that it is unnecessary to require all the elements be multi-modalized in a MMKG, since some of them can hardly find corresponding multi-modal data items.

\begin{table*}[t]
    \centering
    \scriptsize
    \subfloat[Example RDF triples in A-MMKG]{
        \begin{minipage}[t]{0.47\textwidth}
        \begin{tabular}{ccc}
            \toprule
            subject & predicate & object \\
            \midrule
            France & hasImage & The\_flag\_of\_France.jpg\\
            %France & capital & Paris\\
            Anne Hidalgo & hasImage & Anne\_Hidalgo.jpg\\
            Paris & mayor & Anne Hidalgo\\
            %Paris & climate & temperate marine climate\\
            Paris & hasImage & A\_landmark\_of\_Paris.jpg\\
            Eiffel Tower & locateIn & Paris\\
            Eiffel Tower & hasImage & Eiffel\_Tower.jpg\\
            Eiffel Tower & subclassOf & building\\
            building & hasImage & a\_kind\_of\_architectural\_style.jpg\\
            \bottomrule
        \end{tabular}
        \end{minipage}
        \label{tab:rdf-mmkg-def-1}
    }
    \subfloat[Example RDF triples in N-MMKG]{
        \begin{minipage}[t]{0.47\textwidth}
        \begin{tabular}{ccc}
            \toprule
            subject & predicate & object \\
            \midrule
            %Place\_de\_la\_Concorde.jpg & imageOf & Place de la Concorde\\
            %Place\_de\_la\_Concorde.jpg & contain & Luxor\_Obelisk.jpg\\
            %Arc\_de\_Triomphe\_in\_Paris.jpg & imageOf & Paris\\
            Eiffel\_Tower\_in\_Paris.jpg & imageOf & Paris\\
            Eiffel\_Tower\_in\_Paris.jpg & size & 700*1600\\
            Eiffel\_Tower\_in\_Paris.jpg & sameAs & Arc\_de\_Triomphe\_in\_Paris.jpg\\
            Eiffel\_Tower\_in\_Paris.jpg & similar & Eiffel\_Tower.jpg \\
            Eiffel Tower & locateIn & Paris\\
            Eiffel\_Tower.jpg & imageOf & Eiffel Tower\\
            Eiffel\_Tower.jpg.HOG & describes & Eiffel\_Tower.jpg \\
            Eiffel\_Tower.jpg.HOG & value & [0.0775 , 0.0120 , 0.0021 , ...] \\
            \bottomrule
        \end{tabular}
        \end{minipage}
        \label{tab:rdf-mmkg-def-2}
    }
    \caption{Example RDF triples in different types of MMKGs, where items end up with ``.jpg'' are images.} %In A-MMKGs, images are only objects. In N-MMKGs, images could be subjects or objects.}
    \label{tab:rdf-mmkg-def}
    \vspace{-2.0em}
\end{table*}
Existing work on MMKGs mainly adopts two different ways for representing MMKGs.
One way takes multi-modal data (images in this survey) as particular attribute values of entities or concepts, as the example shown in Fig.~\ref{fig:definitions_2mmkg}(a).
We name an MMKG represented in this way as {\bf A-MMKG} for short, denoted as $\mathcal{G=\{E,R,A,V,T_R,T_A\}}$, where $\mathcal{T_A=\mathcal{E}\times\mathcal{A}\times(\mathcal{V_{KG}}\cup\mathcal{V_{MM}})}$ is the set of attribute triples, $\mathcal{V_{KG}}$ is the set of the KG's attribute values and $\mathcal{V_{MM}}$ is the set of multi-modal data.
In A-MMKGs, since multi-modal data are treated as attribute values, in a triple ($s$,$p$,$o$), % describing the multi-modal information of an entity, 
$s$ denotes an entity, $o$ denotes one of its corresponding multi-modal data, and the relation $p$ is ``hasImage'' when $o$ is an image. Some example triples are listed in Table~\ref{tab:rdf-mmkg-def-1}.

The other way takes multi-modal data as entities in KGs, as the example shown in Figure~\ref{fig:definitions_2mmkg}(b). We name an MMKG represented in this way as {\bf N-MMKG} for short, denoted as $\mathcal{G=\{E,R,A,V,T_R,T_A\}}$, where $\mathcal{T_R=(\mathcal{E_{KG}}\cup\mathcal{E_{MM}})\times\mathcal{R}\times(\mathcal{E_{KG}\cup\mathcal{E_{MM}}})}$ is the set of relation triples, $\mathcal{E_{KG}}$ is the set of KG entities and $\mathcal{E_{MM}}$ is the set of multi-modal data.
Since multi-modal data are treated as new entities, more inter-modal and intra-modal relations are discovered and added into the MMKG. For example, in Table \ref{tab:rdf-mmkg-def-2}, the entity \texttt{Eiffel Tower} is associated with an image \emph{Eiffel\_Tower.jpg} by the relation \texttt{imageOf}. Two images can also be associated in one of the following relations: 1) \texttt{contain}: One image entity visually contains another image entity by the relative position of images. 2) \texttt{nearBy}: One image entity is visually nearby another image entity in an image. 3) \texttt{sameAs}: Two different image entities refer to the same entity. 4) \texttt{similar}: Two image entities are visually similar to each other.

In addition, in N-MMKGs an image is usually abstracted into several image descriptors, which are usually summarized into feature vectors of the image entity at the pixel level, such as Gray Histogram Descriptor (GHD), Histogram of Oriented Gradients Descriptor (HOG), Color Layout Descriptor (CLD) and so on. For example, in Table~\ref{tab:rdf-mmkg-def-2}, \emph{Eiffel\_Tower\_in\_Paris.jpg.HOG} is one of the descriptors of the image \emph{Eiffel\_Tower\_in\_Paris.jpg}, and is in the form of a vector. These image descriptors are well interpreted. Thus the relations between images can be obtained by simple calculations (e.g., image similarity obtained via the inner product of vectors of image descriptors).

%
%For example in Figure~\ref{fig:definitions_2mmkg}(a), the entity \texttt{Eiffel Tower} can be grounded to its photo, while the concept \texttt{building} can be grounded with a collection of images of various architectural styles.
%

%
%, the fact that ``the  \texttt{Luxor Obelisk} is located at \texttt{Place de la Concorde}'' can be represented by an image.
%\leeon{the fact that \texttt{Cristiano Ronaldo} serves for team \texttt{Juventus Football Club S.P.A} can be represented by the image that \emph{Cristiano Ronaldo wearing a Juventus jersey}.} %Thus, a MMKG might have its edges multi-modalized. A relation instance or edge in a KG could also be multi-modalized into appropriate images. 
%
%According to the two multi-modalization ways,
%
%we formally give definitions to the two different kinds of MMKGs below. One type of MMKGs take multi-modal data as attribute values, which we name as {\bf A-MMKG} for short, while the other type of MMKGs take multi-modal data as nodes, which we name as {\bf N-MMKG} for short.
%
\eat{
We formally define the two representation ways for MMKGs below:
\newtheorem{definition}{Definition}
\begin{definition}
  (A-MMKG)
  A type-A MMKG is an MMKG with its multi-modal data as attribute values, denoted as $\mathcal{G=\{E,R,A,V,T_R,T_A\}}$, where $\mathcal{T_A=\mathcal{E}\times\mathcal{A}\times(\mathcal{V_{KG}}\cup\mathcal{V_{MM}})}$ is the set of attribute triples, $\mathcal{V_{KG}}$ is the set of the KG's attribute values and $\mathcal{V_{MM}}$ is the set of multi-modal data.
  \label{def:multi-modal_attributes}
\end{definition}

\begin{definition}
  (N-MMKG)
  A type-N MMKG is an MMKG where its multi-modal data also taken as entities, denoted as $\mathcal{G=\{E,R,A,V,T_R,T_A\}}$, where $\mathcal{T_R=(\mathcal{E_{KG}}\cup\mathcal{E_{MM}})\times\mathcal{R}\times(\mathcal{E_{KG}\cup\mathcal{E_{MM}}})}$ is the set of relation triples, $\mathcal{E_{KG}}$ is the set of KG entities and $\mathcal{E_{MM}}$ is the set of multi-modal data.
  \label{def:multi-modal_nodes}
\end{definition}
}

%In MMKGs, an entity is associated with its attribute or another entity via an RDF triple in the form of ($s$,$p$,$o$), an abbreviation of (\emph{subject, predicate, object}). Specifically, the $s$,$p$,$o$ may have different meanings in different MMKGs. 

%Differently, in N-MMKGs multi-modal data are treated as new entity nodes, thus more relations discovered by multi-modal features are added into the MMKG. For example, in Table \ref{tab:rdf-mmkg-def-2}, the entity \texttt{Place de la Concorde} is associated with an image \emph{Place\_de\_la\_Concorde.jpg} by the relation \texttt{imageOf}. Two images can also be associated by each other with a RDF triple in one of the following relations:

%The existing studies on multi-modalization of KGs mainly focus on the multi-modalization of nodes with images. 

\eat{
Table~\ref{tab:mainstream_mmkg} lists the mainstream MMKGs constructed by texts and images with detailed information, including MMKG type, multi-modalized knowledge, source resources, candidate resources, quality control, scale, image per entity etc. 
}

%Table~\ref{tab:mainstream_mmkg} lists the mainstream MMKGs constructed by texts and images with detailed information.
%
We list mainstream MMKGs constructed with image-based visual knowledge extraction systems in Table~\ref{tab:mainstream_mmkg}(a). 
NEIL~\cite{chen2013neil} annotates each image with a single label by pre-trained classifiers and extracts visual relations by heuristic rules about the locations of extracted objects. 
GAIA~\cite{li2020gaia} extracts fine-grained concepts in the news by object recognition together with fine-grained classification.
Based on the framework of GAIA, RESIN~\cite{wen2021resin} extracts visual news events and identifies related visual entities and concepts as arguments on small-scale resources (news documents). Later, MMEKG~\cite{ma2022mmekg} optimizes some modules and adapt to billion-scale universal events extraction.

MMKGs listed in Table~\ref{tab:mainstream_mmkg}(b) are constructed with symbol grounding. %considering usability and universality.
    IMGpedia~\cite{ferrada2017imgpedia} linkes images from Wikimedia Commons\footnote{a multi-media dataset linking to Wikipedia articles, https://wikimediafoundation.org/our-work/commons/} to DBpedia via the structured Wikipedia data in RDF format already extracted in DBpedia Commons~\cite{vaidya2015dbpedia}, which additionally adds the similarity relations between images. ImageGraph~\cite{onoro2017answering} searches images from search engines with entities in KGs as queries. MMKG~\cite{liu2019mmkg} extends this %flexible construction 
    method to several KGs by aligning entities across them. %, which is mainly used for link prediction and entity alignment. 
    These construction methods based on symbolic entity alignment (such as by linked datasets or URI of entities) focus on the representativeness of images, but the diversity of images is also an important issue due to different contexts and views. % and other factors. 
    Richpedia~\cite{wang2020richpedia} trains an additional diversity retrieval model to select diverse images. The categories of entities in Richpedia are limited to cities, sights, and persons. In addition, VisualSem~\cite{alberts2020visualsem} considers that many entities are non-visualizable entities that should not be searched for images. 
    Therefore it starts with the most typical visual entities and mine other related visual entities iteratively. 
    %VisualSem starts with the most typical visual entities and iteratively mine other related visual entities. 
    However, the small scale of VisualSem is far from satisfying the knowledge demands of downstream applications. %KGbench\cite{bloem2021kgbench} creates several vertial-domain MMKGs by combining publicly available datasets for evluating multi-modal link prediction and node labeling. % KGbench不是通用的，是四个领域MMKG

\newcommand{\tabincell}[2]{\begin{tabular}{@{}#1@{}}#2\end{tabular}}  %表格自动换行

\begin{table*}[t]
    \centering
    \scriptsize
    %\subfloat[MMKGs constructed by labeling images]{
    \subfloat[Image-based visual knowledge extraction systems that could be used to construct MMKGs by labeling images]{
        	\resizebox{\textwidth}{!}
        	{
        		\begin{tabular}{|c|c|c|c|c|c|c|}
        			\hline
        			{\bf System} & {\bf \tabincell{c}{MMKG\\Type}} & {\bf \tabincell{c}{Multi-\\modalized\\Knowledge}} & {\bf \tabincell{c}{Source\\Images}} & {\bf \tabincell{c}{Candidate\\KGs}} & \bf{\tabincell{c}{Quality\\Control}} & {\bf Scale} \\

        			\hline
        			
        			NEIL~\cite{chen2013neil} & N-MMKG & \tabincell{c}{entity,\\concept, \\relation}	
        			& \tabincell{c}{images from\\search engine} & WordNet & \tabincell{c}{semi-supervised\\classification with\\labeled seed images} & \tabincell{c}{1,152 objects,1,034 scenes\\87 attributes,1,703 triples\\(2.5 months)} \\
        			
        			\hline
        			
        			GAIA~\cite{li2020gaia} & N-MMKG &  \tabincell{c}{entity,\\concept}
        			%{entity,\\relation,\\event} 
        			& \tabincell{c}{multimedia\\news\\documents} & \tabincell{c}{Freebase,\\GeoNames} & \tabincell{c}{object detection,\\fine-grained classification,\\heuristic rules} & \tabincell{c}{$\textless$ 457K entities, $\textless$ 67K triples,\\$\textless$ 38K events (including textual\\and visual ones)} \\
        			
        			\hline

        			RESIN~\cite{wen2021resin} & N-MMKG &  \tabincell{c}{entity,\\concept,\\event}
        			%{entity,\\relation,\\event} 
        			& \tabincell{c}{multimedia\\news\\documents} & WikiData & \tabincell{c}{event classification,\\object recognition,\\situation recognition,\\weakly-supervised\\event grounding,\\event relation extraction} & \tabincell{c}{$\textless$ 24 entities, $\textless$ 46 relations,\\$\textless$67 events (including textual\\and visual ones)} \\

                        \hline

                        MMEKG~\cite{ma2022mmekg} & N-MMKG &  \tabincell{c}{event}
        			& \tabincell{c}{Wikipedia,\\BookCorpus,\\CC3M\&CC12M,\\C4(news)} & WordNet & \tabincell{c}{event classification,\\object recognition,\\event relation extraction} & \tabincell{c}{$\textless$ 990K events, $\textless$ 644 event relations\\$\textless$ 863M instance events,\\$\textless$ 934M instance events' relations\\(including textual and visual ones)} \\
        			
        			\hline		
        			
        		\end{tabular}
        		\label{tab:labeling}
        	}
    }

    %\subfloat[MMKGs constructed by symbol grounding]{
    \subfloat[MMKGs constructed by symbol grounding]{
        \resizebox{\textwidth}{!}
    	{
    		\begin{tabular}{|c|c|c|c|c|c|c|c|}
    			
    			\hline
    			
    			{\bf MMKG} & {\bf \tabincell{c}{MMKG\\ Type}} & {\bf \tabincell{c}{Multi-\\modalized\\Knowledge}} &  {\bf{\tabincell{c}{Source\\KGs}}} & {\bf \tabincell{c}{Candidate\\images}} & {\bf \tabincell{c}{Quality \\ Control}} & {\bf Scale} & {\bf \tabincell{c}{Images\\per\\entity}} \\
    			
    			\hline
    			
    			\tabincell{c}{IMGpedia\\~\cite{ferrada2017imgpedia}} & N-MMKG & \tabincell{c}{entity,\\concept,\\relation}
    			%entity
    			& DBpedia & \tabincell{c}{Wikimedia\\Commons} & \tabincell{c}{constructed via\\DBpedia Commons\\} & \tabincell{c}{12.7M links to KG (with\\2.6M DBpedia entities\\/concepts), 3000M  triples\\(including 443M triples\\of 1 visual relation)} & $\textgreater$5.6 \\
    			
    			% 12.7M links to dbpedia, 14.7 images. dbpedia has 2.6M entities
    			
    			\hline
    			
    			\tabincell{c}{ImageGraph\\~\cite{onoro2017answering}} & A-MMKG & \tabincell{c}{entity,\\concept}
    			%entity
    			& FB15K & \tabincell{c}{search\\engine} & \tabincell{c}{disambiguation by\\Wikipedia URI}& \tabincell{c}{15K entities/concepts} & 55.8 \\
    			
    			\hline
    			
    			\tabincell{c}{MMKG\\~\cite{liu2019mmkg}} & A-MMKG & \tabincell{c}{entity,\\concept}
    			%entity
                              & \tabincell{c}{FB15K,\\ DBpedia15K,\\ YAGO15K} & \tabincell{c}{search\\engine}
    			& \tabincell{c}{1.entity alignment \\ cross different KGs\\2.disambiguation by\\Wikipedia URI} & \tabincell{c}{15K entities/concepts} & 55.8 \\
    			
    			\hline
    			
    			\tabincell{c}{Richpedia\\~\cite{wang2020richpedia}} & N-MMKG & \tabincell{c}{entity,\\concept,\\relation}
    			%entity
    			& \tabincell{c}{Wikidata} & \tabincell{c}{search\\engine,\\Wikipedia} & \tabincell{c}{1.disambiguation by\\Wikipedia URI\\2.a diversity retrieval\\model to filter images} & \tabincell{c}{2.8M entities/concepts,\\172M triples\\(including 114.5M triples\\of 3 visual relations)} & 99.2 \\
    			
    			% kg entities:29985 images:2883162
    			
    			\hline
    			
    			\tabincell{c}{VisualSem\\~\cite{alberts2020visualsem}} & N-MMKG & \tabincell{c}{entity,\\concept}
    			%entity
    			& \tabincell{c}{BabelNet} & \tabincell{c}{Wikipedia,\\ImageNet} & \tabincell{c}{1.synsets in ImageNet\\as initial entities pool\\2.mining neighbours\\3.a image-text matching \\model to filter noise} & \tabincell{c}{89.9K entities/concepts,\\13 relations} & 10.4 \\

                    \hline

                    % KGbench是否是通用的呢？如果是，那么就补充，如果不是，就不同添加这个KG信息了。
                    %\tabincell{c}{\zxr{KGbench\\~\cite{bloem2021kgbench}}} & \zxr{N-MMKG} & \tabincell{c}{} 
    			%& \tabincell{c}{} & \tabincell{c}{} 
                    %& \tabincell{c}{} & \tabincell{c}{} & xxx \\
    			
    			%\hline
    			
    		\end{tabular}
    		\label{tab:symbol-grounding}
    	}
    }
    \caption{Mainstream MMKGs (or extraction systems for constructing MMKGs)  and their relevant information}
    \label{tab:mainstream_mmkg}
    \vspace{-2.0em}
\end{table*}

%\subsubsection{Symbolic Representation and Design}\label{section:sec2.5.2}

%In symbolic representation of traditional KG, an entity is represented with a unique ID. In RDF, a uniform resource id (URI) is used as the reference of any objects, such as an entity, a concept or an event. %An image (or other multi-modal data) of an entity in a MMKG is also represented with a URI. 

%There are two ways to multi-modalize nodes: 1) \textbf{considering multi-modal data as an attribute of nodes} and 2) \textbf{considering multi-modal data themselves as nodes}, defined in Definition~\ref{def:multi-modal_attributes} and Definition~\ref{def:multi-modal_nodes}. The first type of MMKG does not change the form of representation of traditional KG, and the second type of MMKG introduces new type of nodes and edges into traditional KG leading to different representation, as illustrated in Figure~\ref{fig:definitions_2mmkg}.

\subsection{Preliminaries on Multi-Modal Techniques}
%The construction and application of MMKGs significantly relate to existing multi-modal research in Computer Vision (CV). 

Modality refers to the particular way in which something exists, is experienced or is done~\cite{baltruvsaitis2018multimodal}. 
In computer science and artificial intelligence, a problem is characterized as multi-modal if it involves data of multiple modalities. 
%
%A multi-modal task integrates and models multiple communicative modalities to acquire knowledge or comprehension from the multi-modal data~\cite{baltruvsaitis2018multimodal}.
%
%Typical multi-modal tasks with images and texts include image caption, visual grounding, visual question answering, cross-modal retrieval, etc.
%
Typical multi-modal tasks with images and texts include image caption~\cite{cornia2019show}, visual question answering~\cite{antol2015vqa}, and cross-modal retrieval~\cite{karpathy2014deep}, etc.
We will introduce how MMKGs are applied in these applications in Sec.~\ref{sec:outKG}.
But before MMKGs, people mainly focus on multi-modal learning, and more recently the Vision and Language Pre-trained Models (VL-PTMs), which will be briefly introduced below.

\noindent {\bf Multi-Modal Learning}. Multi-modal learning focuses on modeling the correspondences among multiple modalities, which includes:
1) {\it Multi-modal Representation} aims to use the complementary of multi-modality to learn feature representation. The existing efforts either project the multiple modalities into a unified space~\cite{d2015review}, or represent every single modal in its own vector space which satisfies certain constraints like linear correlation~\cite{kiros2014unifying}.
2) {\it Multi-modal Translation} learns to translate from a source instance in one modality to a target instance in another, including example-based~\cite{karpathy2014deep} and generative translation models~\cite{xu2015show}.
3) {\it Multi-modal Alignment} aims to find the correspondences between different modalities. It can either be directly applied in some multi-modal tasks such as visual grounding or as a pre-training task in VL-PTMs~\cite{ramachandram2017deep}.
4) {\it Multi-modal Fusion} aims to join information from different modalities to perform a prediction~\cite{baltruvsaitis2018multimodal}, where various attention mechanisms~\cite{anderson2018bottom,zhu2020multimodal} are applied to model the interaction between different features in the cross-modal module. 
5) {\it Multi-modal Co-Learning} aims to alleviate the low-resource problems in a certain modality by leveraging the resources of other modalities through the alignment between them~\cite{baltruvsaitis2018multimodal}.

\eat{
\noindent {\bf Multi-Modal Learning}.  The multi-modal learning mainly focus on modeling the correspondences among multiple modalities to understand multi-modal data, which faces some fundamental challenges as listed below:

\begin{enumerate}[1)]\setlength{\itemsep}{0pt}
    \item \zxr{\sout{{\it Multi-modal Representation.} The multi-modal representation uses the potentially complementary of multi-modality to learn feature representation. The existing efforts either projects the multi-modality into a unified space~\cite{d2015review} with VGG~\cite{simonyan2014very}, ResNet~\cite{he2016deep}, or represents every single modal in its own vector expression space which satisfies certain constraints like linear correlation~\cite{kiros2014unifying}.
    }}
    
    \item \zxr{\sout{{\it Multi-modal Translation.} Multi-modal translation learns to translate from a source instance in one modality to a target instance in another. The example-based translation models build bridges between different modality through dictionary~\cite{karpathy2014deep,cao2016deep}, while the generative translation models build a more flexible model which can transform one modal to another~\cite{vinyals2015show,xu2015show}.
    }}
    
    \item \zxr{\sout{{\it Multi-modal Alignment.} Multi-modal alignment aims to find the correspondences between different modalities. It can either be directly applied in some multi-modal tasks such as visual grounding, or be taken as a pre-training task in multi-modal pre-trained language models~\cite{ramachandram2017deep}.
    }}
    
    \item \zxr{\sout{{\it Multi-modal Fusion.} Multi-modal fusion refers to the process of joining information from different modalities to perform a prediction~\cite{baltruvsaitis2018multimodal}, where various attention mechanism such as gated cross-modality attention~\cite{zhu2020multimodal}, bottom up attention~\cite{anderson2018bottom} etc. are applied to model the interaction between different kinds of features in the cross-modal module. 
    }}
    
    \item \zxr{\sout{{\it Multi-modal Co-Learning.} Multi-modal co-learning aims to alleviate the low-resource problems in a certain modality by leveraging the resources of other modalities through the alignment between them~\cite{baltruvsaitis2018multimodal}. 
    }}
    %The typical methods include zero-shot learning\cite{socher2013zero} and transfer learning\cite{albanie2018emotion}.
    
    % \item {\it Multi-modal Co-Learning.} Multi-modal co-learning aids the modality which may have more resources to another\spl{~\cite{baltruvsaitis2018multimodal}}. Co-learning allows the model to learn more alignment and fusion between different modalities data. Co-learning is widely used to solves the problem of unbalanced data and domain gap in the zero-shot learning \cite{farhadi2009describing} which can generates the unseen information of uni-modal through the mapping from the known modal. 
\end{enumerate}
}

%\noindent {\bf \zxr{\sout{Multi-Modal Pre-trained Language Model}

\eat{
Based on a large-scale unsupervised multi-modal data set with text-image pairs, much recent efforts works on learning a multi-modal pre-trained language model with some designed self-supervised pretraining tasks including masked language model, sentence image alignment, masked region label classification, masked region features regression, masked object prediction, etc~\cite{ramachandram2017deep}.
VL-PTMs have proven their effectiveness in improving many downstream tasks~\cite{chen2019uniter,li2020unicoder,kim2021vilt,li2020unimo,yu2020ernie}.

Based on big data and large scales of models, VL-PTMs learn a lot of visual knowledge, such as co-occurrence objects and cross-modal alignment. A typical example is CLIP trained on 400 million text-image pairs, which greatly improved image classification and
cross-modal retrieval performance. To improve fine-grained understanding, many VL-PTMs add cross-modal object alignment~\cite{li2020oscar,yao2021filip,li2022grounded}, relation alignment~\cite{yu2021ernie,cui2021rosita} tasks optimize the pre-training and achieve improvement. However, the increasing number of data and scale of models' parameters does not solve the problem of long-tailed entity, scenarios, common sense, background knowledge and interpretable reasoning in practical applications~\cite{}. It has been confirmed that structural knowledge (such as entity knowledge~\cite{xiong2019pre-trained,lauscher2019specializing,shen2020exploiting} and triple knowledge~\cite{sun2020colake,liu2020k,guan2020knowledge,sun2019ernie,he2019integrating,peters2019knowledge,wang2021kepler}) can rapidly improve the understanding and logical reasoning ability of the pre-trained language models in textual modality~\cite{xiong2019pre-trained,lauscher2019specializing,shen2020exploiting,sun2020colake,liu2020k,guan2020knowledge,sun2019ernie,he2019integrating,peters2019knowledge,wang2021kepler}. MMKGs are still non-substitutable sources of knowledge for VL-PTMs, which is to be fully explored.

}

\noindent {\bf VL-PTMs}. %
Recently, many large companies and research institutions including OpenMind~\cite{radford2021learning}, Microsoft~\cite{chen2019uniter,li2022grounded} and Huawei~\cite{yao2021filip} etc. pay great efforts on training large VL-PTMs based on large-scale unsupervised multi-modal data.
A typical VL-PTM example is CLIP~\cite{radford2021learning} trained on 400 million text-image pairs, which significantly improves the performance of image classification and cross-modal retrieval.
Based on massive multi-modal data and large-scale models, VL-PTMs could learn extensive implicit cross-modal knowledge with some designed self-supervised pretraining tasks, such as masked language model, sentence image alignment, masked region label classification, masked region features regression, masked object prediction, etc.
Furthermore, to improve fine-grained cross-modal understanding, some work also add cross-modal object alignment~\cite{li2020oscar,yao2021filip,li2022grounded}, relation alignment~\cite{yu2021ernie,cui2021rosita} tasks to optimize the pre-training process.

\eat{
Based on big data and large scales of models, VL-PTMs learn a lot of visual knowledge, such as co-occurrence objects and cross-modal alignment. A typical example is CLIP trained on 400 million text-image pairs, which greatly improved image classification and
cross-modal retrieval performance. To improve fine-grained understanding, many VL-PTMs add cross-modal object alignment~\cite{li2020oscar,yao2021filip,li2022grounded}, relation alignment~\cite{yu2021ernie,cui2021rosita} tasks optimize the pre-training and achieve improvement. 
However, the increasing number of data and scale of models' parameters does not solve the problem of long-tailed entity, scenarios, common sense, background knowledge and interpretable reasoning in practical applications~\cite{}. It has been confirmed that structural knowledge (such as entity knowledge~\cite{xiong2019pre-trained,lauscher2019specializing,shen2020exploiting} and triple knowledge~\cite{sun2020colake,liu2020k,guan2020knowledge,sun2019ernie,he2019integrating,peters2019knowledge,wang2021kepler}) can rapidly improve the understanding and logical reasoning ability of the pre-trained language models in textual modality~\cite{xiong2019pre-trained,lauscher2019specializing,shen2020exploiting,sun2020colake,liu2020k,guan2020knowledge,sun2019ernie,he2019integrating,peters2019knowledge,wang2021kepler}. MMKGs are still non-substitutable sources of knowledge for VL-PTMs, which is to be fully explored.
}

\subsection{Discussions}
Although there there is already much research on multi-modal learning and VL-PTMs, there is still an emerging trend to introduce MMKGs to help enhance multi-modal tasks. In general, MMKGs could benefit multi-modal tasks in the following aspects.

\begin{enumerate}[1)]\setlength{\itemsep}{0pt}
	\item MMKGs provide sufficient background knowledge to enrich the representation of entities and concepts, especially for the long-tail ones. 
    For instance, \cite{shi2019knowledge} uses auxiliary commonsense knowledge to enhance the representation of image and text to improve image-text matching.
    %For instance, auxiliary commonsense knowledge is introduced to enhance the representation of image and text, leading to improved performance of image-text matching~\cite{shi2019knowledge}.
	
	\item MMKGs enable the understanding of unseen objects in images. Unseen objects pose a great challenge to statistic-based models. Symbolic knowledge alleviates the difficulty by providing symbolic information about unseen objects or establishing semantic relations between seen objects and unseen objects. For example, ~\cite{mogadala2017describing} uses external symbolic knowledge to guide the generation of captions for unseen novel visual objects.% with parallel captions.
	
	\item MMKGs enable multi-modal interpretable reasoning. For example, the OK-VQA dataset~\cite{marino2019ok}, which contains only questions that require external knowledge to answer, is built to test the reasoning capability of VQA models.
	
	\item MMKGs usually provide multi-modal data as additional features to bridge the information gaps in some NLP tasks. In the case of entity recognition, the image could provide sufficient information to identify whether ``Rocky'' is the name of a dog or a person~\cite{zhang2018adaptive}.

    \item MMKGs provide explicit and fine-grained cross-modal correlation knowledge, which is complementary to the implicit knowledge learned by VL-PTMs. Besides, MMKGs have advantages on providing long-tail knowledge, background knowledge, and fine-grained knowledge compared with VL-PTMs~\cite{marino2021krisp}.

        %\item \zxr{Compared to implicit knowledge from VL-PTMs, MMKGs provide explicit and fine-grained cross-modal correlation knowledge, which could be directly leveraged by these downstream application models in cross-modal understanding and logic reasoning. For example, Oscar~\cite{li2020oscar} adds object tags to connect visual objects and textual tokens explicitly, which has been a solid baseline in cross-modal retrieval, image caption and VQA for a long time.}

        %\item It has been confirmed that structural knowledge (such as entity knowledge~\cite{xiong2019pre-trained,lauscher2019specializing,shen2020exploiting} and triple knowledge~\cite{sun2020colake,liu2020k,guan2020knowledge,sun2019ernie,he2019integrating,peters2019knowledge,wang2021kepler}) can rapidly improve the understanding and logical reasoning ability of the pre-trained language models in textual modality~\cite{xiong2019pre-trained,lauscher2019specializing,shen2020exploiting,sun2020colake,liu2020k,guan2020knowledge,sun2019ernie,he2019integrating,peters2019knowledge,wang2021kepler}. 
        %MMKGs are still non-substitutable sources of knowledge for VL-PTMs, which is to be fully explored.
        
\end{enumerate}

To sum up, previous efforts to use multi-modal information are still limited without the support of large-scale MMKG. Multi-modal tasks can be further improved when MMKGs are available. %Thus the key is reduced to how to construct such a MMKG with acceptable cost, which is the focus of this survey. 

\section{Construction}
\label{sec:construction}

\begin{figure}[t]
	\centering
	\begin{tabular}{@{}c@{}}
		\includegraphics[width=1\linewidth]{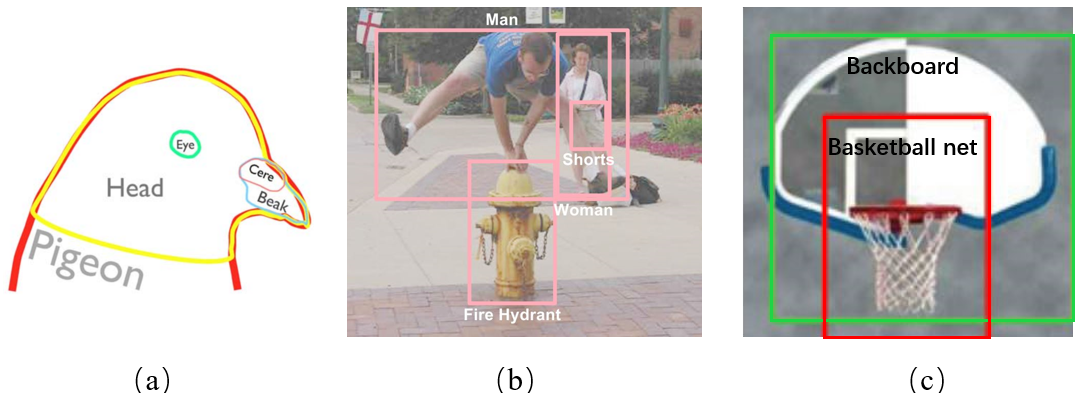} \\% [\abovecaptionskip]
	\end{tabular}
	\caption{Examples of labeling images: (a) labeling components after image segmentation in Visipedia \protect\cite{perona2010vision}; (b) labeling objects with bounding boxes in Visual Genome \protect\cite{krishna2017visual}; (c) labeling two objects where one is a part of the other in NEIL \protect\cite{chen2013neil}, e.g., \emph{PartOf}(\texttt{Basketball net}, \texttt{Backboard}).}
	%是否标明标注是自动还是非自动？  \caption{Examples of labeling images: (a) labeling components after image segmentation in Visipedia \protect\cite{perona2010vision} by experts; (b) labeling objects with bounding boxes in Visual Genome \protect\cite{krishna2017visual} by crowd workers; (c) labeling two objects where one is a part of the other in NEIL \protect\cite{chen2013neil} automatically.}
	\label{fig:label}
	\vspace{-1.0em}
\end{figure}

The essence of MMKG construction is associating symbolic knowledge in a traditional KG, including entities, concepts, relations, etc., with their corresponding images. 
Two opposite ways to complete the task are (1) labeling images with symbols in KG and (2) grounding symbols in KG to images.
%
%One way conducts the association from images to symbols, i.e., labeling images with corresponding symbolic knowledge, while the other way prefers the opposite way, i.e., finding corresponding images for symbolic knowledge.
%
We elaborate on the two categories of solutions in Sec.~\ref{sec:labelingImages} and Sec.~\ref{sec:symbolGrounding} respectively. We finally discuss the differences between the two solutions in Sec.~\ref{secion:sec3.3}.
%We finally discuss the differences between the symbol grounding way and the image labeling way for MMKG construction in Sec.~\ref{secion:sec3.3}.

\subsection{From Images to Symbols: Labeling Images}
\label{sec:labelingImages}

The CV community has developed many image labeling solutions, which could be leveraged in labeling images with structural symbols (e.g., concepts or entities) in KG. For example, NEIL~\cite{chen2013neil} links images to WordNet\cite{miller1995wordnet}, and ImageSnippets~\cite{warren2018bounding,warren2021knowledge} links images to DBPedia\cite{auer2007dbpedia}.
Most image labeling solutions learn the mapping from image content to a wide variety of label sets, including objects, scenes, entities, attributes, relations, events and other symbols. 
The learning procedure is supervised by human-annotated datasets, which require the crowd workers to draw bounding boxes and annotate images or regions of images with given labels, as illustrated in Figure~\ref{fig:label}.
%
%Even in automatic labeling systems, a lot of pre-labeled images and bounding boxes are also needed to train the detection and classification model. %Many datasets are built in this way, as listed in Table~\ref{tab:labeling}. These annotated datasets are used for image labeling models. %

Some well-known image-based visual knowledge extraction systems are %constructed, 
as listed in Table~\ref{tab:mainstream_mmkg}(a), which could be utilized for constructing MMKGs through image labeling.
According to the category of symbols to be linked, the process of linking images to symbols could be divided into several fractionized tasks: {\it visual entity/concept extraction} (Sec.~\ref{section:sec3.1.1}), {\it visual relation extraction} (Sec.~\ref{section:sec3.1.2}) and {\it visual event extraction} (Sec.~\ref{section:sec3.1.3}).
%
%Note that not all the tasks above are completed for building a MMKG as listed in Table~\ref{tab:labeling_3_methods}.
%
%In the following three subsections, we discuss the challenges, progresses and opportunities of each task respectively.

\subsubsection{Visual Entity/Concept Extraction}\label{section:sec3.1.1}

Visual entity (or concept) extraction aims to detect and locate target visual objects in images and then label these objects with entity (or concept) symbols in KG. %These objects are usually coarse-grained such as ...

\vspace{5pt}
\noindent{\bf CHALLENGES}. The main challenge with this task lies in how to learn an effective fine-grained extraction model without a large-scale, fine-grained, well-annotated concept and entity image dataset.
Although there are rich well-annotated image data in CV, these datasets are almost coarse-grained concept images, which could not meet the requirements of MMKG construction for image annotation data of fine-grained concepts and entities.
%
%On the other hand, although there are a large amount of fine-grained image-caption pairs on the Internet, which make it possible to obtain adequate fine-grained data without relying on annotated data. But weakly supervised extraction is difficult and not very accurate. \textbf{How to extract visual entities accurately by weakly supervised image-caption pairs} ?

%Visual entities can be extracted and labeled by two types of methods: supervised methods and weakly supervised methods. If annotated data are sufficient, visual entities and concepts are usually extracted using supervised classification or object recognition approaches. %However, in practice annotated data with high-quality are not always sufficient. 
%
%But in the case of inadequate annotations, an alternative way of extracting visual entities or concepts is to learn the weak correspondence between visual regions and phrases in the massive image-caption pairs from the internet. %Finally, theses labels of extracted visual objects need to be aligned to symbols in a traditional KG to construct the hierarchical structure of visual entities in a MMKG.

\begin{figure}[t]
	\centering
	\begin{tabular}{@{}c@{}}
		\includegraphics[width=0.7\linewidth]{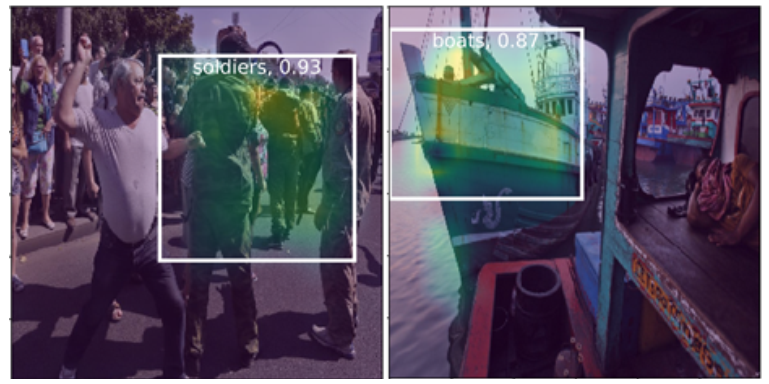} \\% [\abovecaptionskip]
	\end{tabular}
	\caption{The heatmap for detected visual entities (\texttt{Soldier} and \texttt{Boats}) in two example images by visual grounding in GAIA~\cite{li2020gaia}, where the stronger the correlation between a pixel and a word, the warmer the color of the pixel.}
	\label{fig:heatmap}
\end{figure}

\begin{figure}[t]
	\centering
	\begin{tabular}{@{}c@{}}
		\includegraphics[width=1\linewidth]{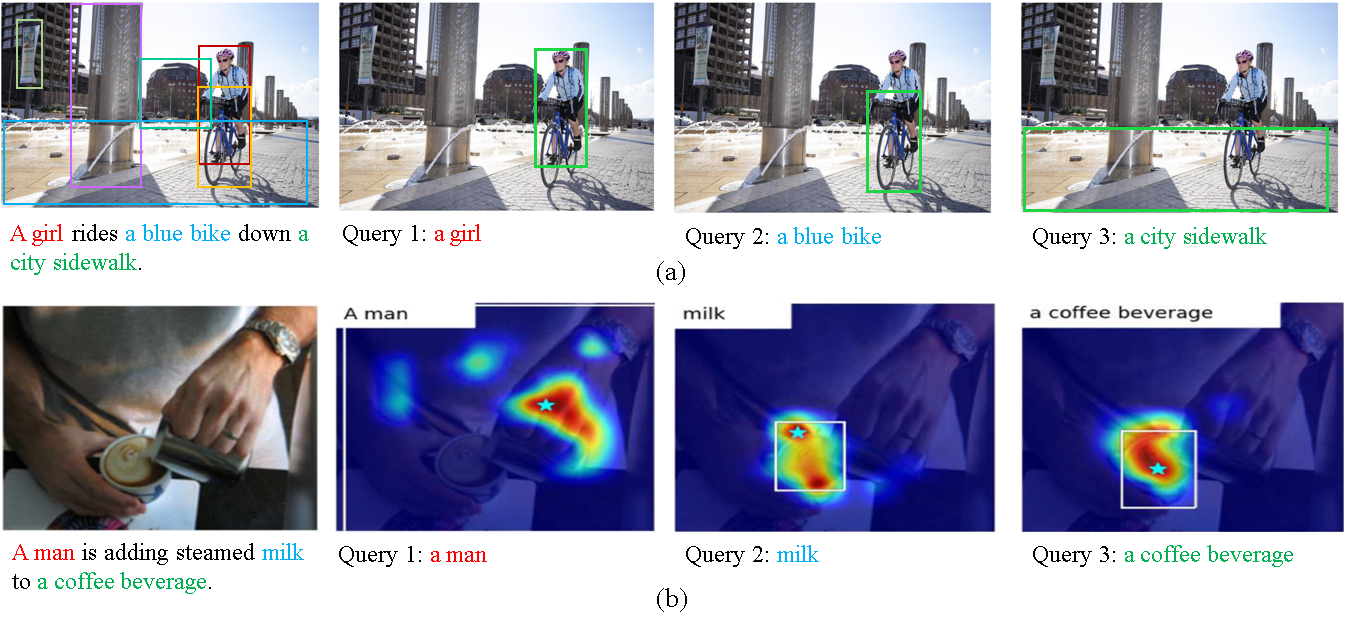}\\
		%[\abovecaptionskip]
	\end{tabular}
	\caption{Two kinds of weakly supervised visual entity extraction: (a) the attention-based method~\cite{chen2018knowledge} and (b) the saliency-based method~\cite{ramanishka2017top}. The first method selects the most relevant bounding boxes to given  phrases. The second method selects the most sensitive pixels to given phrases. }
	\label{fig:weak_supervision}
	\vspace{-1.0em}
\end{figure}

\vspace{5pt} \noindent{\bf PROGRESSES}. The existing efforts with visual entity/concept extraction could be roughly divided into two categories: 1) object recognition methods, which label a visual entity/concept by classifying the region of a detected object; and 2) visual grounding methods, which label a visual entity/concept by mapping a word or phrase in a caption to the most relevant region.

1) {\it Object Recognition Methods.} In early works, images provided by users and researchers are usually simple and there is only one object in one image, which can be processed by classification models. But images in our real life could be too complex to be represented with only one label. Thus we need to tag different visual units with different labels.

In order to distinguish several visual entities in images, pre-trained detectors and classifiers are needed to label visual entities (as well as attributes and scenes) with their locations in the images. These detectors are trained with supervised data from public images-text datasets~\cite{li2020gaia} (such as MSCOCO~\cite{lin2014microsoft}, Flickr30k~\cite{young2014image}, Flick30k Entities~\cite{plummer2015flickr30k} and Open Images~\cite{OpenImages}). %or pre-labeled seed images~\cite{chen2013neil}. 
In the detection process, detectors (e.g., face detectors based on MTCNN or vehicle detectors based on Faster-RCNN) capture a set of region proposals for possible objects. In the recognition process, the pre-trained classifiers pick out region proposals that do contain objects and recognize candidate visual objects with entity-level (e.g., \texttt{BMW 320}) or concept-level (e.g., \texttt{Car}) labels.
%
%The mean average precisions (MAP) of object detection methods used in GAIA\cite{li2020gaia} is 43\% on the benchmark MSCOCO and 95.4\% on the benchmark FDDB. 
%
%\zxr{\sout{
%During the detection procedure, detectors capture a set of region proposals for possible objects and pick out proposals that do contain objects.
%
%At the locations detected by various detectors, such as face detectors based on MTCNN and vehicle detectors based on Faster-RCNN, the pre-trained classifiers recognize candidate visual objects with entity-level (e.g., \texttt{BMW 320}) or concept-level (e.g., \texttt{Car}) labels.
%}}
%
%
%\zxr{The average precision of SOTA model DINO\cite{caron2021emerging}, followed with SwinV2\cite{DBLP:journals/corr/abs-2111-09883}, Florence\cite{DBLP:journals/corr/abs-2111-11432} and GLIP\cite{DBLP:journals/corr/abs-2112-03857}, is 63.3\% on MSCOCO 2017 test-dev.}
%
Since many recognized objects are duplicated instances of the same entities at different viewpoints, positions, poses and appearances, %a common way to process is to do clustering on all the regions with recognized objects, 
a common way to process is to cluster all the regions with recognized objects,
and only the central one of each cluster will eventually be the output as a new visual entity~\cite{li2020gaia}.
However, the disadvantage of these supervised solutions is that only a limited number of visual entities under pre-defined labels could be recognized. 
% ??? 如果不写MAP, pros and cons / SOTA那个问题咋回答呢？？？？
The precision of the visual object extraction model used in GAIA is only 43\% on the benchmark MSCOCO~\cite{li2020gaia}. It requires much pre-processing work for fine-grained recognition~\cite{li2020gaia}, such as pre-defined rules, pre-trained fine-grained detectors, etc.

%, which requires additional pre-defined work for fine-grained recognition~\cite{li2020gaia,baumgartner2020towards. 
%pre-defined rules,predetermined lists of recognizable entities~\cite{li2020gaia,baumgartner2020towards}

%The mean average precision (MAP) of the visual object extraction model used in GAIA is 43\% on the evaluation of benchmark MSCOCO\cite{li2020gaia}.

%xxxSuppose the automatic labeling solution is to support a large number of labels (such as billions of entities), then it requires much more preprocessing work, such as pre-defining rules, predetermining lists of recognizable entities, pre-training fine-grained detectors and classifiers, etc., which deteriorates the practicability of the solutions~\cite{li2020gaia}. A typical scenario is to identify a specific person entity~\cite{li2020gaia,baumgartner2020towards}. The dilemma is that the results will be inaccurate without these additional constraints. For example, the precision of the visual object extraction model used in GAIA is only 43\% on the benchmark MSCOCO~\cite{li2020gaia}.

2) {\it Visual Grounding Methods.} In visual entity extraction, training detectors need a large amount of labeled data with bounding boxes and pre-defined schemas with a fixed set of concepts~\cite{wang2016structured}, which is challenging for large-scale visual knowledge acquisition. Fortunately, many image-captions pairs from the web %(e.g., news websites) 
weakly supervise the extraction of visual knowledge without relying on the labeled bounding boxes. Therefore, the visual entity extraction problem is reduced to an open-domain visual grounding problem, which aims to locate the corresponding image region of each phrase in a caption to obtain visual objects with their labels.
%so as to obtain visual objects in an image with their labels.

%When extracting information from weakly supervised image-caption pairs, we often directly select active pixels for a given word as the region of visual objects based on the spatial heatmap (e.g., the heatmap in Figure~\ref{fig:heatmap}).
In the extraction process, we often select active pixels for the given word as the region of visual objects based on the spatial heatmap, as shown in Figure~\ref{fig:heatmap}. 
In the cross-modal unified vector space, the heatmap of each phrase can be learned by attention-based methods and saliency-based methods, as shown in Figure~\ref{fig:weak_supervision}.
%With text and image representations shared in the same semantic space, the heatmap of each phrase can be learned by attention-based methods and saliency-based methods as cross-modal weight, as shown in Figure~\ref{fig:weak_supervision}. 
%
Saliency-based methods treat the marginal effects~\cite{samek2019explainable} of pixels to a given phrase by gradient computation as the heatmap value, and attention-based methods treat the cross-modal relevance as the heatmap value.
%During the training time, saliency-based methods directly treat the marginal effects~\cite{samek2019explainable} of pixels to a given phrase by gradient computation as the value of heatmap, and attention-based methods treat the cross-modal relevance as the value of heatmap. %It has been demonstrated that salience methods are more faithfully interpretable\cite{bastings2020elephant}. 
%
However, since some salience methods are too sensitive to input changes to produce reliable %and intuitive
results~\cite{hooker2019benchmark,kindermans2019reliability,bastings2020elephant}, thus attention-based methods~\cite{wang2016structured,chen2018knowledge,akbari2019multi,li2020gaia,li2020cross} are more studied than saliency-based methods~\cite{ramanishka2017top,zhang2018top} on locating visual objects.
%However, since some salience methods are too sensitive to input changes to produce reliable and intuitive results~\cite{hooker2019benchmark,kindermans2019reliability,bastings2020elephant}, thus attention-based methods~\cite{wang2016structured,rohrbach2016grounding,chen2018knowledge,xu2018attngan,javed2018learning,akbari2019multi,li2020gaia,li2020cross} are more well-studied than saliency-based methods~\cite{simonyan2013deep,cao2015look,ramanishka2017top,zhang2018top} on locating visual objects.
%
For example, the heatmap values in GAIA~\cite{li2020gaia} are similarities between image regions and entity mentions in a caption, and those in~\cite{li2020cross} are similarities between image regions and possible event argument role types.
%For example, the heatmap in GAIA~\cite{li2020gaia} is generated based on the similarity between image regions and each entity mention in a caption, and that in~\cite{li2020cross} is generated based on the similarity between image regions and possible event argument role types. %, which brings additional type constraint to the matching. 
At test time, the heatmap is thresholded to obtain a suitable bounding box of a visual object. 
If there is no overlap between the new bounding box and existing visual entities/concepts in KGs, the bounding box will be created as a new visual entity or concept.
%If there is no overlap between the bounding boxes of the existing visual entities/concepts in KGs and the new bounding box, the bounding box will be created as a new visual entity or concept.
%
%\zxr{The accuracies of the SOTA model OFA\cite{wang2022ofa}, followed with MDETR\cite{DBLP:conf/iccv/KamathSLSMC21}, VILLA\cite{DBLP:conf/nips/Gan0LZ0020}, UNITER\cite{DBLP:conf/eccv/ChenLYK0G0020}, are 80.7\%-94.0\% on referring expression comprehension datasets\cite{DBLP:conf/eccv/YuPYBB16,DBLP:conf/cvpr/MaoHTCY016}.}

The located visual objects via visual grounding include entities, concepts and attributes with acceptable accuracy. The accuracy of visual grounding methods used in GAIA~\cite{li2020gaia} is 69.2\% on Flickr30k. However, inconsistent semantic scales of images and texts may lead to incorrect matching. For example, \texttt{troops} may be mapped to \emph{several individuals wearing military uniforms}, and \texttt{Ukraine (country)} may be mapped to \emph{a Ukrainian flag}, both of which are relevant but not equivalent.

%\zxr{\sout{
%Although the open-domain visual grounding does not rely on labeled data with bounding boxes, in practice, human verification is still needed due to mismatching. Some efforts attempts to add constraints on common concepts, relations and event arguments into the training stage to increase supervision information. The precision of visual grounding is less than 70\% in works related to the construction of MMKGs~\cite{li2020gaia}. 
%
%The located visual objects via visual grounding could be entities (e.g., Barack Hussein Obama), concepts (e.g., place, car, stone), and attributes (e.g., red, short). % and actions (e.g. throw). 
%However, inconsistent semantic scales of images and text may lead to incorrect matching. For example, \texttt{troops} may be mapped to \emph{several individuals wearing military uniforms}, and \texttt{Ukraine (country)} may be mapped to \emph{a Ukrainian flag}, both of which are relevant but not equivalent. 
%}}

\begin{figure}[t]
	\centering
	\begin{tabular}{@{}c@{}}
		\includegraphics[width=1\linewidth]{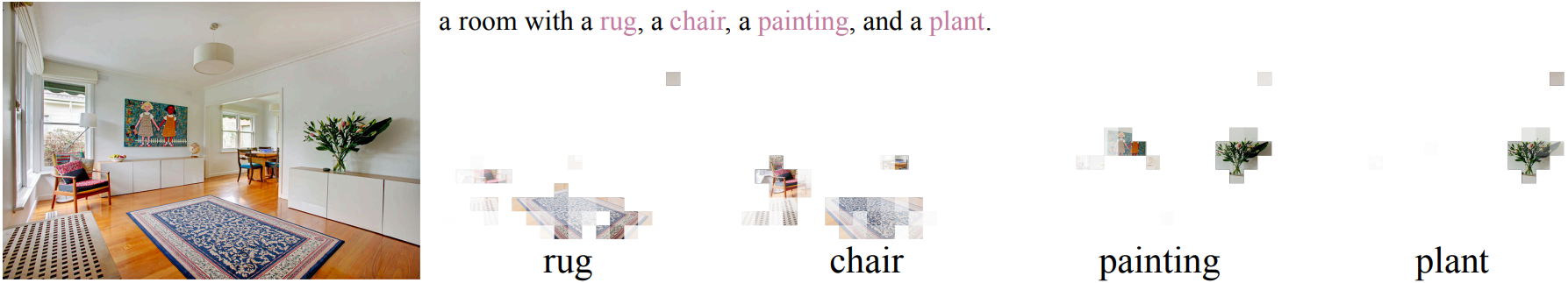} \\% [\abovecaptionskip]
	\end{tabular}
	\caption{Weakly supervised visual entity extraction via %the self-attention mechanism of 
	VL-PTMs. This figure shows the most relevant regions of an image to given words in a caption through self-attention mechanism of ViLT~\cite{kim2021vilt}.}
	\label{fig:vilt_attention_map}
	\vspace{-1.0em}
\end{figure}

\vspace{5pt} \noindent{\bf OPPORTUNITIES}. 
% 这部分要删减换例子了，因为原文SOTA全部是Transformers的
1) {\it VL-PTMs Based Extraction.} VL-PTMs bring new opportunities to nearly all cross-modal downstream tasks, including the detection of visual entities and concepts~\cite{caron2021emerging, wang2022ofa}. The mapping of image patches and words can be directly visualized in the self-attention maps of the model without additional training. An example of the prediction with ViLT~\cite{kim2021vilt} is shown in Fig.~\ref{fig:vilt_attention_map}. 
It is proved that VL-PTMs such as CLIP~\cite{radford2021learning}, trained on hundreds of millions of image-text data, can recognize many popular entities such as famous people and landmarks with high accuracy~\cite{hessel2021clipscore}.
2) {\it Taxonomy Extension.} Some visual objects with multiple reasonable labels indicate different semantic levels. For example, an image of \emph{a boy} can be labeled as \texttt{Person}, \texttt{Man} and \texttt{Boy}. To reduce the ambiguity, 
we should find an appropriate extension semantic level for the labels of images in the taxonomy. \cite{krishna2017visual} fuses aforementioned multiple labels into the lowest common ancestor node of these synsets (i.e., \texttt{Person}), which may lead to many coarse-grained labels. \cite{zhang2019vision} limits the scale of independent concepts' labels by setting a small value of maximum extension level to avoid too many related images. More nodes should be further searched recursively in the taxonomy consisting of hyponyms of the ancestor node to select the most semantically consistent label with the given visual object.
%Except for specific domains such as political persons, landmarks and flags\cite{li2020gaia}, this problem hasn't been well solved in current works.} 
%

\subsubsection{Visual Relation Extraction}
\label{section:sec3.1.2}

Visual relation extraction aims to identify semantic relations among detected visual entities (or concepts) in images and then label them with the relations in KGs~\cite{chen2013neil}.

\vspace{5pt} \noindent{\bf CHALLENGES}. 
Although visual relation detection has been studied extensively in the CV community, most detected relations are superficial visual relationships between visual objects such as \texttt{(Person, standing on, Beach)}. Differently, for the purpose of constructing MMKG, the visual relation extraction task aims to identify more general types of semantic relations that are defined in KGs, such as \texttt{(Jack, spouse, Rose)}.

\vspace{5pt} \noindent{\bf PROGRESSES}. The existing efforts on visual relation extraction can be roughly put into two categories: rule-based relation extraction and statistic-based relation extraction. Some other work mainly focuses on long-tail relations and fine-grained relations, which will also be covered in the following.

1) {\it Rule-based Relation Extraction}. 
Traditional rule-based methods mainly focus on specific relations types, such as spatial relation~\cite{kulkarni2013babytalk, elliott2013image} and action relation~\cite{yao2010grouplet, yao2010modeling, maji2011action, antol2014zero}. 
%The criteria are usually predefined by experts and the discriminative features are scored and selected by heuristic methods.
Experts usually predefine the criteria, and the discriminative features are scored and selected by heuristic methods.

\eat{
Traditional rule-based methods mainly focus on some specific types of relations, such as spatial relation~\cite{kulkarni2013babytalk, elliott2013image} and action relation~\cite{yao2010grouplet, yao2010modeling, maji2011action, antol2014zero, gkioxari2015contextual}. The criteria are usually pre-defined by experts and the discriminative features are scored and selected by heuristic methods.
}

In rule-based methods, the relations are determined based on label types of visual objects and the relative locations of regions. For example, if the bounding box of one object is always within that of another, there may be a \texttt{PartOf} relation between them. Table~\ref{tab:relation} lists several visual relations detected in NEIL, where the average detection accuracy of all 1703 relations is 79\%~\cite{chen2013neil}. During the extraction in NEIL, the detected relation between a pair of objects is, in turn, an additional constraint for new instance labeling. For example, ``\emph{Wheel is a part of Car}'' indicates that it is more likely for a \texttt{Wheel} to appear in the bounding box of a \texttt{Car}. Rule-based methods provide highly accurate visual relations, but require much manual manipulation, which is less practical in large-scale MMKG construction.
\begin{table*}
	\centering
	\scriptsize
	\begin{tabular}{ccc}
		\toprule
		relation type & example & images \\
		\midrule
		%Concept-Concept & Eye is a part of Children. & %\includegraphics[width=0.1\textwidth]{child-3473596_1280.jpg}\\
		Concept-Concept & \tabincell{c}{\texttt{Keyboard} is a part\\ of \texttt{Laptop}.} &
		\includegraphics[width=0.073\textwidth]{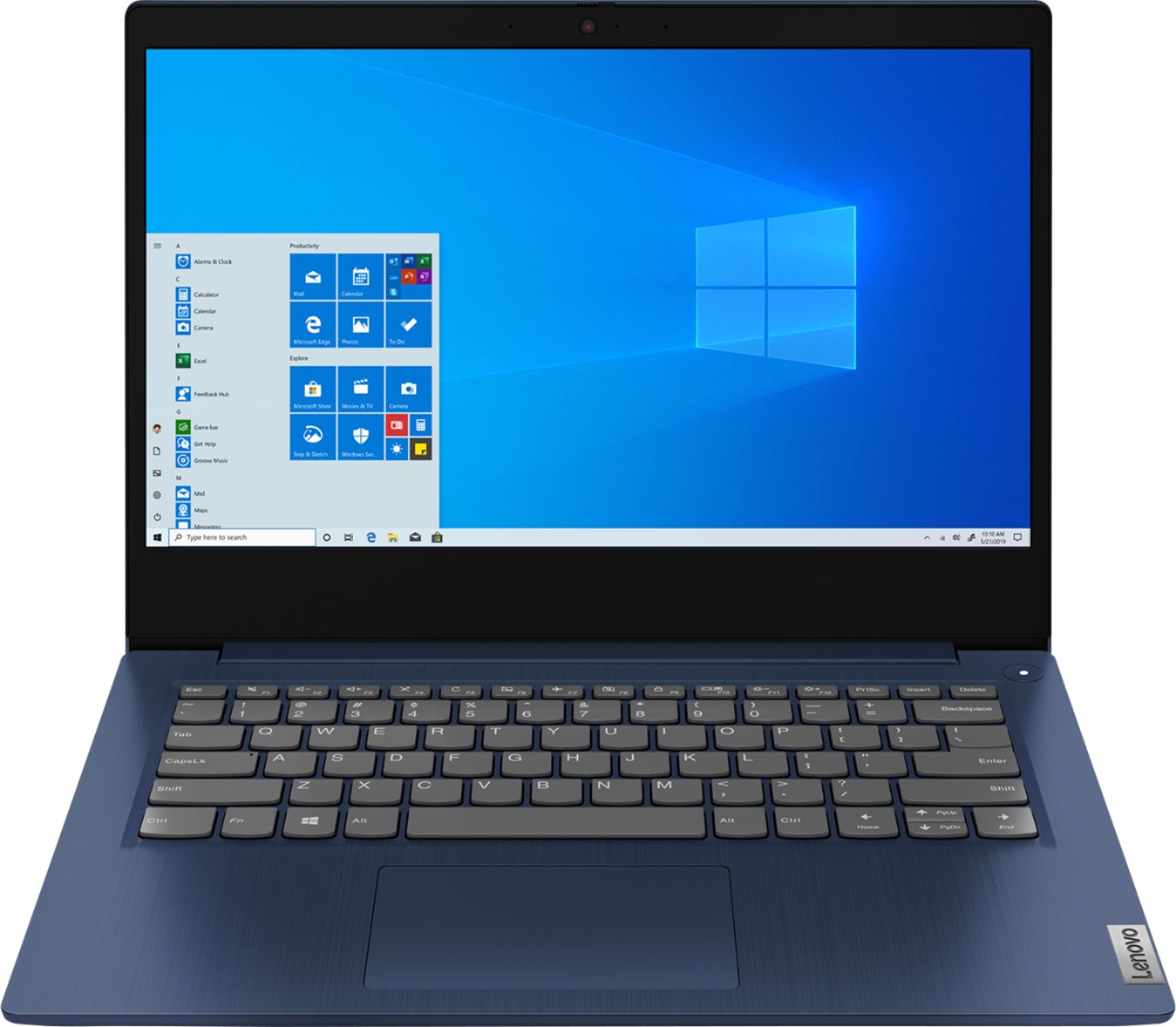}\\
		Entity-Concept & \tabincell{c}{\texttt{BMW 320} is a kind\\ of \texttt{Car}.} & \includegraphics[width=0.1\textwidth]{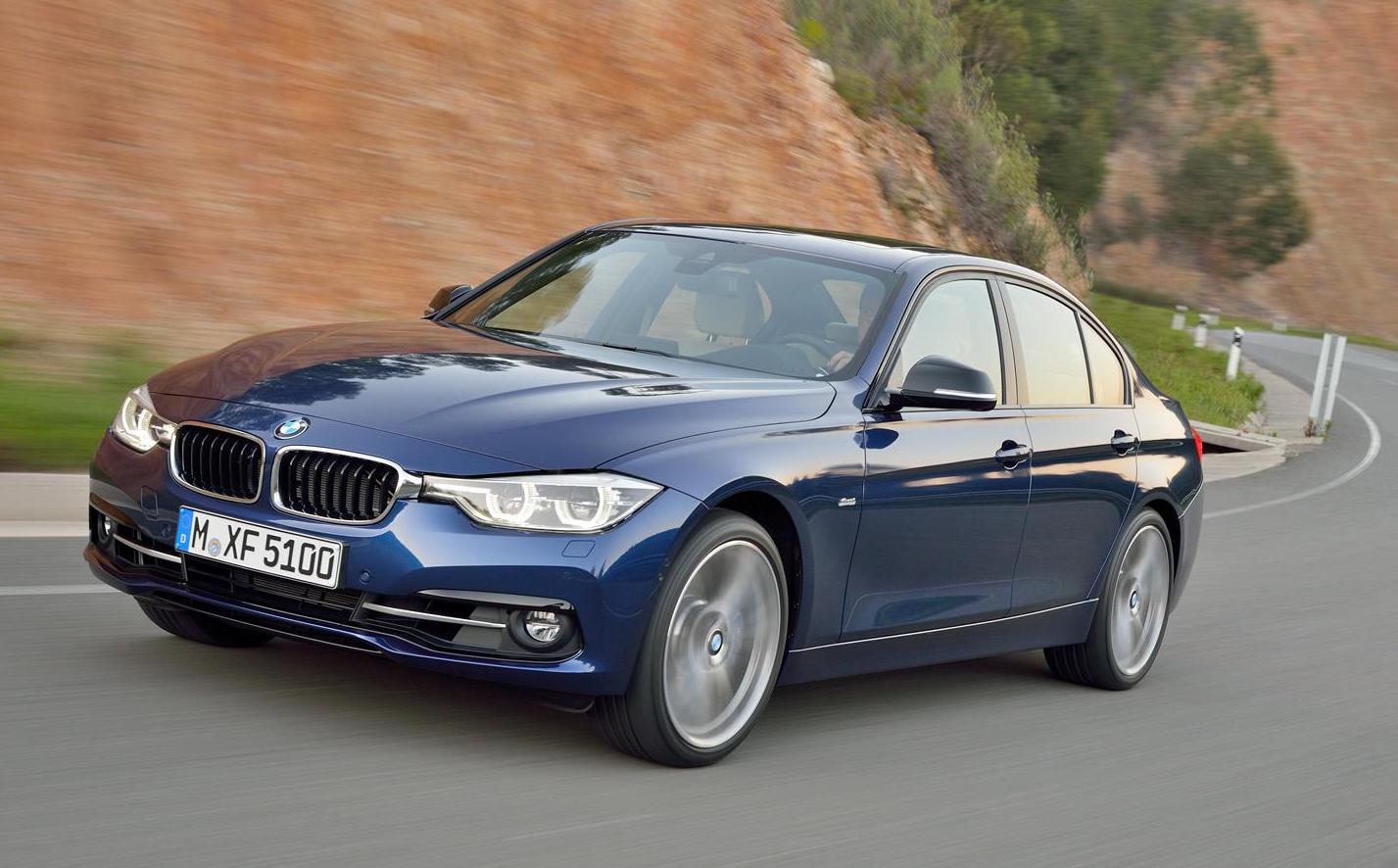}\\		
		%Entity-Attribute & \texttt{Sunflower} is \texttt{Yellow}. & %\includegraphics[width=0.1\textwidth]{sun-flower-2713132_1280.jpg}\\
		\bottomrule
	\end{tabular}
	\begin{tabular}{ccc}
		\toprule
		relation type & example & images \\
		\midrule
		Scene-Entity & \tabincell{c}{\texttt{Ferris wheel} is found\\ in \texttt{Amusement park}.} & \includegraphics[width=0.1\textwidth]{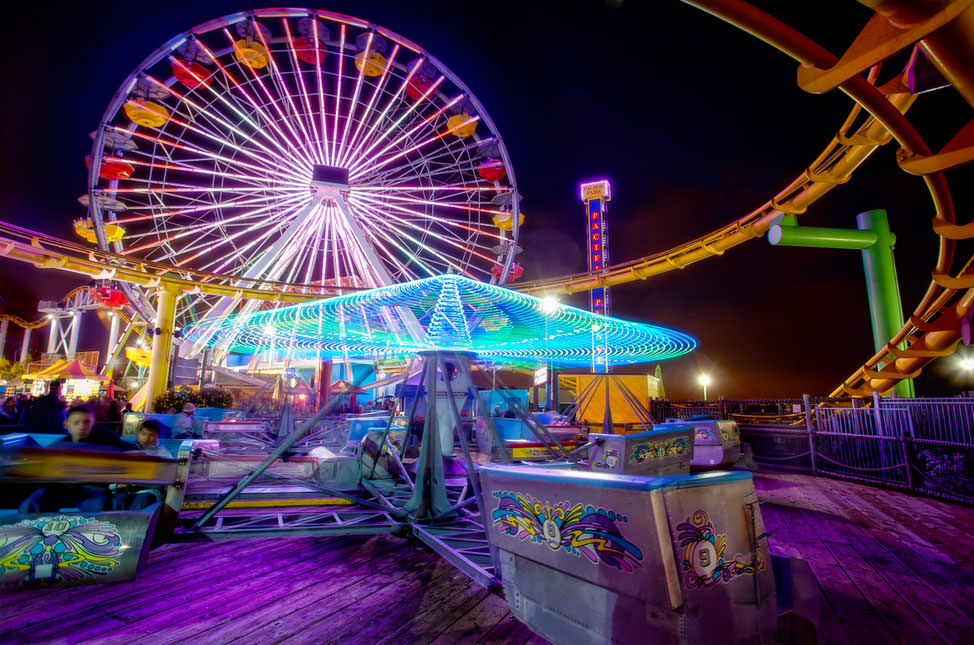}\\
		Scene-Attribute & \texttt{Alleys} are \texttt{Narrow}. &  \includegraphics[width=0.1\textwidth]{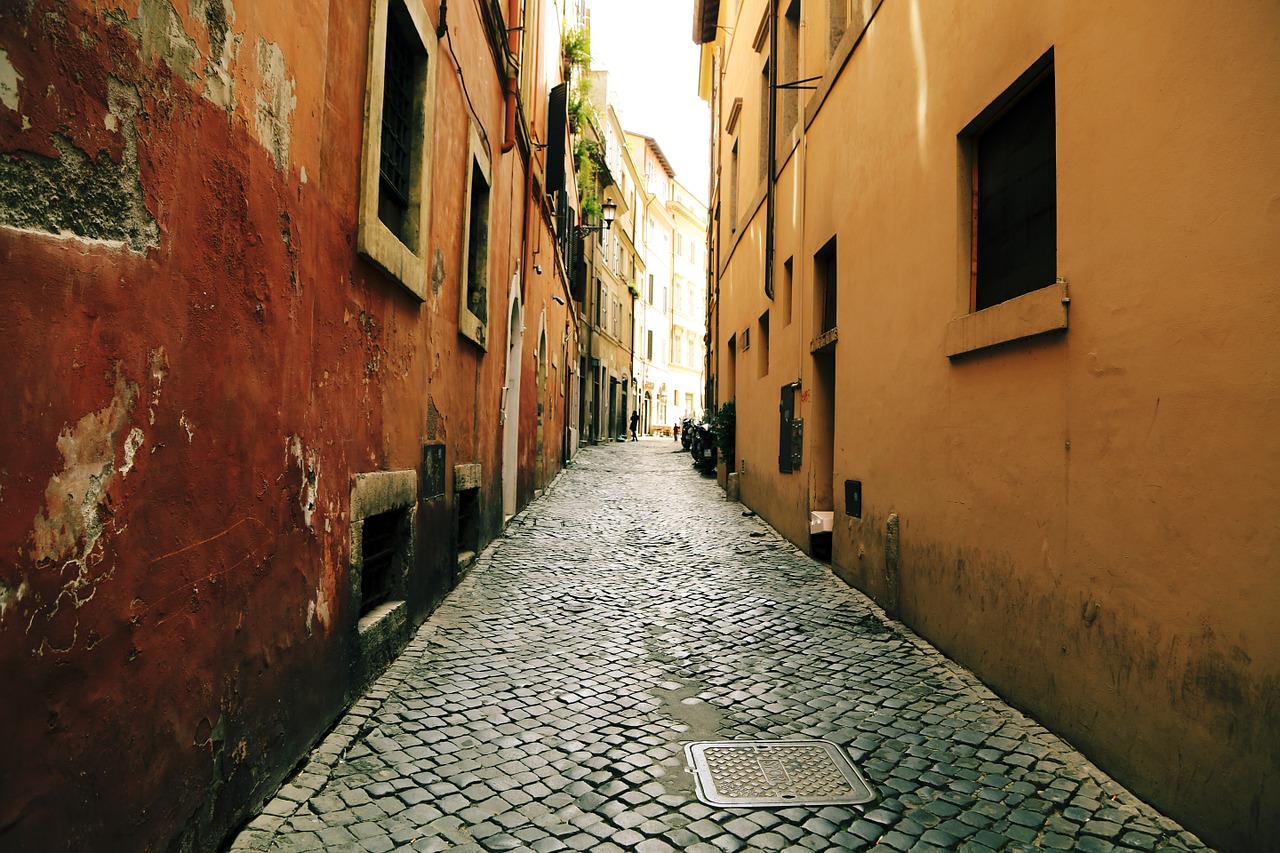}\\
		\bottomrule
	\end{tabular}
	\caption{Examples of visual relations detected in NEIL~\cite{chen2013neil}}%~\cite{chen2013neil} }
	% 这个表里的四个图，用不用画bounding boxes？因为原文中提到了，是根据bbox的位置关系确定的，主要用于concept-concept、entity-concept。scene不用两个bbox,只需要一个，scene分类即可
	\label{tab:relation}
	\vspace{-1.0em}
\end{table*}

2) {\it Statistic-based General Relation Extraction}. The statistic-based methods encode features such as visual, spatial, and statistics of the detected objects into distributed vectors and predict the relation between the given objects by a classification model. Unlike rule-based methods, statistic-based methods can detect all relations in the training set.%Compared to rule-based methods, statistic-based methods are able to detect all relations that have appeared in the training set. 

Some work has proved that predicting the predicates rely heavily on the categories of subjects and objects, but subjects and objects are not dependent on predicates, and there is also no dependency between subjects and objects \cite{zellers2018neural}. For example, in triple (\texttt{Person, ride, Elephant}), \texttt{Person} and \texttt{Elephant} indicate that the relation might be \texttt{ride} rather than \texttt{wear}. Thus to utilize the dependency, \cite{lu2016visual, dai2017detecting, zellers2018neural} add language priors of language models into the statistic model by objects' labels and \cite{zhang2017visual} set a stricter constraint that the hidden layer representation of a triple should satisfy \emph{subject} + \emph{predicate} $\approx$ \emph{object}. It is embarrassing that the language model improves much, but the visual information contributes little~\cite{zellers2018neural}.

Detected objects and relations in an image could be represented as a graph. The graph structure enables the edges to get more messages from other nodes and edges to classify the relation with higher accuracy. For example, \cite{xu2017scene} represents objects and relations as two complementary sub-graphs, where nodes are iteratively updated according to the values of the surrounding edges and vice versa. \cite{yang2018graph} used GCN to learn the context of objects and edges.
Unfortunately, the recall@50 of triple detection in current visual detection models is still less than 23\%, although the recall@50 of predicate detection has been up to 85.64\%\cite{yu2017visual} on the visual relation detection benchmarks.

3) {\it Long-tail and Fine-grained Relation Extraction}. 
It is challenging for statistic-based methods to detect long-tail relations. Frequent relations are more likely to be predicted due to the bias of sample distribution in the training sets. 
%
%For example, models tend to predict \texttt{(Person, play, Violin)} instead \texttt{(Person, hold, Violin)}. 
%
Much work focuses on eliminating the effect of unbalanced samples in the training sets by metric learning ~\cite{tang2019learning, zhang2019graphical}, transfer learning~\cite{wang2020memory}, few-shot learning~\cite{wang2020one} and contrastive learning~\cite{zhang2019large}, which are still limited to the feature fusion of hidden layers.

%\zxr{\sout{
%Although statistics-based methods are able to detect general relations, it is challenging to detect long-tail relations. The reason is that the biased datasets make it more possible to predict the relations with a large number of samples. In order to eliminate the effect of unbalanced samples in the training sets,~\cite{tang2019learning} proposed a new unbiased metric (Mean Recall@K) to average the recall of all types of relations instead of all samples and avoid the neglect of relations with only a few samples. There is much other work focusing on detecting relations with few samples by transfer learning~\cite{wang2020memory}, few-shot learning~\cite{wang2020one} and contrastive learning~\cite{zhang2019large} still being limited to the feature fusion of hidden layers.
%}}

Fine-grained relation is a kind of long-tail relation. Existing studies on long-tail relation problems from the perspective of feature fusion fail to distinguish fine-grained relations well. For example, models tend to predict \texttt{on} instead of fine-grained relation \texttt{sit on/walk on/lay on}. For more informative unbiased predictions, \cite{tang2020unbiased} uses counterfactual causation instead of conventional likelihood to remove the effect of context bias.
% 
%Differently, \cite{baumgartner2020towards} orders relations into a hierarchy from specific to generic ones, and trains a classifier for each node in the hierarchy to decide whether a detected triple belongs to this relation or any of its sub-relations.
Differently, \cite{baumgartner2020towards} orders relations in a hierarchy, from specific ones at the bottom to generic ones towards the top. It trains a classifier for each relation, classifying a detected triple into two types: whether it belongs to a certain relation or its sub-relations in the hierarchy.

\eat{
It is more difficult to detect complex and fine-grained relations such as human-object interaction and action detection.  Because the pose of a person is determined by many components of the body. For example, there is a slight difference between images of \texttt{(Person, play, Violin)} and \texttt{(Person, hold, Violin)}. In the first example, the violin is usually placed on a person's neck, with the left hand pressing the strings and the right hand holding the bow, and in the second example, the violin may be absolutely different poses. In early studies, the action is defined as a series of poses of different parts of the body, and the discriminative features are mined by heuristics approaches~\cite{yao2010grouplet, yao2010modeling, maji2011action}. In the current statistics-based detection, the discriminative feature are filtered by stricter contrastive loss functions~\cite{zhang2019graphical}, which is obviously still too coarse.}

\vspace{5pt} \noindent{\bf OPPORTUNITIES}. Despite much existing work, there still leaves many challenging issues unsolved. For instance: 1) {\it Visual Knowledge Relation Judgement}. Many visual triples extracted from images only describe the scene of the image, which are unqualified to be taken as visual knowledge since they are not widely accepted facts. The challenges (also opportunities) lie in how we recognize the triples of visual knowledge from the triples of scene information.
2) {\it Relation Detection based on Reasoning}. Existing relation detection methods predict the relations by a hidden unified representation fusing visual features and language priors. We cannot explicitly describe the basis of prediction. \cite{li2019hake} builds a human action dataset to help predict an action by body part states. For example, if there is a person and a football in an image and \texttt{(Head, look at, Sth)} \texttt{(Arm, swing, -)} \texttt{(Foot, kick, Sth)} are meanwhile satisfied, the action will be judged as \texttt{(Person, kick, Football)}. Unfortunately, this dataset is built manually. We need to summarize the chain of reasoning for relation detection automatically.

\subsubsection{Visual Event Extraction}\label{section:sec3.1.3}

An event includes a trigger and several arguments with their argument roles.
%An event is usually defined as the dynamic interaction among arguments~\cite{zhang2017improving}, including a trigger and several arguments with their argument roles. 
A trigger is a verb or a noun indicating the occurrence of an event. An argument role is a %semantic 
relation between an event and an argument, %, such as \texttt{time}, \texttt{person} and \texttt{place}, 
and the arguments are entity mentions, concepts or attribute values.
%A trigger is a verb or a noun indicating the occurrence of an event. An argument role is the semantic relation between an event and an argument, such as \texttt{time}, \texttt{person} and \texttt{place}, and the arguments are entity mentions, concepts or attribute values.
%
%For example, in the event \emph{Henry was injured by Mike}, the trigger is \emph{injured}, and an argument is \textt{Henry} who is the subject of the event, while the other is \textt{Mike} who is the object of the event.
%
\eat{
The traditional event extraction task aims to predict the event types by triggers and then extract their arguments according to the pre-defined schema of the event.
Similarly, the visual event extraction can be also divided into two sub-tasks: 1) to predict the visual event types; and 2) to locate and extract objects in source images or videos as visual arguments~\cite{zhang2017improving, chen2021joint, li2020cross, wen2021resin}.
}
The visual event extraction can also be divided into two sub-tasks: 1) to predict the visual event types; and 2) to locate and extract objects in source images or videos as visual arguments~\cite{zhang2017improving, chen2021joint, li2020cross, wen2021resin}. 
This task is different from the situation recognition task~\cite{xiong2015recognize,yatskar2016situation,pratt2020grounded} in CV, which aims to recognize a visual event rather than locating and extracting its visual arguments. Schemas defined in datasets of situation recognition tasks, such as SituNet~\cite{yatskar2016situation} and SWiG~\cite{pratt2020grounded}, can be used to train models in this task. 
%For example, the event \texttt{Clipping} have argument roles like \texttt{Agent}, \texttt{Source}, \texttt{Tool}, \texttt{Item}, \texttt{Place} and in an image of \emph{clipping a sheep’s wool} they are respectively \texttt{Man}, \texttt{Sheep}, \texttt{Shears}, \texttt{Wool}, \texttt{Field}.

\vspace{5pt} \noindent{\bf CHALLENGES}. %There are several challenges with the task: 1) Visual event extraction requires pre-defined schema for different event types, but a large number of visual events have not been defined by experts. How to mine visual pattern as event schema automatically?
The task has several challenges: 1) Visual event extraction requires pre-defined schemas for different event types, but there are a large number of visual events that experts have not defined. How to mine visual patterns as event schemas automatically?
2) How to extract visual arguments of a visual event from images or videos?
%
%3) There are also some events which are made up of a serial of temporal sub-events in videos. How we extract these temporal sub-events from the videos?

%Existing studies on visual event are mainly extracted from images or video key frames, in which there are only one event to be extracted. However, an event in a long video may contain multiple steps, which existing in most videos. How to extract a high-level event being made up of a serial of temporal sub-events in videos?

\vspace{5pt} \noindent{\bf PROGRESSES}. The existing work on visual event extraction mainly focuses on two aspects: 1) visual event schema mining, which detects and labels the most relevant visual entities (or concepts) as a new schema; 2) visual event arguments extraction, which extracts argument role regions from visual data according to the event schema.

1) {\it Visual Event Schema Mining}. 
\eat{
In CV community, there is a similar task called situation recognition task, the schema of which includes the main activity (event type), the participants (arguments), the roles theses participants play in the activity (argument role). Some datasets, such as SituNet~\cite{yatskar2016situation} and SWiG~\cite{pratt2020grounded}, have defined a lot of visual event schemas. For example, the event \texttt{Clipping} have argument roles like \texttt{Agent}, \texttt{Source}, \texttt{Tool}, \texttt{Item}, \texttt{Place} and in an image of \emph{clipping a sheep’s wool} they are respectively \texttt{Man}, \texttt{Sheep}, \texttt{Shears}, \texttt{Wool}, \texttt{Field}. The task mainly aims to recognize an visual event rather than locating and extracting its visual arguments. The visual arguments detection relies on the co-occurrence~\cite{yatskar2016situation, pratt2020grounded} and relative positions~\cite{xiong2015recognize} of detected objects. 
}
In large-scale visual event extraction, such as news, the visual schemas of many events have not yet been manually defined, which requires much experts' work. Large numbers of image-caption pairs from the web make it possible to mine and label the visual pattern for event schemas. 
%
%However, in large-scale visual event extraction, such as news, the visual schemas of many events have not yet been manually defined, which requires a lot of experts' work. A large number of image-caption pairs from Web make it possible to mine and label the visual pattern for event schemas. % and label the extracted visual pattern. 
Thus this task is reduced to finding a frequent itemset of visual patterns which indicate the correct event type from the images of a given event. %The collection of all images of a given event can be retrieved from the collection of image-caption pairs with the triggers of the event as queries.
The collection of images of an event can be retrieved from the image-caption pairs with the event's triggers as queries.
%The collection of images of an event can be retrieved from the image-caption pairs with the triggers of the event as queries.
%By weakly supervision of images and corresponding captions, 
Words or phrases in captions label the candidate image patches through visual grounding. Heuristic approaches (e.g., the Apriori algorithm) can be utilized to mine frequent visual %itemsets of 
%Through visual grounding methods, the candidate image patches are labeled by words or phrases in captions. Heuristic approaches (e.g., Apriori algorithm) can be utilized to mine frequent visual %itemsets of 
image patches to find association rules for predicting the event type by visual patterns~\cite{li2016event, zhang2017improving}. 

Mining and labeling methods can correct wrong arguments or add missing ones in manually defined visual event schemas. For example, an ontology expert may consider \texttt{Explosion} and \texttt{Weapon} as important items in the schema of event \texttt{Attack}, but in some news corpus, these concepts %items 
are not discovered and \texttt{Smoke} and \texttt{Police} appears much more frequently, which is not expected in advance~\cite{li2016event}.

2) {\it Visual Event Arguments Extraction}. 
This task aims to extract a group of visual objects with the constraint of relations. The event types are classified according to the global features of images, and the event arguments are extracted as the most sensitive region to the event type by object recognition or visual grounding.
%The visual event arguments extraction is actually a task of extracting a group of visual objects with the constrain of relations. 
%Visual arguments are able to be labeled by fully supervised methods like object recognition or weakly supervised methods like visual grounding. 
%According the two sub-tasks of visual event extraction, the event types are classified by the global features of events' images and the event arguments are extracted as the most sensitive local region to the event type. 
%
The quality of the two sub-tasks on a large corpus is acceptable. In MMEKG~\cite{ma2022mmekg}, the instance-level evaluation has a precision score of about 64\% on visual events and cross-modal triples.

In addition, the relations in visual and text arguments should also be aligned to ensure that the relations among visual objects are consistent with the relations in text. \cite{li2020cross} aligns the situation graph~\cite{yatskar2016situation} extracted from the image and the abstract meaning representation graph (AMR graph)~\cite{banarescu2013abstract} extracted from the caption of an event in terms of the semantics and categories of cross-modal arguments. Many constraints on semantic, event type, event argument role and the consistency between modalities are also added into joint extraction~\cite{li2020cross, wen2021resin}.

\eat{
However, during weakly supervised methods, we are unsure whether the relations among extracted visual objects are consistent with the relations in text. Thus, the relations in visual and text arguments should also be aligned.
~\cite{li2020cross} aligns the situation graph~\cite{yatskar2016situation} extracted from an image of an event to the abstract meaning representation graph (AMR graph)~\cite{banarescu2013abstract} representing the semantic structure of the caption of this event in terms of the semantic and categories of cross-modal arguments. 
Many studies~\cite{li2020cross, wen2021resin} add constraints on semantic, event type, event argument role and the consistency between modalities in joint extraction.
%Many potential constraints on semantic, event type, event argument role and the consistency between modalities are also added into joint extraction~\cite{li2020cross, wen2021resin}. % to identify the consistency~\cite{li2020cross, wen2021resin}. 
%
%Many potential constrains on semantic, event type, event argument role and the consistency of visual and text information are also added into joint extraction to identify the consistency~\cite{li2020cross, wen2021resin}.
}

Videos are more suitable for event extraction than images because the temporal bounding box of an event may be across the video, and all arguments may not appear in a single frame. ~\cite{chen2021joint} simplifies this task and extracts arguments from three keyframes derived from short video segments including only one event, and the keyframes are the most matching ones to the captions of the videos. 
%Compared to images, videos are more suitable for event extraction, because the temporal bounding box of an event may across the video and all arguments may not be shown in a single frame. To simplify the task,~\cite{chen2021joint} extracted arguments from three key frames derived from short video segments including only one event, and the key frames are the most matching ones to the captions of the videos. 

\vspace{5pt} \noindent{\bf OPPORTUNITIES}. The research on this task is still in an early stage, and many problems are still worth exploring.
%The research on this task is still in an early stage, and there are still many problems worth exploring. 
For instance: 1) The extraction of sequential events from a long video containing multiple events has not yet been addressed. 2) {\it Video Event Extraction with multiple Sub-events.} For example, the event \texttt{Making Coffee} is divided into a sequence of steps, such as \texttt{Cleaning coffee machine} $\rightarrow$ \texttt{Pour in the coffee beans} $\rightarrow$ \texttt{Turn on the coffee machine} and each step can be also considered as an event. The sequential steps need to be extracted and listed by the timeline of the steps, which are difficult to be solved by current methods. % 

%Some events are a progress including multiple sub-events. For example, the event \texttt{Making Coffee} is divided into a sequence of steps, such as \texttt{Cleaning coffee machine} $\rightarrow$ \texttt{Pour in the coffee beans} $\rightarrow$ \texttt{Turn on the coffee machine} and each step can be also considered as an event. The sequential steps need to be extracted and listed by the timeline of the steps, which are difficult to be solved by current methods. First, we have to detect and locate the temporal boundaries of each steps in a long video. Second, the textual information of a video can be obtained from the sounds and captions. The sequential steps in the caption and video should be aligned. Finally, how to determine a high-level event based on the schema in which arguments are multiple small-scale sub-events is also a difficult issue.

\subsection{From Symbols to Images: Symbol Grounding}\label{sec:symbolGrounding}

Symbol grounding refers to the process of finding proper multi-modal data items such as images to describe a symbol knowledge in a given KG, such as an entity, a concept or a relational triple.
Some popular MMKGs constructed in the symbol grounding way are listed in Table~\ref{tab:mainstream_mmkg}(b).

In the rest of this subsection, we cover the process of grounding symbols to images in several fractionized tasks: {\it Entity Grounding} (Sec.~\ref{section:sec3.2.1}), {\it Concept Grounding} (Sec.~\ref{section:sec3.2.2}) and {\it Relation Grounding} (Sec.~\ref{section:sec3.2.3}).

%We discuss the challenges, progresses and opportunities of each symbol grounding task in Sec.~\ref{section:sec3.3.1}, Sec.~\ref{section:sec3.3.2} and Sec.~\ref{section:sec3.3.3} respectively.

% 根据三个任务拆分

\subsubsection{Entity Grounding}
\label{section:sec3.2.1}
Entity grounding aims to ground entities in KGs to their corresponding multi-modal data such as images, videos and audios~\cite{harnad1990symbol}. The existing work mainly focuses on grounding entities to their corresponding images.

\vspace{5pt} \noindent{\bf CHALLENGES}. The main challenges of grounding entities to images are the following: 1) How to find enough images with high quality for entities at a low cost? 2) How to select the images that best match an entity from much noise?

\vspace{5pt} \noindent{\bf PROGRESSES}. There are two major sources to find images for entities: (1) from \emph{online encyclopedia} (such as Wikipedia), or (2) from the Internet through \emph{Web search engines}.

1) {\em From Online Encyclopedia}. In Wikipedia, an article usually describes an entity with images. Wikipedia and DBpedia provide many facilities (such as Wikimedia Commons%\footnote{a multi-media dataset linking to Wikipedia articles, https://wikimediafoundation.org/our-work/commons/}
) to help build the connection between an entity in DBpedia and corresponding images or data in other modalities in Wikipedia. It is easy for researchers to use an online encyclopedia like Wikipedia to build the first version of a large-scale MMKG. 

However, the encyclopedia-based approach has several major disadvantages: 1) First, not all entities are attached to many high-quality images in an online encyclopedia. We investigate that the average number of images per entity in Wikipedia is only 0.83. 
Second, many images of entities in Wikipedia are only indirectly related to that entity but can not accurately represent that entity. For example, there are several images of \emph{animals}, \emph{buildings}, \emph{plaques}, \emph{carvings} in images of \texttt{Beijing Zoo} in Wikipedia. 
Third, the images of the non-visualizable entity may bring mistakes. For example, in the Wikipedia article of \texttt{Gaussian Progress}, there is an image of \emph{Gaussian processes with different prior conditions}, which should not be mapped to any image. 
Finally, the coverage of MMKG built from Wikipedia alone still needs to be improved. English Wikipedia has 6 million entities (articles), which is the upper bound of the capacity of the MMKG harvested from English Wikipedia. According to our investigation, 79.35\% of Wikipedia articles in English have no corresponding images, and only 6.7\% of them have at least 3 images.

\eat{
However, not all entities are attached to many high-quality images in an online encyclopedia. We investigate that the average number of images per entity in Wikipedia is only 0.83.%~\cite{ferrada2017imgpedia}.
In addition, many images of entities in Wikipedia are only indirectly related to that entity but can not accurately represent that entity. For example, there are several images of \emph{animals}, \emph{buildings}, \emph{plaques}, \emph{carvings} in images of \texttt{Beijing Zoo} in Wikipedia. 
Also, the images of the non-visualizable entity may bring mistakes. For example, in the Wikipedia article of \texttt{Gaussian Progress}, there is an image of \emph{Gaussian processes with different prior conditions}, which should not be mapped to any image. 
Finally, the coverage of MMKG built from Wikipedia alone still needs to be improved. English Wikipedia has 6 million entities (articles), which is the upper bound of the capacity of the MMKG harvested from English Wikipedia. According to our investigation, 79.35\% of Wikipedia articles in English have no corresponding images, and only 6.7\% of them have more than 3 images (including 3).
}

2) {\em From Search Engines}. 
Search engine based solutions are proposed to improve the coverage of an MMKG. We can easily find images from the search results of a commercial search engine by specifying entity names as queries, where the top-ranked image is more likely to be the correct image of the searched entity. Thus we can select these images for the entity to be searched. Compared to the Wikipedia based approach, the coverage of MMKG is significantly improved in the search engine based approach. 
%To improve the coverage of an MMKG, search engine based solutions are proposed. We can easily find images from the search result of commercial search engine by specifying entity names as queries. The top ranked result image in general has a large possibility to be the correct image of the entity to be searched. Thus we can select these images for the entity to be searched. Compared to the Wikipedia based approach, the coverage of MMKG is significantly improved in search engine based approach. 

However, the search engine based approach is easy to introduce noisy images into MMKGs. It is well recognized that the search engine results might be noisy. Another reason is that it is not trivial to specify the search keywords. For example, the search query ``\emph{Bank}'' is not good enough to find the image for \texttt{Commercial Bank}, since it also incurs the images of \texttt{River Bank}. Hence, many efforts have been made to clean candidate images. The query words are usually extended for disambiguation by adding parent synsets~\cite{deng2009imagenet} or entity types~\cite{liu2019mmkg}. Diversity is also a non-negligible issue when selecting the best images for the entity. An image diversity retrieval model is trained to remove similar redundant images so that the grounded images are as diverse as possible~\cite{wang2020richpedia}. 

Compared to the encyclopedias-based approaches, search engine based approaches are better in coverage but worse in quality. The two approaches are often used together since in most cases the knowledge acquired by these two approaches complements each other~\cite{wang2020richpedia}. For example, the coverage of MMKG harvested from Wikipedia can be improved by collecting more images for each entity from search engines~\cite{wang2020richpedia}.

\begin{figure}
	\centering
	\begin{tabular}{@{}c@{}}
		\includegraphics[width=1\linewidth]{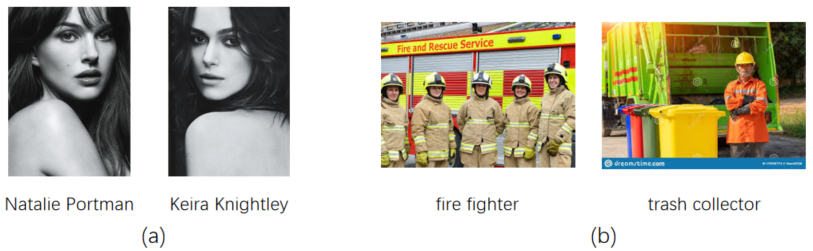} \\% [\abovecaptionskip]
	\end{tabular}
	\caption{Examples that can hardly be distinguished by visual entity extraction methods. (a) Similar visual entities: \texttt{Natalie Portman} and \texttt{Keira Knightley}; (b) Similar visual concepts: \texttt{fire fighter} and \texttt{trash collector}.}
	\label{fig:similar_entities}
	\vspace{-1.0em}
\end{figure}

Due to the decoupling of entities and their visual features, an MMKG constructed with encyclopedias or search engines can distinguish visually similar entities, as shown in Fig.~\ref{fig:similar_entities}. Entity grounding methods make it possible to build a domain-oriented fine-grained MMKG (e.g., a movie/product/military MMKG).

%
%3) {\it From a Visual-Language Corpus.}
%如果场景中包含大量弱监督的图文对，那么也可以考虑训练一个检索系统来辅助实体接地。搜索引擎方法是一种特殊的无需预训练的检索系统，主要依靠标题和query的匹配来召回图片。但是，如果实体是比较复杂的不常见的短语组合，那么这些短语不方便用搜索引擎检索的到。我们可以利用领域的大量语料训练检索模型，来获得对应短语的图片。

%
%The search engine based approach is a special case of retrieval systems which does not need  training, but only retrieves entities' images only based on the similarity between the query and captions. %
%However, if we want to search by both textual semantics in captions and visual semantics in images when searching, the results of search engines may be not enough.
%
%Training a retrieval model is an alternative that is suitable for the construction of domain-oriented MMKG from a large corpus of text-image pairs. \cite{xu2021alime} trains a cross-modal matching model based on images from E-commerce products' detail pages and OCR text in these images, and convert the entity grounding task to a text-image retrieval task.

\begin{figure}[t]
	\centering
	\small
	\includegraphics[width=0.5\textwidth]{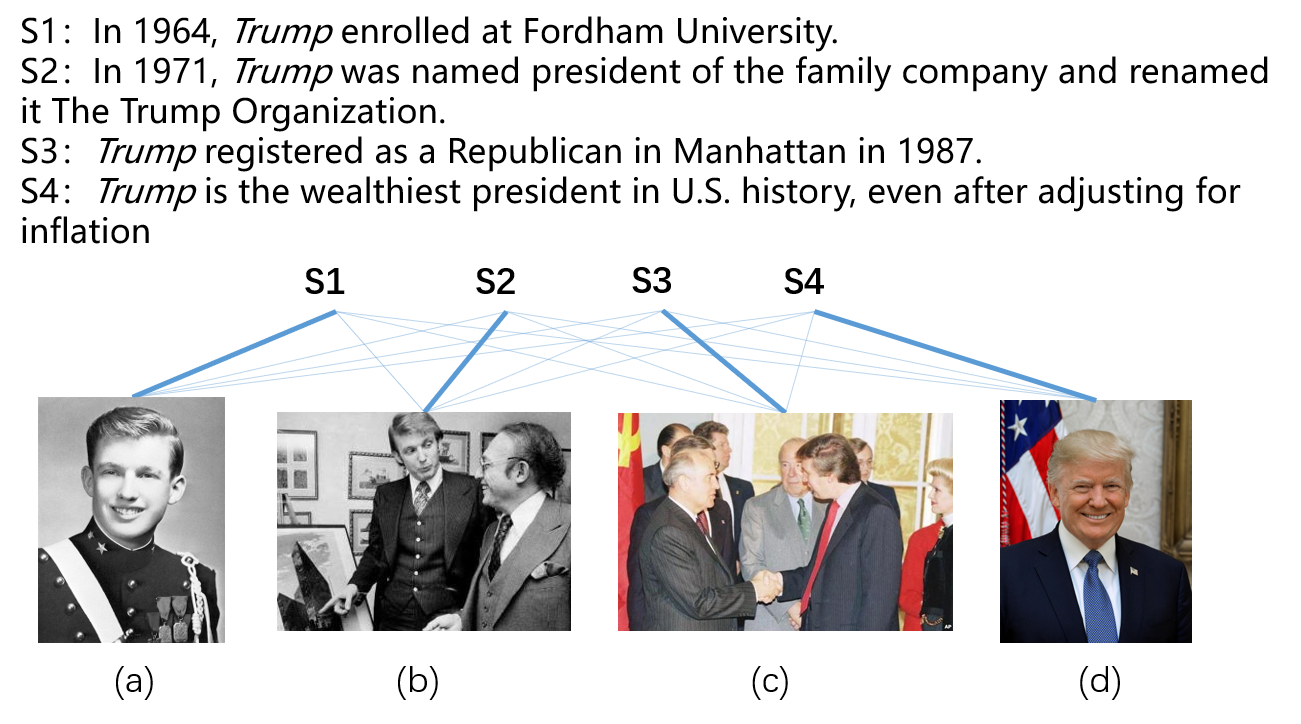}
	\caption{Take \texttt{Trump} as an example to illustrate that an entity needs different images to express its different aspects (Trump as (a) a young student, (b) a businessman, (c) a politician, or (d) the president of the USA) in different contexts}% There are four examples of multiple grounding of \texttt{Trump}. Images are respectively photos of \sout{Sentences and Images about \texttt{Trump}.} \sout{(a) Photos of Trump's early life} (a) Trump as a youth}, (b) Trump as a businessman, (c) Trump as a politician, and (d) Trump as the president of the USA. Obviously, (a) is more related to {S1}, (b) is more related to {S2}, (c) is more related to {S3} and (d) is more related to {S4}.}
	\label{fig:multiGrouding1}
	\vspace{-1.0em}
\end{figure}

\vspace{5pt} \noindent{\bf OPPORTUNITIES}. There are many unsolved problems in this direction.
1) Entities are grounded into several images, each of which is only an aspect of the entity. For example, the image collection of a person may be images of different ages, life photos, event photos, single photos and family photos. How do we determine the most typical subset?%How to determine the most typical subset?
% to do: 这个(2)和原来的(3)【现在的(2)】重复了, 只留一个即可。不错， (3)的例子是否需要替换为国家？？？？
%2) Each image of an entity has a matching context, and perhaps the context of images in a same collection are diverse. For example, \texttt{the United States} might be grounded into \texttt{a map of the United States}, \texttt{the American flag}, \texttt{Martin Luther King} and \texttt{George Washington}. How do you determine the picture in a particular context? how to find the right image for the entity given a specific context?
%
%3) 
2) Real-world entities are multi-faceted, and it is desirable to associate an entity with multiple images in different contexts. The demand motivates us to propose a new task \emph{multiple grounding} that selects the most related images from the entity given a specific context.
%2) Real-world entities are multi-faceted, and it is desirable to associate an entity with multiple images in different contexts. This motivates us to propose a new task \emph{multiple grounding} which selects the most related images from the entity given a specific context. %This problem could be formulated as $\argmax\limits_{i}{P(i|c)}, i \in I_e$, where $e$ denotes an entity and $c$ denotes a specific context. $I_e$ denotes an image pool of entity $e$ and $P(i|c)$ is the probability that image $i$ is an appropriate image for entity $e$ conditioned on the given context $c$.
%
For example, \texttt{Donald Trump} %, the 45th president of the United States, 
has a lot of different images that can be collected from the web. But as shown in Figure~\ref{fig:multiGrouding1},  any single image is not appropriate for all the different contexts. Thus, \texttt{Trump} should be multi-grounded when constructing the knowledge graph.
%
%However, mapping different aspects of entities to the most related images in different contexts is not a trivial task. First, the image pool of an entity is difficult to build, because the completeness of the image pool cannot be guaranteed and it is easy to miss some related images for certain contexts. Second, disambiguating images for the entity for a specific context is challenging, because the context is usually noisy and contains sparse information, and more background information is needed to guide the acquisition of semantic information. Finally, as a new task, the lack of labeled data is a big issue .%and it is not trivial to annotate data.
%
3) If there is an objective domain corpus containing a large amount of texts with attached images, we may convert the entity grounding task into a text-image retrieval task, such as the work done on the E-commerce domain~\cite{xu2021alime}.
\eat{
3) If there is a large visual-language corpus containing a large amount of texts with attached images of the objective domain, we may convert the entity grounding task into a text-image retrieval task, such as the work done on the E-commerce domain~\cite{xu2021alime}.
}

%\textbf{Domain Entity Grouding} Addition to visual related cross-modal KG, MMKG consist of other modal data are also promising, such as chemical formula in chemical KG, which are natural multi-modal. Though it need specific knowledge, it reason based on its shape and structure which is multi-modal.

\subsubsection{Concept Grounding}\label{section:sec3.2.2}
Concept grounding aims to find representative, discriminative and diverse images for visual concepts. 

\noindent{\bf CHALLENGES}
Although some visually unified concepts (such as \texttt{man}, \texttt{woman}, \texttt{truck} and \texttt{dog}) can also be grounded to images with the entity grounding methods introduced in Sec.~\ref{section:sec3.2.1}, the symbol grounding to the other concepts faces new challenges: 
1) Not all the concepts could be adequately visualized. For example, \texttt{irreligionist} cannot be grounded to a specific image. How to distinguish visualizable concepts from non-visualizable ones?
2) How to find representative images for a visualizable concept from a group of relevant images? Note that the images of a visualizable concept might be very diverse. For example, when it comes to \texttt{Princess}, people often think of several diverse images: \emph{Disney princesses, ancient princesses in historical movies or modern princesses in the news}. Therefore, we have to consider the diversity of images..
    
\vspace{5pt} \noindent{\bf PROGRESSES}. In response to the above challenges, related studies are divided into three tasks: visualization concept judgment, representative image selection and image diversification.

1) {\em Visualization Concept Judgment}. The task aims to automatically judge visualizable concepts and is a new task to be solved. \cite{yang2020towards} discovers that only 12.8\% of the synsets of \texttt{Person} subtree have well-accepted imageability (i.e., the score is greater or equal to 4 and the total score is 5), and many of the rest synsets have no corresponding visual descriptions. For example,~\texttt{Rock star} is imageable, and~\texttt{Job candidate} is non-imageable. So what are the criteria for recognizing visual concepts? The manual annotation in~\cite{yang2020towards} is unpractical in constructing a large-scale MMKG.

In order to automatically judge visual concepts, there has been much effort based on syntax and semantics.~\cite{torralba200880} thinks that abstract nouns concepts are non-visualizable so that TinyImage dataset~\cite{torralba200880} removes all hyponyms in the subtree of~\texttt{Abstraction} in WordNet and only collects images for non-abstract noun concepts. However, these methods are not very accurate. For example, \texttt{Anger} or \texttt{Happiness} can be grounded in an image of a person who feels angry or happy. Since the images come from the web, it is possible to use search engine hits to judge visual concepts. For example, a word might be visualizable if the number of Google image hits is larger than that of Google web hits~\cite{yan2017graph}.
\cite{jiang2022visualizable} assumes that if images of a concept from Google are similar (with a small variance), this concept is more likely to be visualizable. This assumption may lead to a low recall, so it is used to correct the false negative predictions (non-visualizable) of classifiers.

%\zxr{ \sout{In addition, some characterises of high-qualified images of visual concepts could be used to recognize visual concepts, such as representiveness and discriminativeness. \cite{divvala2014learning} believes the foreground of representative images are similar and are easily to be separated from background and have a small inter-class variance. Thus in turn \cite{divvala2014learning} trains a classifier to select the concepts whose images collection have these characterises.}}

2) {\em Representative Image Selection}. Based on the methods of Sec.~\ref{section:sec3.2.1}, we get a collection of images for each visual concept. This section focuses on selecting visually representative and discriminative images in the collection. 
%In this section, we focus on selecting visually representative and discriminative images in the collection. 

The task %essentially 
aims to re-rank the images according to their representativeness. The representative scores of images derive from results of cluster-based methods, such as K-means, spectral clustering, etc. The smaller the variance within a cluster, the higher the scores of images in the cluster. After re-ranking the representative scores of images, the top may be representative images. In addition, the expected images are also constrained by rules to distinguish different clusters. For example,~\cite{qimeibin2014representative} adds a new metric to rank images together with similarity within clusters, which is the ratio of inter-class distances and intra-class distances, and the bigger a ratio, the more discriminative the image is. 
%

%In addition to traditional clustering methods, graph clustering methods are also a choice. In this graph, images are nodes and image similarity are edges. Based on graph cluster information, more graph-based methods (e.g. Markov Random Walk) are convenient to follow up. 

The captions and tags of images from search engines could also be utilized to evaluate the representativeness and discrimination of images at the level of semantics. Captions and tags provide semantic information that images do not have. For example, a photo of \emph{Icelandic landscapes} and a photo of \emph{British landscapes} may look similar, but text tags can help us distinguish their differences in concepts. In~\cite{yu2014joint, yan2017graph, wang2021social}, tags are clustered based on semantic features and images are reassigned into each cluster according to their tags' semantic clusters.

3) {\em Image Diversification}. The task requires that images in which concepts are grounded should balance diversity and relevance. The images should also be re-ranked after clustering, but the difference from representative image selection is that we want to show the results of as many clusters as possible. Specifically, in each selection step, images from unselected clusters are preferred to be selected.

There are two types of scores for ranking the priority of selection: diversity scores and relevance scores, where diversity scores evaluate the topics of images and relevance scores penalize the difference of images to avoid semantic drift. For fusing the two conflicting scores, \cite{van2009visual, wang2010towards} use Max-Min methods to choose candidates: assign a higher score to images that are not similar to the selected set, and choose the dissimilar one with the highest score among the remaining similar ones.
\cite{jiangxueyao2022entity} mines topics (e.g., \texttt{View, Flag, Map}) from image captions of popular entities (e.g., \texttt{Greenland}) to expand queries of long-tail entities of the same type (e.g., \texttt{Country}) during image retrieval. Then images of long-tail entities are filtered by local outlier factors based on the distribution of similar popular entities' images. Diversity is achieved by pattern mining, and relevance is achieved by pattern transferring.
%In \cite{jiangxueyao2022entity}, the diversity is achieved by mining topics (e.g., \texttt{View, Flag, Map}) in captions of popular entities' images (e.g., \texttt{Greenland}) and transferring these patterns to expand queries of other long-tail entities of the same type (e.g., \texttt{Country}) in the image retrieval process, and the relevance is achieved by filtering the local outlier factors according to the distribution of the same type popular entities' images.

We can also resolve the ranking problem by graph algorithms. A set of images could be represented as a graph, where images are nodes and visual similarities between images are weights of edges. 
Thus, the ranking of representative images reduces to finding an optimal path in a fully connected graph concerning re-weighted values of edges.
%Thus, the ranking of representative images reduces to finding an optimal path in a fully connected graph with respect to re-weighted values of edges.
~\cite{deselaers2009jointly} uses dynamic programming to search for the optimal sequence in an image graph, where the value of edges is a joint criterion combining diversity score and relevance score. Markov random walk is also used for the optimal sequence in~\cite{ji2011diversifying, yan2017graph}, where~\cite{ji2011diversifying} weights the values by Max-Min methods and~\cite{yan2017graph} reassigns the visits values between nodes according to their source clusters by a two-layer graph model.

These studies concentrate on text-image retrieval, and only \cite{jiangxueyao2022entity} is related to MMKGs. There are still many unsolved biases on the diversity of images of concepts derived from the Internet on gender, race, color and age, and the problem now relies heavily on crowdsourcing~\cite{yang2020towards}. 
\begin{comment}
    \begin{table}[t]
    	\small
    	\centering
    	\begin{tabular}{lcc}  
    		\toprule
    		symbol & entity & concept \\
    		\midrule
    		example & Li Na (tennis star)  & scientist \\
    		image    & \includegraphics[width=0.1\textwidth]{LiNa.png}  & \includegraphics[width=0.1\textwidth]{Einstein.jpg} \\
    		\bottomrule
    	\end{tabular}
    	\caption{Examples of concept grounding via entity grounding. The photo of Einstein can be grounded to the concept ``scientist'' since it is a typical entity of ``scientist''. }
    	\label{tab:concept_grounding_via_entity}
    \end{table}
\end{comment}

\begin{table}[t]
	\scriptsize
	\centering
	\begin{tabular}{lcc}  
		\toprule
		\tabincell{c}{concept\\type} & \tabincell{c}{visualizable\\concept} & \tabincell{c}{non-visualizable\\concept} \\
		\midrule
		example & Surgeon  & Physicist \\
		image    & \includegraphics[width=0.14\textwidth]{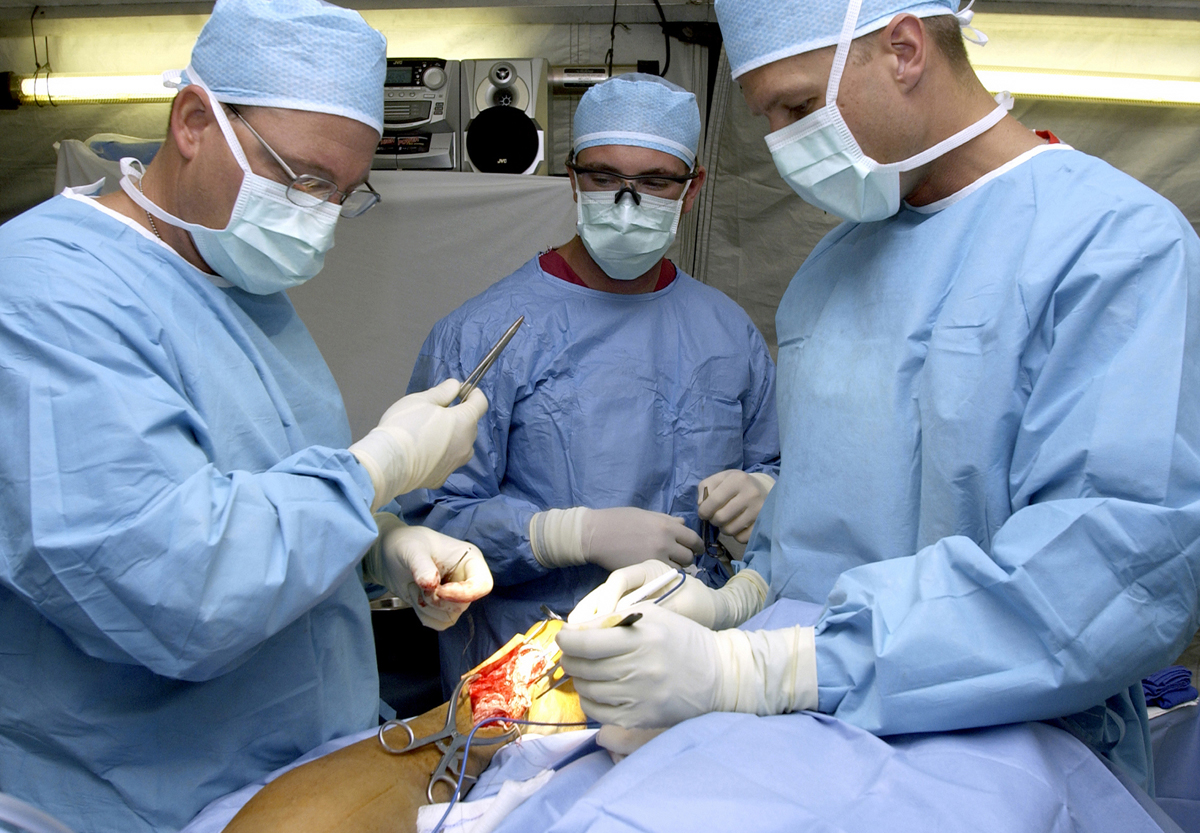}  & \includegraphics[width=0.08\textwidth]{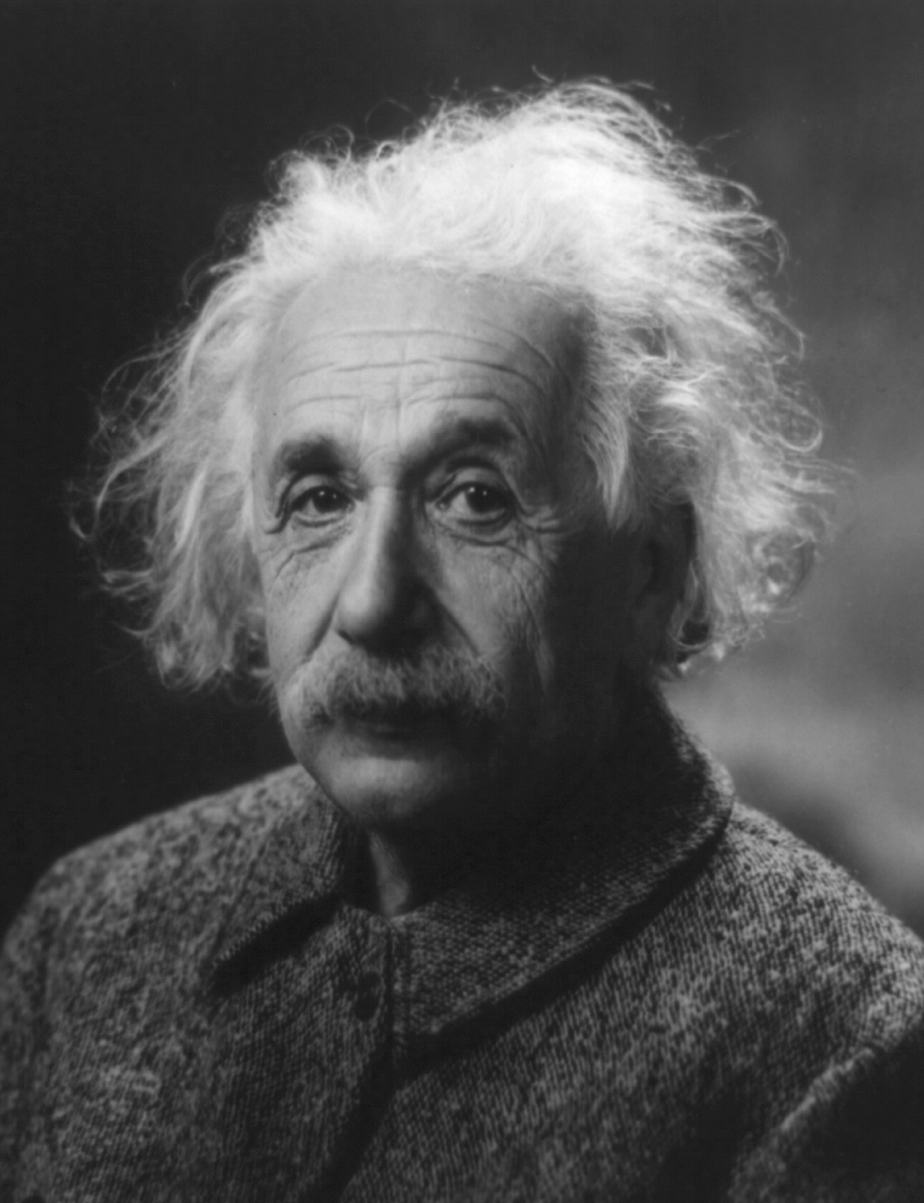} \\
		\bottomrule
	\end{tabular}
	\caption{Examples of visualizable concept grounding and non-visualizable concept grounding. The visualizable concept \texttt{Surgeon} can be grounded to \emph{the photo of doctors wearing surgical suits and performing surgery in the operating room}, and the non-visualizable concept \texttt{Physicist} can be grounded to \emph{the photo of Einstein} since \texttt{Einstein} is a typical entity of \texttt{Physicist}.}
	\label{tab:concept_grounding_via_entity}
	\vspace{-1.0em}
\end{table}

\vspace{5pt} \noindent{\bf OPPORTUNITIES}. As a fledgling area, many unsolved problems are left for future research. We give two examples below:

1) {\it Abstract Concept Grounding}. Previous work on concept visualization judgment seldom considers abstract concepts. But the abstract concepts could also be grounded in images. For example, \texttt{Happiness} are usually associated with \emph{smile}, and \texttt{Anger} are usually associated with \emph{an angry face}. Some abstract nouns have a diverse but fixed visual association, such as nature, human and action. For example, in~\cite{raguram2008computing} the images of \texttt{Beauty} are associated with following word clusters: \emph{woman/girl}, \emph{water/beach/ocean},  \emph{flower/rose}, \emph{sky/cloud/sunset}. Similarly, the image of \texttt{Love} are associated with following word clusters: \emph{baby/cute/newborn}, \emph{dog/pet}, \emph{heart/red/valentine}, \emph{beach/sea/couple}, \emph{sky/cloud/sunset}, \emph{flower/rose}. %It can be seen 
It shows that some abstract nouns often have generic and fixed images in terms of sentiment and discriminative images in terms of semantics.

2) {\it Gerunds Concept Grounding}. Gerunds are a special kind of nouns that could be transformed into verbs, such as \emph{singing} $\rightarrow$ \emph{sing}. \cite{antol2014zero} grounds many gerunds to images through crowdsourcing, such as \texttt{arguing with}, \texttt{wrestling with} and \texttt{dancing with}. These verbs about human interaction are sensitive to the features of body angle, gaze angle, the position of the joints and expression.

% done: scientist是一个视觉概念，所以Scientist--Einstein不合适。改为Physicist--Einstein
3) {\it Non-visualizable Concept Grounding via Entity Grounding}. 
If a concept is non-visualizable but its hyponym entities could be visualized, the concept could also be grounded via its entities.
%If a concept is non-visualizable but the entities of the concept could be visualized, the concept could also be grounded via its entities. 
For instance, a reasonable selection of the grounded image for such a concept is to use the image of the concept's most typical entity. As shown in Table \ref{tab:concept_grounding_via_entity}, we use a photo of \texttt{Einstein} to ground the concept \texttt{Physicist}. %\texttt{Scientist}. 
It is reasonable since most of us will think up with \texttt{Einstein} when we mention a \texttt{Physicist}. %\texttt{Scientist}. 
However, there are still a lot of unresolved questions: (a) In general, different people will come up with different typical entities for a concept, so we should address such subjectivity in concept grounding. Whether an entity is a typical one in the constrain of its concept? (b) We should choose several typical entities’ images to present that concept. How do we summarize and select typical entities to represent concepts? (c) Whether should we abstract common visual features from multiple images of entities?

% 添加Relation grounding is often considered as a fine-grained text-image retrieval problem. 
\subsubsection{Relation Grounding}\label{section:sec3.2.3}

Relation grounding is to find images from an image data corpus or the Internet that could represent a particular relation. The input could be one or more triples of this relation, and the output is expected to be the top-ranked representative images for the relation. For example, \texttt{(Justin Bieber, couple, Selena Gomez)} could be grounded to an image of ``\emph{Selena Gomez and Justin Bieber Kissed}'' instead of ``\emph{Selena Gomez and Justin Bieber worked out together}''.

\vspace{5pt}
\noindent{\bf CHALLENGES}.
% 1. top-rank ： 图片从哪儿来？要写一下，来自search engine   
% 2. 检索结果有噪声，所以已经过滤过entity, concept的matching model来过滤, 但是仍然不能体现relation.
% 确认一下，richpedia的工作，是否有确认两个实体都存在。
When we take a triple as a query to retrieve images for the relation, the top-ranked images are often more relevant to the subject and object of the triple but not to the relation itself. How to find images that could reflect the semantic relation of the input triples?
%by taking into account subject, object and relation?
%
%3) Multiple triples make up a graph with complex relations. Perhaps there are an appropriate image for the graph. \textbf{How to find the right image for a multi-relational graph if the graph is visualizable} ?

\vspace{5pt}
\noindent{\bf PROGRESSES}. Existing efforts on relation grounding mainly focus on the co-occurrence of visual objects in images or textual entities in captions. Richpedida~\cite{wang2020richpedia} proposes a very strong assumption that if there is a pre-defined relations (e.g., \texttt{nearBy} and \texttt{contain}) between two entities in the Wikipedia descriptions, the same relations also exist between two entities' corresponding visual objects. %entities. 
But in reality, it is more likely that the two objects do not simultaneously appear in one image. 
Even if they do, the relation shown in the image may not be the expected one. 
%Even if they do, there may be no expected relation in one image. 

% 输入是什么，输出是什么。做法是xxxxx。两三篇工作，分别输入是什么，输出是什么。
% 分两种：检索；或者多模态关系抽取。分别是怎么解决的。
Relation grounding could be modeled as a fine-grained text-image retrieval problem, where the triple \emph{(subject, relation, object)} is the query and candidate images are represented with the implicit or explicit structure information of the scene graph extracted. Specifically, each image could also be represented as a combination of multiple ($s$, $p$, $o$) by multi-branch CNN~\cite{elhoseiny2017sherlock} or graph convolutional neural network (GCN)~\cite{guo2020visual}.

\begin{figure}
	\centering
	\begin{tabular}{@{}c@{}}
		\includegraphics[width=0.9\linewidth]{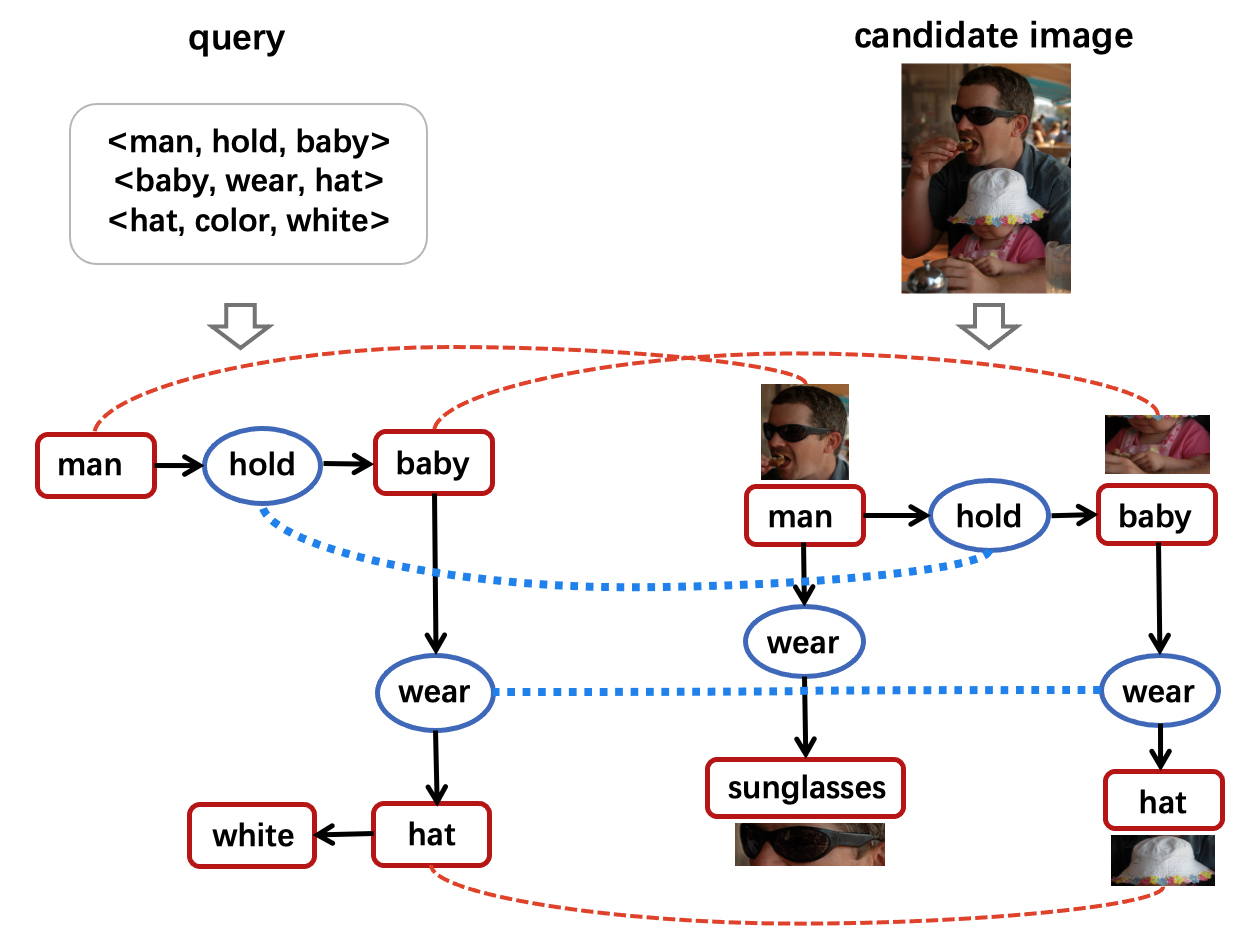} \\% [\abovecaptionskip]
	\end{tabular}
	\caption{Relation grounding is often considered as a fine-grained text-image retrieval problem. The queries are one or more triples, and the expected images should be consistent with entities and relations in the query. The figure shows an example of relation grounding by graph matching in~\cite{wang2020cross}.}
	\label{fig:graph_matching}
	\vspace{-1.0em}
\end{figure}

Instead of global matching of cross-modal embeddings, we expect the item-by-item matching of objects and relations. If we represent the textual query and candidate images into graphs, the relation grounding task turns into a task of graph matching, as illustrated in Figure~\ref{fig:graph_matching}. \cite{wang2020cross} represents the two graphs by GCN, in which objects are updated from themselves and relation nodes are updated from the aggregations of their neighbors. In predicting, the similarity between two graphs is measured by matching object nodes and relation nodes, respectively.

\eat{
Existing studies on relation grounding focus on spatial or action relations such as \texttt{left of}, \texttt{on}, \texttt{ride} and \texttt{eat}.

While textual queries can be represented as structured data in the format of \emph{(subject, relation, object)} through AMR graph~\cite{banarescu2013abstract}, candidate images could also be structured into a scene graph~\cite{johnson2015image}. 
Then, the structured text and structured images could be matched in fine-grained level by means of text-image matching or graph matching, which are introduced in the following.

1) {\em Text-Image Matching}. In text-image matching tasks, text and images are usually represented as vectors in a unified semantic embedding space. The images that best match the query is found by the similarity score of cross-modal representations. The cross-modal representations are usually fused by the mechanism of attention, so the disadvantage of global representation is lacking semantics of explicit fine-grained relations~\cite{wang2019camp}. In addition to representation-based retrieval, a more convenient method is caption-based retrieval like search engines on the Internet. The disadvantage of caption-based retrieval is that the visual features have not been used for matching. 

To represent explicit relation between objects, many studies concentrate on a image encoder considering local structure of images. The final image representations are the fusion of global visual features, local structure features and text aligned embeddings~\cite{elhoseiny2017sherlock, huang2018learning, guo2020visual}. In  \cite{elhoseiny2017sherlock}, all one-order (entity or concept), two-order (attribute or action), three order (triple) facts are modeled by a unified setting ($s$, $p$, $o$), which are respectively represented by the output of different branches of a multiple layer image encoder.~\cite{guo2020visual} uses a scene graph to represent all triple ($s$, $p$, $o$) in an image and uses graph convolutional neural network (CNN) to learn the visual relations. Finally, all the visual representations with relation features learned for each image have to be close to the text embeddings of corresponding words in the captions. Thus, the matching images can be directly retrieved by using a triple as a query instead of a sentence. 

VL-PTMs are a new alternative to image encoders that consider objects (entities or concepts) and triples. 
For each image-caption pair, a scene graph parser is employed to generate a scene graph with objects, attributes and relations from the caption of an image, then UNIMO~\cite{li2020unimo} randomly replaces the object, attribute and relation nodes of a scene graph with a different object, attribute or relation from the corresponding vocabularies to generate a large number of hard negative samples. 
ERNIE-ViL~\cite{yu2020ernie} enhances the capability of visual and language model by adding three pre-training tasks, object prediction, attribute prediction and relation prediction. 

2) {\em Graph Matching}. We expect the relation grounding through explicit matching of objects and relations, rather than implicit matching of unified cross-modal embeddings. A more convenient method is caption-based retrieval like search engines on the Internet,  matching tokens of entities and relations between the query and captions. The disadvantage of caption-based retrieval is that the visual features haven not been used for matching. For example, Richpedida~\cite{wang2020richpedia} proposes a very strong assumption that if there is a pre-defined relations (e.g., \texttt{nearBy} and \texttt{contain}) between two entities in the Wikipedia descriptions, the same relations also exist between two entities' corresponding visual entities. But in reality, it is more likely that the two objects are not simultaneously appear in one image. Even if they do, there may be no expected relation in one image.

If we represent the textual query and candidate images into graphs, the relation grounding task turns into a task of graph matching, as illustrated in Figure~\ref{fig:graph_matching}.  An image could be structured into a graph in which nodes are objects and edges are relations. The dependencies in the textual query could be modeled as the dependency parse tree, which is also a graph. A simple solution is that to only match objects and co-occurring relations in two graphs without predicting the relation types~\cite{johnson2015image}.~\cite{johnson2015image} assumes that if there are a relation between two entities, the relation is considered to be a match, which is also a strong assumption. Obviously, relation prediction module is essential. 
~\cite{wang2020cross} respectively represents two scene graph by GCN, in which objects updates from themselves and relation nodes update from the aggregations of their neighbors. When predicting, the similarity of two graphs in different forms is measured by object nodes matching and relation nodes matching.
}

% 把however前面的内容，放到progress
% progress分为两部：检索和多模态辅助的关系抽取。第一步，第二步，第三步分别是什么。
% 前者关注shallow relation, 后者问题很相关，但是人工构造的。 
\vspace{5pt} \noindent{\bf OPPORTUNITIES}. %Many problems are unsolved for relation grounding. For instance, 
Existing studies mainly focus on grounding spatial relations and action relations, such as \texttt{leftOf}, \texttt{on}, \texttt{ride} and \texttt{eat}, which could be observed visually in images. However, most semantic relations such as \texttt{isA}, \texttt{Occupation}, \texttt{Team} and \texttt{Spouse} may not be that visually obvious in images. There is a lack of training data for these relations, thus it is difficult to train models to retrieve images with the above solutions. Fortunately, some datasets~\cite{zheng2021mnre,zheng2021multimodal} of relation extraction based on textual named entities and visual relations could be helpful.
%
%There are also multi-hop relation \texttt{Trump $\xrightarrow{presidentOf}$ USA $\xrightarrow{capital}$ Washington D.C.}.
%
%1) {\it Semantic Relation} The main challenge is the lack of datasets on this type of relations. Therefore it is difficult to train models to retrieve images by learning unified representation or matching nodes of graphs.

%2) {\it Complex Multiple Relational Grounding}. In the real applications, multiple relational facts might has close semantic relations with each other, which make up a connected graph. Complex multiple relational grounding is to ground a path or a subgraph, the challenge of which is that these relation groundings are usually interleaved with each other in a complicated way. We have to find the images that fully embody the composite semantic relations. The composite semantic in many cases is only implicit expressed and might change over time. As shown in Table \ref{tab:other_relation_grounding}, if a subgprah in KG contains \texttt{Trump}’s wife, daughter, grandson etc, the best grounding might be a \texttt{Trump}'s family photo. Obviously, it is not trivial for machines to automatically select the best image from all image relevant to \texttt{Trump}.

\begin{table*}[htbp]
    %\footnotesize
    \scriptsize
    \centering
    \begin{tabular}{m{0.15\textwidth}<{\centering}|m{0.18\textwidth}<{\centering}|m{0.57\textwidth}<{\RaggedRight}}
         \hline
            \tabincell{c}{Multimodal\\Application} & \tabincell{c}{Benchmark Datasets} & \makecell[c]{Advantages with MMKG} \\ 
            
            \hline
            
            \multirow{4}*{\tabincell{c}{Entity Recognition\\and Linking}} & \multirow{4}*{\tabincell{c}{Twitter2015~\cite{zhang2018adaptive} \\  Twitter2017~\cite{Seungwhan2018Zeroshot} \\ Weibo~\cite{zhang2021attention} \\
            WikiDiverse~\cite{wang2022wikidiverse}} } & 1.background knowledge provides deep features of images \\ & & 2.images provide necessary complementary information, help to capture the relationship among mentions and entities \\ & & 3.learn distributed representations for each entity with multi-modal data \\
            
            \hline
            
            \multirow{5}*{\tabincell{c}{VQA}} &
            \multirow{5}*{\tabincell{c}{GQA~\cite{hudson2019gqa} \\ OK-VQA~\cite{marino2019ok} \\ FVQA~\cite{wang2018fvqa} \\ KVQA~\cite{shah2019kvqa} \\ KB-VQA ~\cite{wang2015explicit}}} & 1.provide knowledge about the named entities and their relations in the image, leading to \\ & & deeper visual content understanding \\ & & 2.conduct the reasoning process and predict the final answers in a more explicit way with symbolic knowledge from MMKG \\ & & 3.refine the answers with more interpretability and generality \\ 
            
            \hline
            
            \multirow{3}*{\tabincell{c}{Image-text\\Matching}} & \multirow{3}*{\tabincell{c}{Flickr30k~\cite{young2014image} \\ MSCOCO~\cite{lin2014microsoft} \\ Visual Genome~\cite{krishna2017visual}}} & 1.expand more semantic concepts \\ & & 2.introduce informative relationships between visual concepts by constructing scene graphs \\ & & 3.enhance the reasoning capabilities of multi-modal data with graph-structured information \\
            
            \hline
            
            \multirow{1}*{\tabincell{c}{Image Tagging}} & \multirow{1}*{\tabincell{c}{NUS-WIDE~\cite{chua2009nus} }} & help disambiguation the concept and relate them better to images \\
            
            \hline
            
            \multirow{3}*{\tabincell{c}{Image Captioning}} & \multirow{3}*{\tabincell{c}{MSVD~\cite{guadarrama2013youtube2text} \\ MSCOCO~\cite{lin2014microsoft} \\
            GoodNews~\cite{tran2020transform}}} & 1.enable the understanding of unseen objects with MMKG symbolic knowledge \\ & & 2.leverage MMKG for relational reasoning to generate more accurate and reasonable captions \\ & & 3.capture fine-grained relationships between entities in different modalities \\
            
            \hline
            
            \multirow{2}*{\tabincell{c}{Visual Storytelling}} & \multirow{2}*{VIST Dataset~\cite{huang2016visual}} & 1.triples in MMKG provide explanation and traceability for described facts \\ & & 2.provide a strong logical inference between images for more fluent story \\ 
            
            \hline
            
            %\multirow{2}*{\tabincell{c}{Scene Graph \\ Generation}} & \multirow{2}*{\tabincell{c}{VRD~\cite{lu2016visual} \\ VG-MSDN~\cite{li2017scene}}} & 1.complement missing relations in the scene graph \\ & & 2.capture more semantic information between objects, alleviating data bias \\
            %\hline
            
            \multirow{4}*{\tabincell{c}{Recommender \\ System}} & \multirow{4}*{\tabincell{c}{MovieLens~\cite{harper2015movielens} \\ IntentBooks~\cite{uyar2015evaluating} \\ Dianping~\cite{sun2020multi} \\ KKBOX~\cite{huangkkbox}}} 
            & 1.provide background knowledge for items with rich semantics to solve the cold-start problem \\ & & 2.learn through rich path semantics across different modalities in MMKG and produce an interpretable and explicit recommendation \\ & & 3.construct personalized MMKG for items and model entity relation reasoning between them \\
            
            \hline
            
            %\multirow{4}*{\tabincell{c}{\\ Video Event \\ Recognition\\}} & \multirow{4}*{\tabincell{c}{\\ TRECVID MED~\cite{over2013trecvid} \\ TRECVID AVS~\cite{awad2017trecvid} \\ EVVE~\cite{revaud2013event}}} 
            %& \\ & & 1.generalize concept-event relations by constructing MMKG to align concept with its related images \\ & & 2.get better presentations for video streams with concepts of great diversity \\ & & \\
            %\hline
            
    \end{tabular}
    \caption{Benchmark datasets for their corresponding multimodal applications incorporating MMKGs.}
    \label{tab:multimodal_application}
    \vspace{-1.0em}
\end{table*}

\subsection{Comparing Two Construction Ways}
\label{secion:sec3.3}

There are several differences between the image labeling and symbol grounding solutions for constructing an MMKG in the aspects of applicable scenarios, construction efficiency, quality, etc. We analyze kinds of MMKGs in which multi-modal data are not only images but also code, audio or video, and summarize these differences as follows:

1) {\it Applicable Scenarios}. If the multi-modal data are treated as first-class citizens in some scenarios, the multi-modal data labeling way is more preferred to construct the MMKG, such as unearthed oracle bones' photos in oracle bones recognition system~\cite{xiong2021oracle}, teachers' class audios in educational services~\cite{li2022medukg} and the movies' videos in deep video understanding tasks~\cite{baumgartner2020towards}. 
If the multi-modal data collected is redundant and noisy, the multi-modal data labeling way may produce many low-quality (such as repeated or mismatching) visual entities. In this case, the symbol grounding way is preferred to construct the MMKG because the symbols in KGs have already been well filtered and refined, such as the movie ontologies in recommendation systems~\cite{sun2020multi}, product ontologies in e-commerce dialogue systems~\cite{xu2021alime} and paper ontologies in academic information retrieval and KBQA~\cite{kannan2020multimodal,deng2021gakg}. 

Whether multi-modal data or symbolic knowledge is first-class citizen depends on what kind of knowledge we want the MMKG to provide. For example, in \cite{deng2021gakg} when we want to know the relations between geoscience academic papers and maps in them, the papers are first-class citizens; when we want to know the relations between maps and regions pointed in these maps, the maps are first-class citizens. 

2) {\it Efficiency.}  
The symbol grounding solutions are usually retrieval-based methods~\cite{liu2019mmkg,wang2020richpedia,alberts2020visualsem,bloem2021kgbench,xu2021alime,zheng2021pay} and the multi-modal data labeling solutions are usually classification and detection methods~\cite{li2020gaia,wen2021resin,ma2022mmekg,baumgartner2020towards,xiong2021oracle}.
Extracting entities, concepts and relations in multi-modal data labeling solutions is time-consuming~\cite{chen2013neil}. Therefore, it will be an excellent choice to start the construction of an MMKG from scratch using the symbol grounding solutions.% from the perspective of cost. 
For example, NEIL~\cite{chen2013neil} initially collects image datasets by retrieving images from search engines with ontologies of NELL~\cite{carlson2010toward} as queries and then extracts objects and relations in these images.

3) {\it Quality.} Except for the quality of extraction models, the multi-modal data labeling solutions have to solve the problem of coarse-grained labeling and inappropriate semantic hierarchies. Symbol grounding solutions could solve these problems. 
However, the symbol grounding way also faces the problem of missing and mismatching images of symbols. For example, it is easy to find a bad image for a long-tail entity from search engines. Because such an entity might have no image on the web, any clicked image is misleading to a mistake grounding.

\section{Application}
\label{sec:application}
%There is an increasing trend to use multi-modality to improve the performance of real wold tasks as introduced in Sec.~\ref{sec:mmtasks}. However, the current efforts to use multi-modal information is still limited without the support of large-scale MMKGs.
%
After a systematic review of MMKG construction, this section explores how the knowledge in MMKGs can be applied to and benefit a wide variety of downstream tasks. For a quick overview, Table~\ref{tab:multimodal_application} lists some mainstream application tasks, their benchmark datasets, and the advantages brought by MMKGs.
We categorize such tasks into (i) in-KG applications (Sec.~\ref{sec:inKG}) , (ii) out-of-KG applications (Sec.~\ref{sec:outKG}) and (iii) domain applications(Sec.~\ref{sec:domain}), discussed as follows.

\subsection{In-MMKG Applications}
\label{sec:inKG}

In-MMKG applications refer to tasks conducted within the scope of the MMKG where the embeddings of entities, concepts and relations are already learned. Thus, before introducing in-MMKG applications, we briefly go through the distributed representation learning of the knowledge in MMKGs, also named MMKG embedding.

The MMKG embedding models are developed from the embedding models on conventional KGs, 
i.e., {\it semantic matching based} models, RESCAL\cite{maximilian2011athree} and its variants\cite{zhaolizhang2022multiscale, pezeshkpour2018embedding}, which measure the possibility of existence of triple ($\bm{h}$, $\bm{r}$, $\bm{t}$) by the calculation of $\bm{h}$, $\bm{r}$, $\bm{t}$ in vector space,
%i.e., {\it distance-based} models~\cite{bordes2011learning}, which consider that the head entity and tail entity of the same triple should be close in the projection space, 
and {\it translational distance based} models, TransE~\cite{bordes2013translating} and its variants~\cite{wang2014knowledge, lin2015learning, ji2015knowledge}, which should conform to the assumption: $\bm{t} \approx \bm{h} + \bm{r}$. $\bm{h}$, $\bm{t}$, $\bm{r}$ is respectively the vector representation of head entity, tail entity and relation in a triple. 
There are two additional issues in dealing with multi-modality data: how we effectively encode the vision knowledge and information contained in images, and how we fusion knowledge of different modalities.
1) {\it Vision Encoders}. 
%There are many ready techniques for image information encoding in CV, including numerous conventional explicit visual features~\cite{...} such as GHD, HOG and CLD, and some hidden features from neural network such as Convolutional Neural Network~\cite{bordes2013translating,lin2015learning,xie2016representation,lin2015modeling}.
%
With the development of deep learning, hidden features gotten from  CNN~\cite{mousselly2018multimodal,Rettinger2017Towards,pezeshkpour2018embedding} or Transformers~\cite{chen2022hybrid} are the main image embeddings used in MMKG representation, while other explicit visual features such as GHD, HOG, CLD can hardly be leveraged in MMKG representation.
2) {\it Knowledge Fusion}. There are two ways to fuse the knowledge embeddings of multi-modalities: combining every single modal representation trained in its own vector space (such as concatenation, average pooling, SVD and PCA)~\cite{onoro2017answering,liu2019mmkg,Rettinger2017Towards}, or further learning a unified embedding by projecting different modal representations into the same space~\cite{xie2016image,mousselly2018multimodal,pezeshkpour2018embedding}.
%To fuse the knowledge embeddings of multi-modalities, various fusion ways are considered, including simple concatenation, average of multiple modalities' embeddings and normalization-based or weighted SVD and PCA~\cite{Rettinger2017Towards}. 
%
While some methods~\cite{Rettinger2017Towards} take the fused results as the MMKG embedding directly, the other methods~\cite{pezeshkpour2018embedding} further train the uni-modal representations on a well-designed objective function.

In the following, we introduce four well-studied in-MMKG applications including {\it link prediction}(Sec.~\ref{section:sec4.1.1}), {\it triple classification}(Sec.~\ref{section:sec4.1.2}), {\it entity classification}(Sec.~\ref{section:sec4.1.3}), and {\it entity alignment}(Sec.~\ref{section:sec4.1.4}).

%All these applications are sorts of refinement (e.g., comple- tion or de-duplication) of the input KG [123], from different viewpoints and application context.

%In-MMKG applications refer to the tasks aiming at addressing the issues within MMKGs, such as multi-modal knowledge graph completion, multi-modal triple classification, and entity alignment between MMKGs. %~\cite{xie2016image,mousselly2018multimodal,liu2019mmkg, guo2021multi, chen2020mmea}. %Knowledge graph completion and triple classification are also usually used to evaluate the performance of MMKG representation or to guide the design of objective function in representation learning.

\subsubsection{Link Prediction}\label{section:sec4.1.1}

Link prediction in MMKG~\cite{xie2016image,mousselly2018multimodal} aims to complete a triple $(h,r,t)$ when one of the entities in $h,r,t$ is missing, i.e., predicting $h$ in $(?,r,t)$ or predicting $t$ in $(h,r,?)$.
A similar task is to predict the missing relation between two given entities, i.e, predicting $r$ in $(h,?,t)$.

Conventionally, link prediction on KGs can be processed with a simple ranking procedure, which finds the best fit entity to complete a triple from all the candidate entities.
Specifically, in the training stage, the embedding model learns an embedding for each entity or relation, for instance, with the training objective $\bm{t} \approx \bm{h} + \bm{r}$ defined by TransE~\cite{bordes2013translating}. Then in the prediction stage, the most matching $h$ in $(?,r,t)$ is found by ranking all candidate head entities $h^*$ according to a score function like $\mathop{\arg\max}_{h^*}\phi(h^*, r, t )$, where the score function is diverse in different embedding models~\cite{rossi2021knowledge}. %The prediction of $r$ or $t$ is performed analogously. 

%Compared to the task in traditional KGs, the images attached to entities and relations in MMKGs could provide extra visual information to enhance the embedding learning quality.
Compared to the task in traditional KGs, the images fused into representations of entities and relations in MMKGs could provide extra visual knowledge to enrich the information of embedding.
For instance, the images of a person might provide evidence for the person's age, profession, and designation~\cite{pezeshkpour2018embedding}.
%
% 放在了前面一节的knowledg fusion
%\zxr{The visual information can be added into entities and relations through two ways: combining every single modal representation separately trained in its own vector space (such as concatenation, average pooling, SVD and PCA)~\cite{onoro2017answering,liu2019mmkg,Rettinger2017Towards}, or further learning a unified embedding by projecting different modal representations into the same space~\cite{xie2016image,mousselly2018multimodal,pezeshkpour2018embedding}.}

%: 1) the concatenation of different modal embeddings. 2) the unified embedding by project different modal data into the same space. 
%现在加图像的方式有直接concate拼接两个模态的embedding的方式，也有对齐之后使用对齐的embedding的

%\zxr{\sout{Some other work}}
This task is different in existing MMKGs depending on the scenario.
IMAGEgraph~\cite{onoro2017answering} proposes to express the relation prediction between unseen images and multi-relational image retrieval as visual-relational queries, such that these queries could be leveraged for MMKG completion. %, with the query type $(h, ?, t)$ for link prediction task and the query types $(h, r, ?)$ and $(?, r, t)$ for entity prediction task for completing the triples. 
%
%These queries could express the relation prediction between unseen images and multi-relational image retrieval as visual-relational queries. %Motivated by the fact the number of entities is often orders of magnitude greater than the number of relations in a typical KG, the presented visual-relational KG inputs high-level features of images directly and focuses on combining visual information both during the learning and query time.
%
Compared to the conventional way, IMAGEgraph performs better on the relation and head/tail entity prediction tasks and is able to be generalized to unseen images, to answer some zero-shot visual-relational queries. 
For example, given an image of an entirely new entity not part of the KG, this approach can determine its relation with another given image for which we do not know the underlying KG entity.
%For example, given an image of an entirely new entity that is not part of the KG, this approach can determine its relation with another given image which we do not know the underlying KG entity.

Similarly, MMKG~\cite{liu2019mmkg} constructs three datasets to predict the multi-relational links between entities, with all the entities associated with numerical and visual data. However, it only focuses on the \texttt{sameAs} link prediction task and answers such queries for MMKG completion. Three quite heterogeneous knowledge makes MMKG a vital benchmark to measure the performance of multi-relational link prediction methods and validates the hypothesis that different modalities are complementary for the \texttt{sameAs} link prediction task.

\subsubsection{Triple Classification}\label{section:sec4.1.2}

Triple classification aims to distinguish correct triples from incorrect ones, which can also be seen as a sort of KG completion task.
Based on the embedding model learned on an MMKG, each triple could be calculated with an energy score $E(h,r,t)$.
Different threshold $\delta_r$ is set for each relation $r$, and a triple will be predicted to be negative if its energy score is higher than $\delta_r$.
In classification models, correct triples are corrupted by replacing one of the $h, r, t$ to generate negative data~\cite{xie2016image,mousselly2018multimodal}.
%To prepare the training data for the classification model, correct triples are corrupted by replacing one of the $h, r, t$ to generate negative data~\cite{xie2016image,mousselly2018multimodal}.

\subsubsection{Entity Classification}\label{section:sec4.1.3}

Entity classification categorizes entities into semantic categories, i.e., concepts of different grains in the MMKG.
Entity classification can also be regarded as a special link prediction task, where the relation is %set to 
\texttt{IsA} and the tail of the triple to be predicted is a concept in the MMKG. 

Various entity classification models have been proposed for traditional KGs, which could also be adopted in MMKGs.
But the rich multi-modal data for entities and concepts in MMKGs cannot be fully utilized without a good MMKG embedding model. For instance, some efforts~\cite{wilcke2020end,bloem2021kgbench} work on learning embeddings for entities and concepts from several different types of modalities and then encode them to a joint representation space.
However, \cite{bloem2021kgbench} argues that this task in KGs cannot be solved purely by node embedding models, and the graph structures should also be considered. Therefore, \cite{bloem2021kgbench} proposes a collection of extensive and high-qualified multi-modal benchmarks for precisely evaluating node classification tasks on MMKGs.

% a high enough number of labeled nodes in the test set for precise %measurement of performance, and (b) a rich enough variety of multimodal
%information to learn from.

\subsubsection{Entity Alignment}\label{section:sec4.1.4}

Entity alignment works on aligning entities that refer to the same real-world identity in different MMKGs. It is a viable way to integrate two MMKGs into one when there are overlaps.

The core idea is to learn representations for entities in different KGs and then evaluate the similarity between each entity pair between the two KGs.
The features used in entity embedding between two traditional KGs include in-KG context information (e.g., the semantics of OWL properties, co-occurrence of neighbors, compatible attribute values) and external information (e.g., external lexicons and Wikipedia links).
For MMKGs, due to the introduction of multi-modal features, some entity-alignment oriented MMKG embedding models are proposed~\cite{guo2021multi, chen2020mmea}.
Feature vectors are encoded for different modalities respectively and then merged into one to represent the entity by the knowledge fusion techniques mentioned at the beginning of this subsection. 
One work~\cite{guo2021multi} uses ranking loss as the loss function, while another~\cite{chen2020mmea} designs a loss function $L = \alpha||e-e_s|| + \beta||e-e_n|| + \gamma||e-e_i||$ to enhance the complementarity of multiple modalities, where $e_s, e_n, e_i$ is the embedding of three different modalities respectively $e$ is the final embedding of the entity, and $\alpha, \beta, \gamma$ is ratio hyper-parameters for each modality.

Another line of work~\cite{liu2019mmkg} elaborates a Product of Experts (PoE) model to answer queries such as $(h?, sameAs, t)$ or $(h, sameAs, t?)$ where $h$ and $t$ are from different KGs. 
By incorporating \cite{garcia2017kblrn} and extending it to visual features, the end-to-end learning framework is superior to the concatenation and an ensemble type of approach for entity alignment.
%By incorporating \cite{garcia2017kblrn} and extending it to visual features, the end-to-end learning framework has found to be superior to the concatenation and an ensemble type of approach as for entity alignment.

%The metrics are as same as the knowledge graph completion. Specially, the performance will be evaluated under different percentages of the given alignments $P(\%)$. And the percentages are often set to $20\%$, $50\%$ and $80\%$.

%\noindent {\bf OPPORTUNITIES.} As is depicted in the overall variation of energy function, embedding of entities in multiple modalities including structure information (i.e. $\bm{h_S}$) and other multiple modalities (i.e. $\bm{h_I}$) are encoded respectively and used to optimize the energy function, while for embedding of relations, only structure information (i.e. $\bm{r}$) in graph is encoded.

%Similarly, relations can also be characterized in multiple modalities. Take images for example, the predicate \textit{president\_of} can also be depicted with an image of \textit{a president speaking at the podium}. Besides, the predicate \textit{wife\_of} can be associated with an image of \textit{a couple holding hands}.

%With the proper encoders of relations in multiple modalities, the embedding of relation $\bm{r}$ can be further extend to structure-based representation (i.e. $\bm{r_S}$) and other-modality-based representation (i.e. $\bm{r_I}$).

\subsection{Out-of-MMKG Applications}
\label{sec:outKG}

The out-of-KG applications refer to the downstream applications that are not limited to the boundary of MMKGs but could be assisted by them.
In the following, we introduce several such applications as examples. Instead of providing a systematic reviews to all the solutions of these tasks, we mainly focus on introducing how MMKGs are utilized, and the advantages of MMKGs compared with other solutions.

\subsubsection{Multi-modal Entity Recognition and Linking}\label{section:sec4.2.1}

Named entity recognition (NER) with plain texts has been studied extensively. Ambiguity and diversity of entity mentions have always been the key challenges.
Recent work focusing on detecting entities from texts attached with images is defined as multi-modal NER (MNER)~\cite{zhang2018adaptive, Seungwhan2018Zeroshot}, where images could provide necessary complementary information for entity recognition.

MMKGs can enhance MNER by providing vision features of entities to enhance the representation of images or text. 
For instance, \cite{chen2021multimodal} compares the given image with the images of candidate entities (from text) and two-hop neighborhood entities in the MMKG to find the most relevant entity as external background knowledge for disambiguation.
%
%\zxr{\sout{MMKG plays an important role in MNER by providing vision features to describe different types of entities, such that the MNER models could better use the vision features of the images attached with the text for entity recognition.}}
%
%\zxr{\sout{For instance, \cite{chen2021multimodal} proposes to use the background knowledge of images in MMKGs to help capture deep features of image to avoid error from shallow features. }}
%
\cite{wang2022wikidiverse} also employs MMKGs to retrieve more labels as related words based on the co-occurrence frequency between entities. With the expansion of entity type labels from MMKGs, more task-specific salient features are highlighted, avoiding being neglected in cross-modal interactions and improving the performance of MNER.

Given a text with images attached, multi-modal entity linking (MEL) uses textual and visual information to map an ambiguous mention in the text to an entity in a given KG~\cite{adjali2020multimodal}.
Although some early efforts do MEL based on a traditional KG, increasingly recent work uses MMKGs for linking.
%Although some early efforts does MEL based on a traditional KG, more and more recent works uses MMKGs for linking.
%
MEL utilizes the knowledge with images in an MMKG in two ways: (1) providing the target entities to which the entity mentions should be linked; (2) learning distributed representations for each entity with multi-modal data, which are then used to measure the correlation between a mention and an entity. 
%The knowledge with images in an MMKG are utilized by MEL in two ways: (1) Providing the target entities to which the entity mentions should be linked; (2) Learning distributed representations for each entity with multi-modal data, which are then used to measure the correlation between a mention and an entity.
%
%A core techniques in MEL lies on how to learn a good distributed representation for both mentions and entities with cross-modal data.
%
The usage of visual information with images would help to capture the relationship among mentions and entities~\cite{moon2018multimodal, adjali2020multimodal}, but the irrelevant part with images may also become noises and bring negative impact to the representation learning for both mentions and entities.
To remove the side effect, a two-stage image and text correlation mechanism is proposed to filter out the irrelevant images based on the pre-defined threshold, and the multiple attention mechanisms are also utilized to capture the critical information in the mention representation and entity representation by querying multi-hop entities around the mention's candidate entities~\cite{zhang2021attention}.

\subsubsection{Visual Question Answering}\label{section:sec4.2.2}

Visual question answering (VQA) is challenging, requiring accurate semantic parsing of the questions and an in-depth understanding of the correlations between different objects and scenes in the given image.
%Visual question answering (VQA) is a challenging task, which requires not only accurate semantic parsing to the questions, but also in-depth understanding to the correlations between different objects and scenes in the given image.
%
In most recent VQA benchmark datasets such as GQA~\cite{hudson2019gqa}, OK-VQA~\cite{marino2019ok} and KVQA~\cite{shah2019kvqa}, many questions require visual reasoning combined with external knowledge. The newly proposed VQA tasks bridge the discrepancy that humans can easily combine knowledge from various modalities to answer visual queries.
%
%For example, the question ``\emph{What color is the apple on the white table?}'' needs the following reasoning through several triples: table $\rightarrow$ $filter$: white $\rightarrow$ $relate(subject, on)$: apple $\rightarrow$ $query$: color~\cite{hudson2019gqa}. 
%
%However, in most cases, the visual content is not sufficient to answer the proposed question, which requires knowledge beyond the image, such as the knowledge about the named entities.
%
%Several typical efforts includes the OK-VQA~\cite{marino2019ok}, FVQA~\cite{wang2018fvqa} and KVQA~\cite{shah2019kvqa}, which only contains questions that require external knowledge to answer.
%
For example, in the question ``\emph{Which American President is associated with the stuffed animal seen here?}'', if the stuffed animal in the image is detected as ``\texttt{Teddy Bear}'', the answer inferred through KG will be ``\texttt{Theodore Roosevelt}'', who is often referred as ``\emph{Teddy Roosevelt}'', and after whom \emph{Teddy Bear} is named~\cite{marino2019ok}.

Obviously, reasoning only by semantic parsing and matching can not answer the above question~\cite{wang2015explicit}. 
In this case, MMKGs could help in three aspects. % and also enhance the interpretability of answers.
First, MMKGs provide external knowledge about the named entities and their relations in the image, leading to deeper visual content understanding. 
Second, the facts about visual entities in the image and textual entities in the question from existing MMKGs help to re-weight the answer~\cite{yu2020cross}, which also benefits from the unified representation of all modal resources including images, questions and structured facts. 
Third, entities and relation triples of different modal in MMKGs can be represented as nodes and edges in a heterogeneous graph and represented in a unified format, which facilitates explicit reasoning with heuristic rules, SPARQL queries~\cite{wang2015explicit} or weighted passing messages between GNN nodes~\cite{wang2015explicit,yu2020cross,marino2021krisp}.

%\zxr{\sout{Extracting relationships between visual concepts and understanding semantic information in questions are two key issues for VQA. However, without incorporating more knowledge of various modalities, it is not scalable to reason just over image-question-answer triples by semantic parsing and matching and can hardly generalize to more complicated cases~\cite{wang2015explicit}. MMKG helps to deal with the problems and enhances the interpretability to answers. First, MMKG provides knowledge about the named entities and their relations in the image which leads to deeper visual content understanding.Second, the structured symbolic knowledge in MMKG makes it a more explicit way to conduct the reasoning process and predict the final answers.}} \wxd{\sout{Second, the structured symbolic knowledge in MMKGs including entities and relations can be modeled by graph representation learning or represented by graph neural network, to further conduct more explicit reasoning and predict more explainable results.}}

Some recent efforts tend to construct MMKGs for VQA by combining existing KGs and well-annotated image datasets. For example, the explicit knowledge in \cite{marino2021krisp} has four sources: \texttt{hasPart} triples from hasPart KB~\cite{bhakthavatsalam2020dogs}, \texttt{hasPart/isA} triples from DBpedia~\cite{auer2007dbpedia}, commonsense triples from ConceptNet~\cite{liu2004conceptnet}, and location triples of visual objects from Visual Genome~\cite{krishna2017visual}.
The model fusing explicit symbolic knowledge from the MMKG and implicit knowledge from VL-PTMs outperforms the pure VL-PTMs, and most of the knowledge in the MMKG is non-overlapping with the implicit knowledge in VL-PTMs\cite{marino2021krisp}.

\subsubsection{Image-Text Matching}\label{section:sec4.2.3}

Image-text matching is a fundamental task in many cross-modal applications like image-text and text-image retrieval, which aims to output a semantic similarity score between the input image and text pair~\cite{ma2015multimodal, huang2018learning, lee2018stacked, ma2019matching, zheng2020dual}.

Image-text matching is usually achieved via mapping texts and images into a joint semantic space and then learning unified multi-modal representations for the similarity calculation.
A general method is to exploit a multi-label detection module to extract semantic concepts and then fuse these concepts with the global context of image~\cite{huang2018learning, huang2018image, wang2019language}. However, it is difficult for pre-trained detected-based models to find long-tail concepts, which constrains models to those detected concepts and leads to poor performance.

%As an enhancement of semantic information, MMKGs can be well applied here to provide external knowledge to improve the interpretability of model as well as visual representation learning.
%
%...\leeon{Xiaodan: summarize the ways the MMKGs are applied here with one paragraph.}

To overcome the bias in the training data for retrieval tasks, MMKG could be leveraged to expand more visual and semantic concepts leveraging the relations between multi-modal entities. Besides, MMKGs can also help to construct scene graphs, which introduce informative correlation knowledge between visual concepts and further enhance image representations. For example, the concept pairs that frequently co-occurred in the multimodal triples of an MMKG, such as \texttt{house-window} and \texttt{tree-leaf}, can be extracted to enhance the representation of concepts in images, thus providing a solid context signal for semantic understanding of images and leads to improved performance of image-text matching~\cite{shi2019knowledge}. 
Besides, considering that one key step in the image-text matching task is to align both local and global representations across different modalities, some efforts propose incorporating relations in MMKGs to represent both image and text with higher-level semantics~\cite{wang2020consensus}. %instead of exploiting only image-text instance pairs
Such graph-structured information better enhances the reasoning and inference capabilities of multi-modal data with more interpretability. MMKG also helps cross-modal alignment by learning a more unified multimodal representation.

%\eat{
%One powerful solution is to incorporate external knowledge to expand more semantic concepts precisely. 
%
%For example, \cite{shi2019knowledge} constructs an auxiliary common-sense knowledge base by extracting frequently co-occurred concept pairs in all triples of Visual Genome (e.g. ``house-window'' and ``tree-leaf'') to enhance the representation of concepts in images, which provides strong context signal for semantic understanding of images and leads to improved performance of image-text matching.

%In addition to incorporating external knowledge to expand concepts and enhance image representations,~\cite{wang2020consensus} introduces graph-structured information by constructing a concept correlation graph between instantiated concepts, and aligns cross-modal information at the consensus level to measure cross-modal similarity via disentangling higher-level semantics for both image and text, as well as to further improve its interpretability.

%In the model architecture of~\cite{wang2020consensus}, CVSE relies on image captioning corpus to exploit semantic concepts and their relations. The concept correlation graph is then built by constructing a conditional probability matrix P to model the correlation between different concepts and the consensus-aware concept representations are extracted from graph convolutional network afterwards. The instance-level representations and consensus-level representations are then integrated to comprehensively characterizing the semantic meanings of the visual and text modalities.
%}

\subsubsection{Multi-modal Generation Tasks}\label{section:sec4.2.4}
Several vision-text generation tasks, such as image tagging, image captioning, visual storytelling, etc., could benefit from MMKGs.

\noindent {\bf Image Tagging}. Traditional image tagging methods are limited by biased distribution, noise and imprecise tags.
MMKGs not only establish a well-organized taxonomy of concepts (such as synonyms, hypernyms and hyponyms) but also provide corresponding representative and discriminative images for concepts, thus they could greatly alleviate the effects of distribution bias of tags and noisy tags.
For example, \cite{chaudhary2019enhancing} constructs an MMKG called VTKB containing hierarchical concepts, linking concepts of original tags to images and linking images by the similarities of embeddings.
The candidate concept set is a subset of the union of the parent, the child, the part, the whole, synonyms, hypernyms, hyponyms and related concept sets of the original coarse-grained tags of images.
Finally, the re-generated fine-grained tags are those concepts that best match nearest neighbor images, where the candidate concept set depends on the type of bias specified in advance. 
%
% 结果：有无MMKG的对比
The experimental results show that the proposed method with MMKGs achieves higher mean average precision than the baselines without MMKGs. MMKGs help to generate more relevant candidate tags and are more capable of disambiguating them than ConceptNet, WebChild and ImageNet.

\noindent{\bf Image Captioning}. The mainstream statistic-based image captioning models have two weaknesses:
First, they heavily rely on the performance of object detectors. The encoder-decoder framework with separate procedures of detection and captioning always leads to semantic inconsistency between the pre-defined objects/relations and target textual descriptions. 
Second, unseen objects always pose great challenges to them. The models trained on image-caption parallel corpora always fail to describe unseen objects and concepts.

%\leeon{The following two paragraphs need to be confirmed and concise. How effect MMKG bring to the task, and how it is used.}

Fortunately, MMKGs could help to alleviate the two obstacles in the following ways:
1) Some efforts~\cite{hou2019relational} propose to leverage MMKG for relational reasoning, which results in more accurate and reasonable captions. 
More specifically, a semantic graph could be built for visual and knowledge vectors embedded from candidate image proposals, and the semantic graph could then be encoded for textual description generation. In this way, the semantic constraints summarized in MMKGs can be fully used, which may further endow the MMKGs ability and readily extended for more advanced reasoning.
2) The symbolic knowledge from MMKGs may enable the understanding of unseen objects~\cite{mogadala2017describing}, which are made visible by the semantic relation between seen objects and unseen objects in MMKGs.
In the knowledge-guided image-caption task containing novel objects, the key module is a multi-label image classifier for grounding depicted visual objects to knowledge base entities, unveiling a way to build a connection between real-world objects to their multi-modal information with the assistance of MMKGs~\cite{mogadala2017describing}. By introducing external knowledge from an MMKG-based multi-label classifier, image representations are also expanded.

A more complex task, named entity-aware image captioning, asks for more informative descriptions of named entities based on the background knowledge in the given article. 
In this task, these methods that only focus on textual knowledge and neglect the associations between named entities and visual cues in the image perform badly.
However, MMKGs are very handy for the task requiring fine-grained cross-modal alignment between named entities and their images and further extension. 
In \cite{zhao2021boosting} the textual scene graph and visual scene graph extracted from the input article and images are aligned by the cross-modal entity matching module pre-trained on Wikipedia articles and images. Incorporating the aligned cross-modal scene graphs and external knowledge from Wikipedia, more accurate named entities and relevant events are chosen and refined.
The results show that the structurization of cross-modal data improves the value of BLEU, METEOR, ROUGE, CIDEr and entity F1, where structurization with external knowledge significantly improves the performance.
%
%\zxr{\sout{Though some studies extract and encode textual knowledge to construct a more fine-grained attention mechanism, they neglect the associations between named entities and visual cues in the image and therefore  perform badly in some complicate scenarios. 
%
%However, MMKGs can capture the fine-grained relationships between entities in the context and objects in the image, for generating captions with more accurate named entities and more relevant events~\cite{zhao2021boosting}.
%
%More specifically, two different MMKGs are leveraged for various functions of different modules. First, in the cross-modal entity linking module, a complete MMKG is constructed by connecting a text sub-graph and an image sub-graph extracted from the input article and image respectively, while in the meantime incorporating an external MMKG as an assistance. The well-established MMKG then together with the image and article greatly benefits the entity-aware caption generation procedure afterwards. }}

\noindent {\bf Visual Storytelling}. Visual storytelling is more challenging, aiming to tell the story according to several successive images.
%
%Visual storytelling is a more challenging task than image captioning, which aims to tell the story according to a number of successive images.
%
This task requires discovering the relations between the images and the objects associated with the images. 
%
%Given a number of successive images, the task should not only present objective descriptions of objects in each image, but also need to discover the relations between the images and the objects associated with the images. 
%
Traditional visual storytelling approaches usually treat the task as a sequential image captioning problem and ignore the relation between images, which may produce monotonous stories. Besides, these approaches are limited to the vocabulary and knowledge in a single training dataset.
To tackle these problems, \cite{hsu2020knowledge} resorts to an MMKG for help within a distill-enrich-generate three-stage framework.
After extracting a set of words from each image, all words from two consecutive images are paired to query the MMKG (such as Visual Genome) to enrich possible triples. Then story sentences are generated based on the most reasonable triple step by step.
The methods using the relations in KGs show a strong ability of logical inference between images, generating more fluent stories than non-KG methods, and the triples from Visual Genome perform better than those from OpenIE in this task.

\subsubsection{Multi-modal Recommender System}

Recommender systems aim to recommend items that users might like/buy through the analysis of historical data, where accuracy, novelty, dispersity, stability and other factors should be balanced~\cite{bobadilla2013recommender,xiaoxuanshen2021deep}. %,hailiu2022edmf,xiaoxuanshen2021deep,duantengchuan2021carm}. 
%
%Recommender system aims to recommend items that users might like/buy through the analysis of historical data.
%
\eat{
Various factors needs to be balanced during the process, such as accuracy, novelty, dispersity and stability~\cite{bobadilla2013recommender}.
}
%
%Collaborative Filtering (CF) methods play a significant role in the recommendation. Other filtering techniques are along with the basic methods and can be categorized into memory-based CF, model-based CF or Hybrid methods~\cite{su2009survey}.
%
Where there are multi-modal data such as image and text in a recommending scenario, we say it is a multi-modal recommender system, where the information of different modalities should be leveraged jointly.

It has been proved that MMKGs could greatly enhance multi-modal recommender system~\cite{su2009survey}.
%Recent years have proved that MMKGs could greatly enhance multi-modal recommender system~\cite{su2009survey}.
%
First, MMKGs incorporate different modal data with a hierarchical structure, enriching the representations of items~\cite{sun2020multi}, which can be used to solve the cold-start problem long existing in collaborative filtering based on recommending strategies~\cite{zhang2016collaborative}.
Second, MMKGs can be used to select better logical reasoning paths for more explicit and explainable recommendations. For instance, \cite{tao2021multi} takes advantage of the the graph structure of MMKGs to design a hierarchy-based attention-path, which reduces the size of the action space and lets the model be more focused on critical intermediate items (entities).
The results imply that additional structured textual and visual knowledge can significantly improve the recommendation quality~\cite{zhang2016collaborative,sun2020multi,tao2021multi}.

%\zxr{\sout{Some approaches obtain the representations with rich semantics for items by leveraging external MMKGs. Incorporating information of MMKG across different modalities can help solve the cold-start problem long existing in Collaborative Filtering (CF) based recommending strategies~\cite{zhang2016collaborative}. }}%
%
%\zxr{\sout{Some other approaches find other ways to utilize MMKG for more personalized and explainable recommendations. For instance, \cite{tao2021multi} fully exploits the graph structure of MMKG and designs a novel approach hierarchy attention-path over MMKGs for the reasoning over items with information across different modalities. }}
%
%\zxr{\sout{Rich path semantics could be learned through entities and images within the path in MMKG, thus producing an interpretable and explicit recommendation with higher knowledge level.% but not just incorporating MMKG embeddings.
%
%Differently, some recent efforts~\cite{sun2020multi, tao2021multi} novelly proposes to construct a personalized MMKG from the images and texts of items in various ways, and then the entity relation reasoning between items can be better modeled by taking the relations in MMKG into account. % and therefore completing entity information.}}

\eat{
%One important approach is to construct a MMKG in a specific recommending scenario given that multimodal data of the items for explaining recommendation~\cite{sun2020multi}. The images associated with the item entity describe its appearances of the entity and enriches the entities representation with their hidden semantics. Moreover, entity relation reasoning between items can be better modeled taking into account the relations in MMKG and therefore completing entity information.

%Just as the different ways MMKG is constructed discussed above, there exists some minor distinctions in how these works construct MMKGs and how they represent them. Regarding as an entity grounding problem, CKE ~\cite{zhang2016collaborative, tao2021multi} constructs MMKG in a feature-based manner, which requires images for every entity. While MKGAT ~\cite{sun2020multi} handles differently by treating different types of information as first-class citizens of the MMKG instead of auxiliary information, which reduces data sources requirements for building the MMKG. 

%Not limited to constructing a personalized MMKG using item data, it is also helpful to obtain the representations with rich semantics by leveraging external MMKG, so as to further optimize the system’s recommendation strategy. Incorporating information of MMKG across different modalities can help solve the cold-start problem long existing in CF-based recommending strategies ~\cite{zhang2016collaborative}. 

Some other approaches find other ways to utilize MMKG for more personalized and explainable recommendations. ~\cite{tao2021multi} fully exploits the graph structure of MMKG and designs a novel approach hierarchy attention-path over MMKGs for the reasoning over items with information across different modalities. Rich path semantics could be learned through entities and images within the path in MMKG, producing an interpretable and more explicit recommendation with higher knowledge level but not just incorporating MMKG embeddings.
}

\eat{Among the recommendation strategies, collaborative filtering (CF) based approaches make good use of historical interactions of preferences and have shown great superiority. However, under scenarios which item set is quite large such as online shopping, the performance is often limited to the cold start problem for sparse user-item interactions. Incorporating multimodal information,~\cite{zhang2016collaborative} automatically extracts items’ structural, textual and visual content from an MMKG and then jointly learns different representations in a unified model integrating with CF, named CKE, to capture the implicit relationship between users and items. The item's latent vector generated from representations from different modalities and item's historical interactions are learned with user's latent vector to get the final recommendation. With experiments on two different domains (book and movie), CKE has shed the light on the insight that the usage of heterogeneous information from MMKG has proved to be effective for recommendation.

Another KG-based recommendation framework, MKGAT~\cite{sun2020multi}, has also constructed an MMKG and further developed an MMKG representation model to exploit them more efficiently. Compared to CKE~\cite{zhang2016collaborative}, the innovation of MKGAT lies in its exploitation of graph attention neural network. By information propagation on the MMKG, the entity embedding which fuses with information in different modalities can be better learned for recommendation. Besides, while CKE requires images for every entity in a feature-based manner, MKGAT treats different types of information as first-class citizens of the KG instead of auxiliary information, which reduces data sources requirements for building the MMKG. For each specific data type, different encoders are leveraged and dense layers are added to unify all modal of entities into the same dimension for the training procedure. The recommendation module then receives each entity’s corresponding embedding as the input and finally compute the user and item matching score after the concatenation of representations at each step. 

Compared to other KG-based recommendation method only using single modality such as KGAT~\cite{wang2019kgat}, MKGAT~\cite{sun2020multi} shows improving performance with the help of well-established entity-based MMKG. Some ablation studies to explore the effects of different modalities also lay more importance of visual modality than text. In this way, we can insight that MMKG plays a significant role for recommender systems, as visual information tends to attract a user's attention in real scenarios and provide more information while text only cannot contain.

Moreover, the MMKG embedding can be not only used to calculate the similarity between users and items for Top-N recommendation. With some policy-guided approaches, such as reinforcement learning (RL), the recommendation reasoning can be processed on the knowledge-aware path over the MMKG and therefore provides explanations for the recommendation results. 

Inspired by this idea and the wide application of MMKGs and deep RL in recent years,~\cite{tao2021multi} proposes a multimodal method MKRLN, incorporating both, for explainable recommendation systems. In the RL framework for recommendation, each entity in the MMKG is associated with an image and the initial state is composed of the concatenated embedding of entity and image. The MMKG here is not only used for the agent from the policy network to interact with, but also generates reward as feedback signal to choose the next state. An entity in the KG and its neighbors are presented as a continuous low-dimension vector space with additional context information extracted to complement its identification and enhance its semantic representation. The associated image of the entity from MMKG is taken as the input to process visual features, where an entity is related to its corresponding region. Besides, the introduction of hierarchy attention-path mechanism also helps to reduce the size of the action space due to the large number of neighbors of an entity. The path with the highest attention weight will be chosen, interpreting the reasoning process of recommendation.~\cite{tao2021multi} not only proves the effectiveness of integrating images and knowledge into recommendation, but also demonstrates that MMKG incorporating reinforce learning network further shows improving performance.}

\eat{
\subsubsection{Real-world Applications}
In addition to the above well-studied multi-modal downstream applications, MMKGs have already been applied to some real-world scenarios as introduced below.

\noindent {\bf COVID-19 Diagnosis}. In medical field, inferences of diseases can greatly benefit from multimodal information.~\cite{zheng2021pay} leverages this idea and present a multi-modal knowledge graph attention mechanism for auxiliary diagnosis of COVID-19. First construct the MMKG using 4 types of nodes including X-ray, CT, ultrasound and diagnose textual descriptions. Then offer embeddings for different layers and unify them afterwards. The proposed attention mechanism encoded with MMKG dramatically improves the classification performance of COVID-19 diagnosis, making full use of different modality medical images as well as doctor-patient dialogue while significantly reducing the risk of close contact.

\noindent {\bf Life Event Detection}. Another interesting application of MMKG lies in the life event detection and personal knowledge base construction from social media posts, which helps people recall wonderful memories and provides living assistance sometimes. To overcome the shortage of brief and informal written of user generated text as well as detect more general life events rather than only major ones, \cite{yen2020multimodal} presents multi-modal joint learning model, fully take advantage of both visual and textual information shared on Twitter, to extract life events to construct a personal life MMKG, supporting memory recall of individuals.

\noindent {\bf Academic Assistant}. Multi-modal information in literature of a particular domain can also be represented as a unified knowledge graph, making it more convenient for people to quickly learn about the latest progresses on concerned topics and find the most appropriate methods to solve their own problems.

\cite{kannan2020multimodal} considers text, diagram and source code of as three modalities and extracts them separately, the complementary information represented from KGs of different modalities are then aligned together. With the structured RDF format using DCC schema ontology, researchers can easily query the MMKG to look for new trends in Deep Learning areas and dive deep into model specific implementations.

}

\subsection{Domain Applications}\label{sec:domain}
In addition to applications on movie recommender~\cite{sun2020multi} or e-commerce KBQA systems~\cite{xu2021alime}, MMKGs are also applied in multi-modal tasks such as cross-modal retrieval, dialogue system and object detection in some domain applications. For instance, \cite{deng2021gakg} uses a geoscience academic MMKG to help to retrieve multi-hop queries, such as papers about specific geographic locations with a certain affiliation. \cite{kannan2020multimodal} uses an academic MMKG about papers and codes to offer retrieval on the implementation level.
In some other works, MMKGs are adopted to enrich the representation of entities with the help of images (e.g., X-rays, CT and ultrasound) and textual description, improving the performance of doctor-patient dialogue systems of COVID-19~\cite{zheng2021pay} and further reducing the risk of close contact. 
In the archaeology field, MMKGs also contribute to oracle bones detection and recognition, not only taking into account edges, textures, cracks, scratches, splinters and background, but also offering relevant literature, location and institutions to assist decision making\cite{xiong2021oracle}.

\section{Open Problems}
\label{sec:opportunities}

%This section discusses some open problems on MMKG construction and application leaving for future research.

\subsection{Complex Symbolic Knowledge Grounding}
Besides entities, concepts and relations, some applications require the grounding of complex symbolic knowledge consisting of multiple relational facts with close semantic relations.
%Besides the grounding of entities, concepts and relations, some downstream applications also require the grounding of complex symbolic knowledge which consists of multiple relational facts that have close semantic relations with each other.
%
These multiple relational facts may be a path or a subgraph in a KG. For example, for a subgraph in a KG containing \texttt{Trump}’s wife, daughter, grandson etc., a proper grounding image might be a \emph{Trump's family} photo.
This motivates \emph{multiple relational grounding}, which aims to find images to express the knowledge in a path or a subgraph in a KG.
%
%These complex symbolic knowledge grounding tasks pose new research challenges, but we can hardly find relevant literatures yet.
%
Multiple relational grounding is challenging since it involves the grounding of more than one relation, which is usually interleaved with each other in a complicated way.
%
%Multiple relational grounding is challenging since it involves the grounding of more than one relations and these multiple groundings are usually interleaved with each other in a complicated way. 
%We have to find the images that fully embody the composite semantic relations.  The composite semantic in many cases is only implicit expressed and might change over time. %All these factors together pose great challenge for multiple relational grounding. 
%
%
%As shown in Table \ref{tab:otherComponents}, 
%
%For example, for a subgprah in KG containing \texttt{Trump}’s wife, daughter, grandson etc, a proper grounding image might be a \texttt{Trump}'s family photo. %Obviously, it is nontrivial for machines to retrieve such a photo. %automatically select the best image from all image relevant to \texttt{Trump}.

\eat{
% 修改后
\begin{table}
	\small
	\centering
	\begin{tabular}{lccc}  
		\toprule
		symbol & subgraph & path\\
		\midrule
		example & Trump's family & Trump $\rightarrow$ Obama \\
		image    & \includegraphics[width=0.15\textwidth]{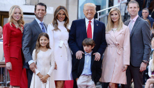} & \includegraphics[width=0.15\textwidth]{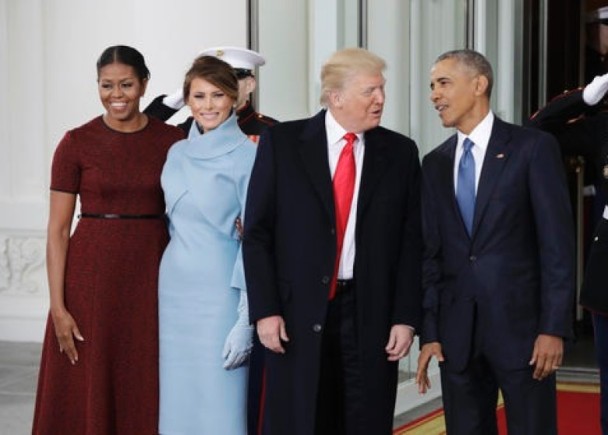} \\
		\bottomrule
	\end{tabular}
	\caption{Examples of multiple relation grounding.}% Trump's family members constitute a subgraph in the knowledge graph, so a set of family photos can present this subgraph well. Trump and Obama may not directly related in the KG, but there will be a path between them like (Trump - President - USA - ex-president - Obama), so the photo of them standing on the East steps of the U.S. Capitol during the 58th Presidential Inauguration is appropriate.
	\label{tab:otherComponents}
\end{table}
}

%\noindent {\bf 1) Quality Control}
\subsection{Quality Control}

%In general, we rely on data-driven approaches to build a large scale MMKG. 
%
%An MMKG that automatically harvested from big data inevitably suffer from quality issues, i.e., the MMKG might contain errors, missing facts or outdated facts. 
%
%For example, it is easy to find a wrong image for a long-tail entity from search engines, because such entity might have no image on the Web, thus any clicked image are misleading to a mistakenly grounding.

%Text information of an entity will be used such as entity name, description and category when searching for its relevant images on the Internet. It's the directest way to get candidates that we discussed in the last section. However, text information will introduce some inaccuracy into our system which affects the quality of our Multi-modal KG as shown in Table \ref{tab:Inaccuracy}.

\begin{table}[t]
	\centering
	\scriptsize
	\begin{tabular}{lp{1.8cm}p{1.8cm}p{1.8cm}}  
		\toprule
		entity & Pluvianus aegyptius & The Wandering Earth & arrogance\\
		\midrule
		image    & \includegraphics[width=0.12\textwidth]{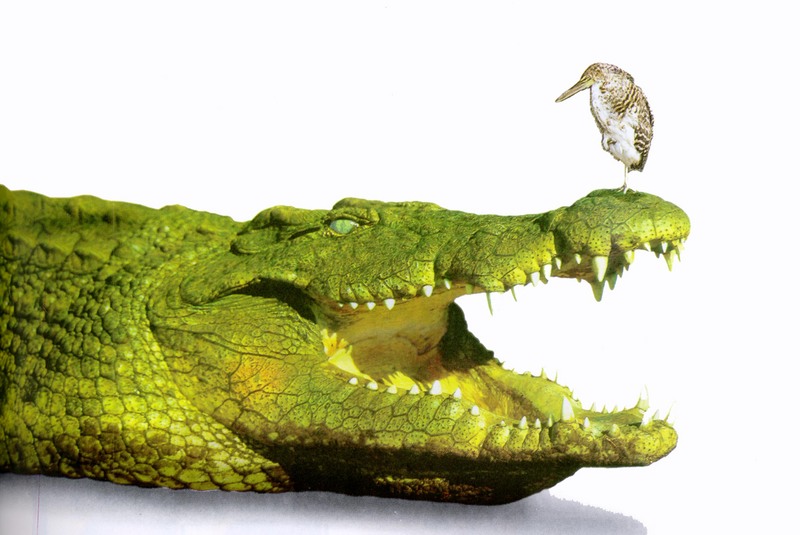} & \includegraphics[width=0.07\textwidth]{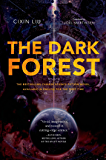} & \includegraphics[width=0.11\textwidth]{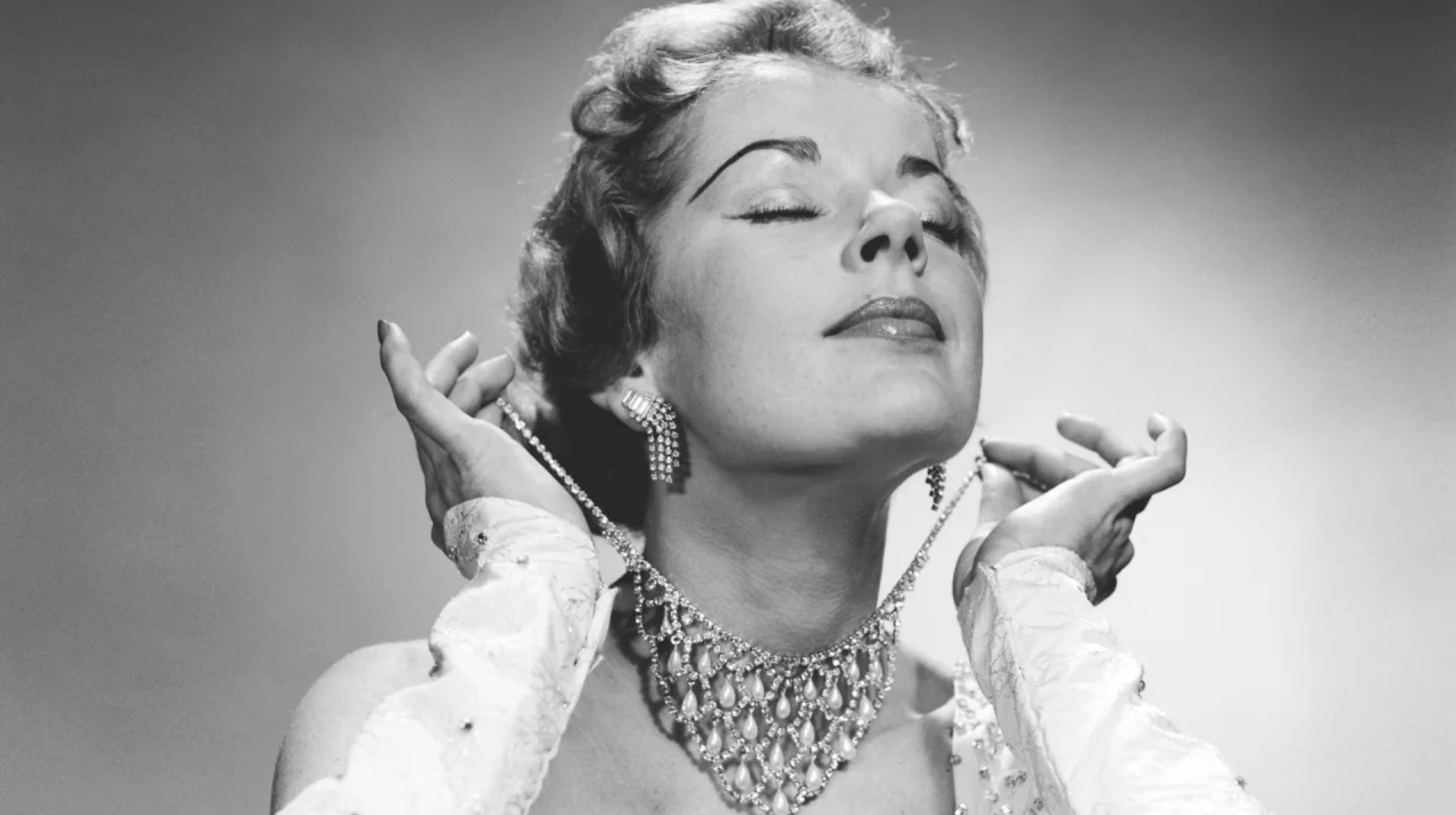}\\
		\bottomrule
	\end{tabular}
	\caption{Examples of quality problems in MMKG, such as images of frequently co-occurring entities, long-tailed entities, and abstract entities.}
	\label{tab:Inaccuracy}
\end{table}

Besides the common quality problems studied extensively in traditional KGs (e.g., accuracy, completeness, consistency and freshness), MMKGs have some special quality issues that concern the images (e.g., wrong, missing or outdated facts), as shown in Table~\ref{tab:Inaccuracy}.
%Besides the common quality problems in accuracy, completeness, consistency and freshness, which are discussed and studied extensively in traditional KG, MMKG has some special quality issues that concerns about the images.
%
Firstly, the image of some entity might be easily mixed with another when the two entities are closely related.
%
%See the first example in Table~\ref{tab:Inaccuracy}, 
\texttt{Pluvianus aegyptius} is a kind of bird that has a symbiosis with crocodiles, so we always get a picture of both the crocodile and the bird when searching for it. 
Secondly, the images of a more famous entity may easily appear in the entity grounding results of its closely-related entities. 
%
%See the  second example in Table~\ref{tab:Inaccuracy}, 
\texttt{The Wandering Earth} is written by the famous Chinese science fiction writer \emph{Liu Cixin}. While searching for this book, we always get a picture of his another more famous book, named \texttt{The Dark Forest}. 
Thirdly, some abstract concepts' visual features are not clear enough. For example, visual features of the %abstract noun 
\texttt{arrogance} are unfixed, so we always get some completely irrelevant pictures.
%
%
%To tackle the above problems, more visual analysis and background information might be needed to guide click-through rates and text information to avoid this misdirection.

\subsection{Efficiency}

Efficiency is always a non-negligible issue when building a large-scale KG. The efficiency problem of constructing an MMKG is more striking, since the extra complexity of processing multimedia data needs to be considered.
For example, it takes NEIL~\cite{chen2013neil} around 350K CPU hours to collect 400K visual instances for 2273 objects, while in a typical KG we need to ground billions of instances. The scalability of the existing solutions in building MMKGs will be greatly challenged. If the grounding objective is video data, the scalability issue might be amplified.

Besides the construction of MMKG, the online application of MMKG also needs to carefully address the efficiency issue since the MMKG needs to serve applications in real-time. The solution's efficiency is crucial for online MMKG-based applications.
% 这段废话有点多。最后一句和两句，一个意思重复说了三遍。

% Knowledge Graphs are scaling out and audio and video data also need to be added. How long will it take should really be considered. 

%Here we divide this task into two phases, the first one of which is looking for candidates and the second one is screening out the appropriate images from the candidates. The time that the second phase takes is determined by the number of candidates, the model we choose and the machine's performance. Among these three factors, the number of candidates affects the first phase as well. Thus, it really matters to work out how to find the candidates and how many candidates should be collected.

%First, as for images, candidates can be collected from search engine, but there is no direct search interface for audio and vedio data on the whole network, which makes it more time-consuming to collect audio and vedio candidates.

%Second, the resources of different objects on the network are not balanced and long-tail. It will takes much longer to collect the same number of candidates for long-tail entities. How to deal with these long-tail entities is also a problem.

\section{Conclusion}
\label{sec:conclusion}

We are the first to thoroughly survey the existing work on MMKGs constructed by texts and images. We systematically review the existing work on MMKG construction and application. We compare mainstream MMKGs in terms of what they contain and how they construct. We analyze different solutions' strengths and weaknesses in MMKG construction and applications. We not only point out some potential opportunities with the existing tasks in both MMKG construction and application, but also list some promising future directions with the construction and application of MMKGs.

\vspace{10pt}

\noindent
{\bf Acknowledgements:} This work is supported by National Key Research and Development Project (No.2020AAA0109302),
National Natural Science Foundation of China (No.62072323; No.62102095), Shanghai Science and Technology Innovation Action Plan (No.19511120400), Shanghai Municipal Science and Technology Major Project (No.2021SHZDZX0103).

\ifCLASSOPTIONcaptionsoff
  \newpage
\fi

\bibliographystyle{IEEEtran}
\bibliography{IEEEabrv_new}
% 参考文献有一些冗余信息，是否可以以统一的方式精简一下？参考simBiber？说不定可以压缩一些空间

\vspace{-40pt}

\begin{IEEEbiography}
[\psfig{figure=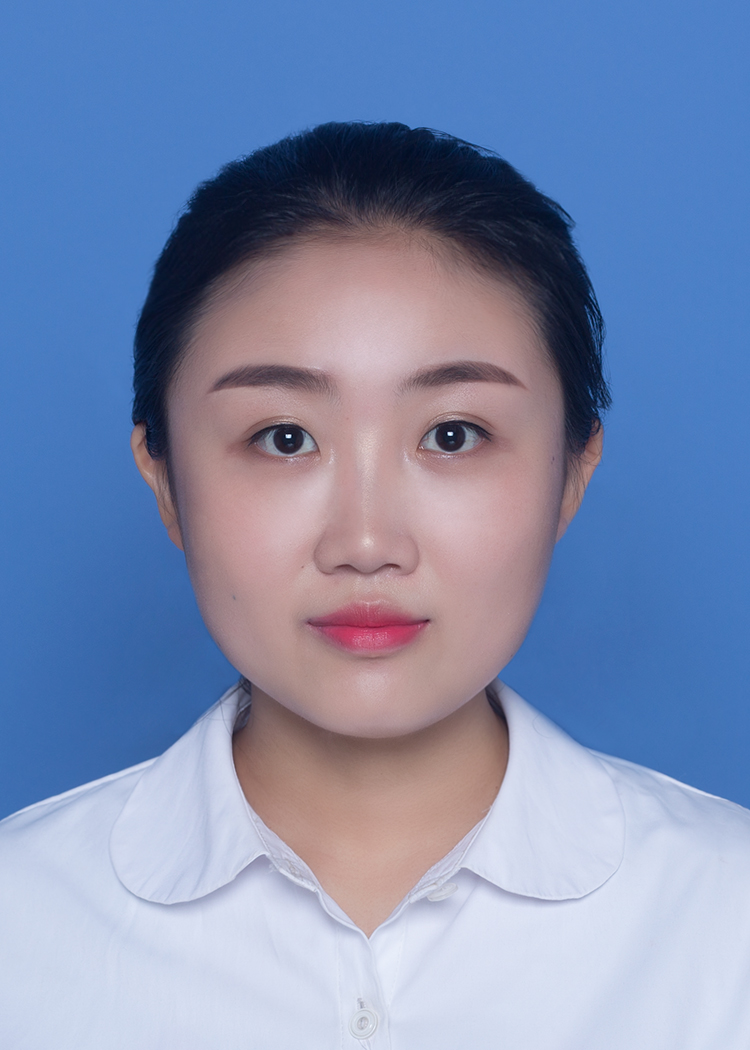,width=\textwidth}]
{Xiangru Zhu} is a Ph.D. student with the School of Computer Science at Fudan University, China. Her research interests include multi-modal knowledge graph and vision-language pre-trained model. 
\end{IEEEbiography}
\vspace{-40pt}

\begin{IEEEbiography}
[\psfig{figure=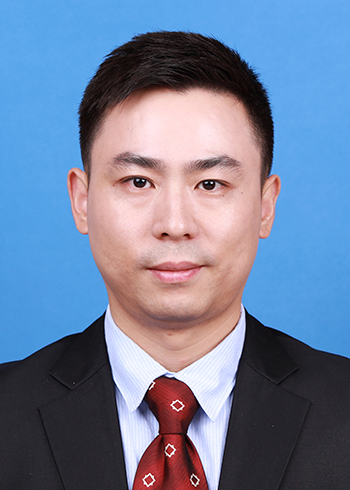,width=\textwidth}]
{Zhixu Li} is a professor with the School of Computer Science at Fudan University, China. %He is the director of Data Engineering and Multimodal Intelligence (DEMI) Group, and the deputy director of Knowledge Works Lab, Fudan University. 
He used to be a professor at Soochow University between 2014 and 2021. He received his Ph.D. degree in Computer Science from the University of Queensland in 2013.%, and his B.S. and M.S. degree from Renmin University of China in 2006 and 2009 respectively. 
His main research interests are Data \& Knowledge Engineering, and Cognitive Intelligence, and he is particularly interested in Multi-modal Knowledge Graph and Cross-Modal Cognitive Intelligence.
\end{IEEEbiography}
\vspace{-40pt}

\begin{IEEEbiography}
[\psfig{figure=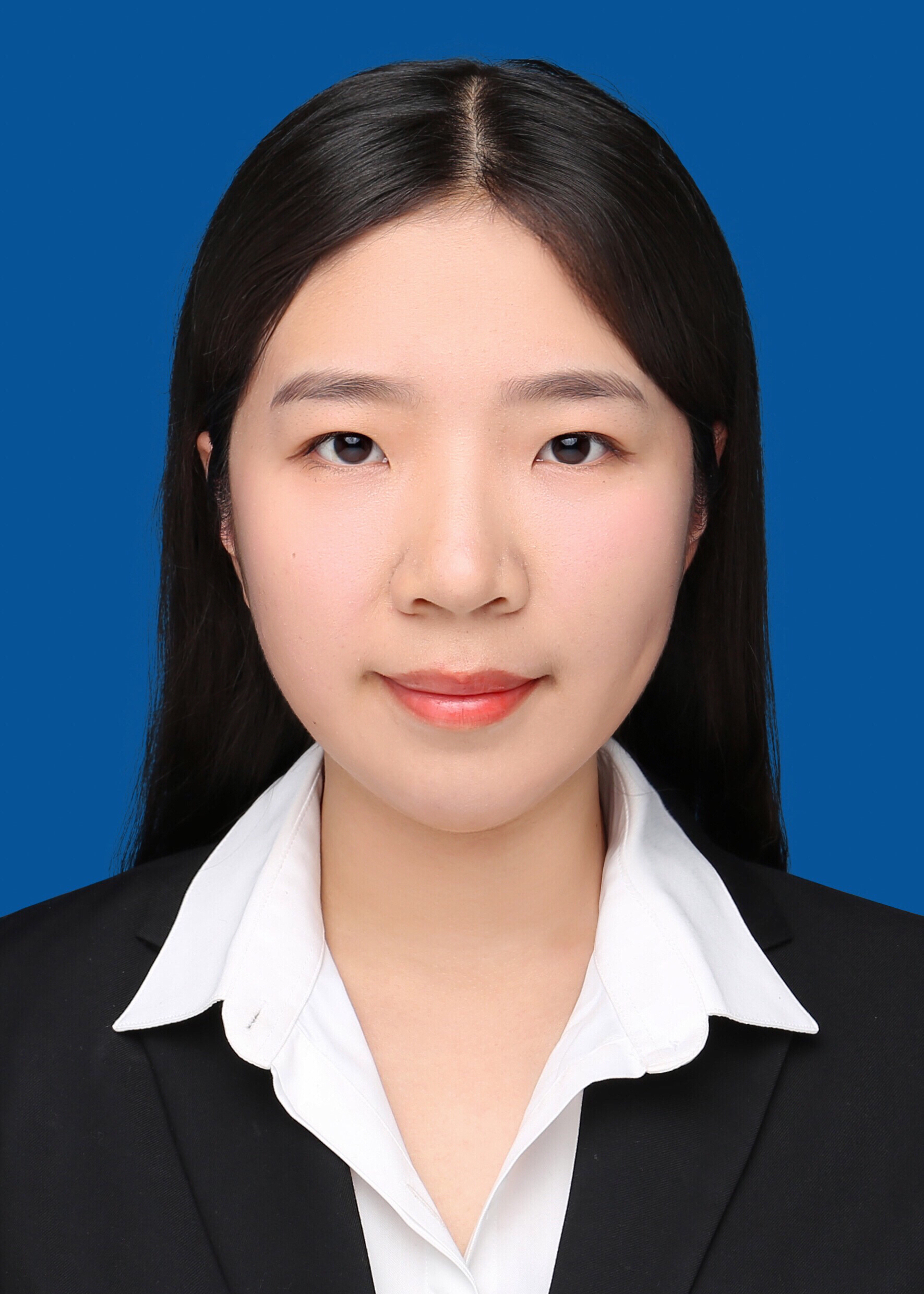,width=\textwidth}]
{Xiaodan Wang} is a Master student with the School of Computer Science at Fudan University, China. Her research interests include image-text retrieval and multi-modal knowledge graphs construction.
%Xiaodan Wang received the BS degree in software engineering from Southeast University, in 2020. She is currently working toward the master’s degree in software engineering in the School of and Computer Science, Fudan University. Her research interests include image-text retrieval and multi-modal knowledge graphs construction.

\end{IEEEbiography}
\vspace{-40pt}

\begin{IEEEbiography}
[\psfig{figure=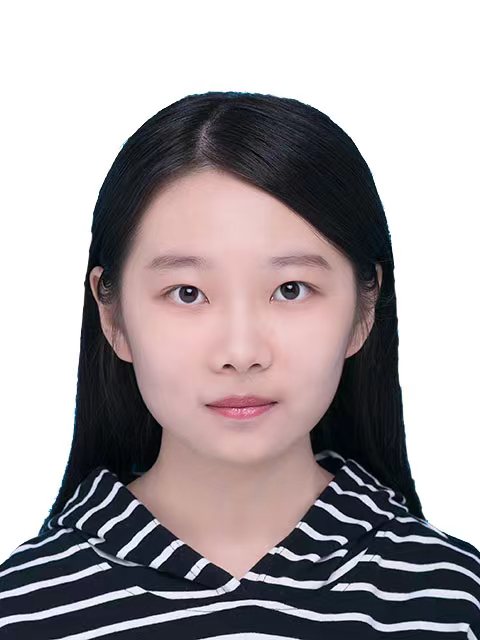,width=\textwidth}]
{Xueyao Jiang} received her Master degree in computer science from Fudan University in 2022. Her research interests include multi-modal knowledge graphs construction and application.
\end{IEEEbiography}
\vspace{-40pt}

\begin{IEEEbiography}
[\psfig{figure=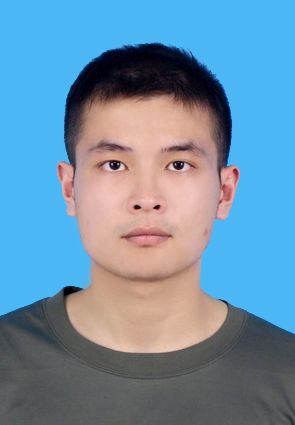,width=\textwidth}]
{Penglei Sun} is a master student with the School of Computer Science at Fudan University, China. His research interests include scenario-driven multi-modal knowledge graph construction and application. 
\end{IEEEbiography}
\vspace{-40pt}

\begin{IEEEbiography}
[\psfig{figure=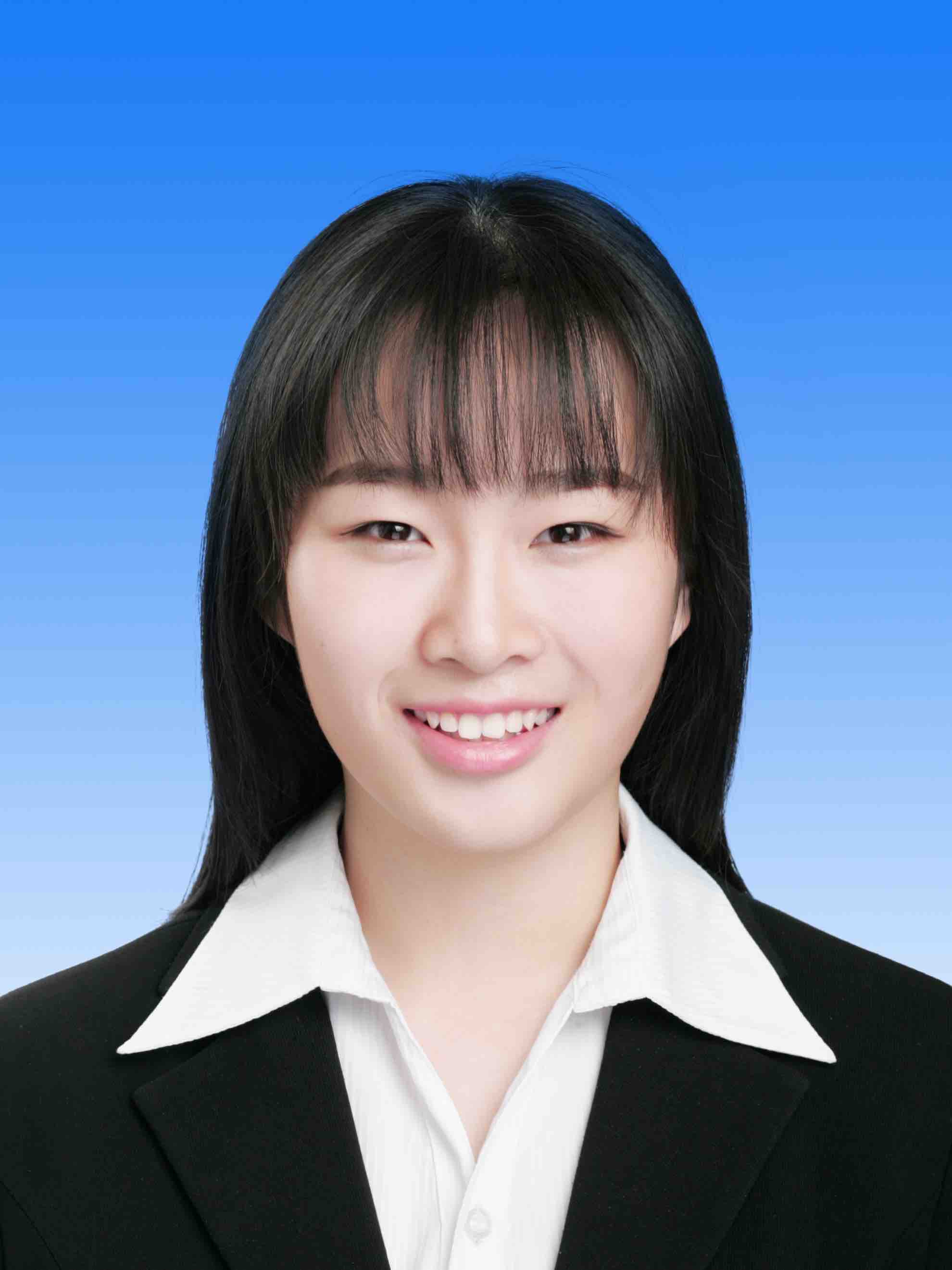,width=\textwidth}]
{Xuwu Wang} is a Ph.D. student with the School of Computer Science at Fudan University, China. Her research interests mainly focus on entity recognition, entity linking and multi-modal knowledge acquisition and application. She has already published several papers on ACL, ICME, DASFAA etc.
\end{IEEEbiography}

\vspace{-40pt}

\begin{IEEEbiography}
[\psfig{figure=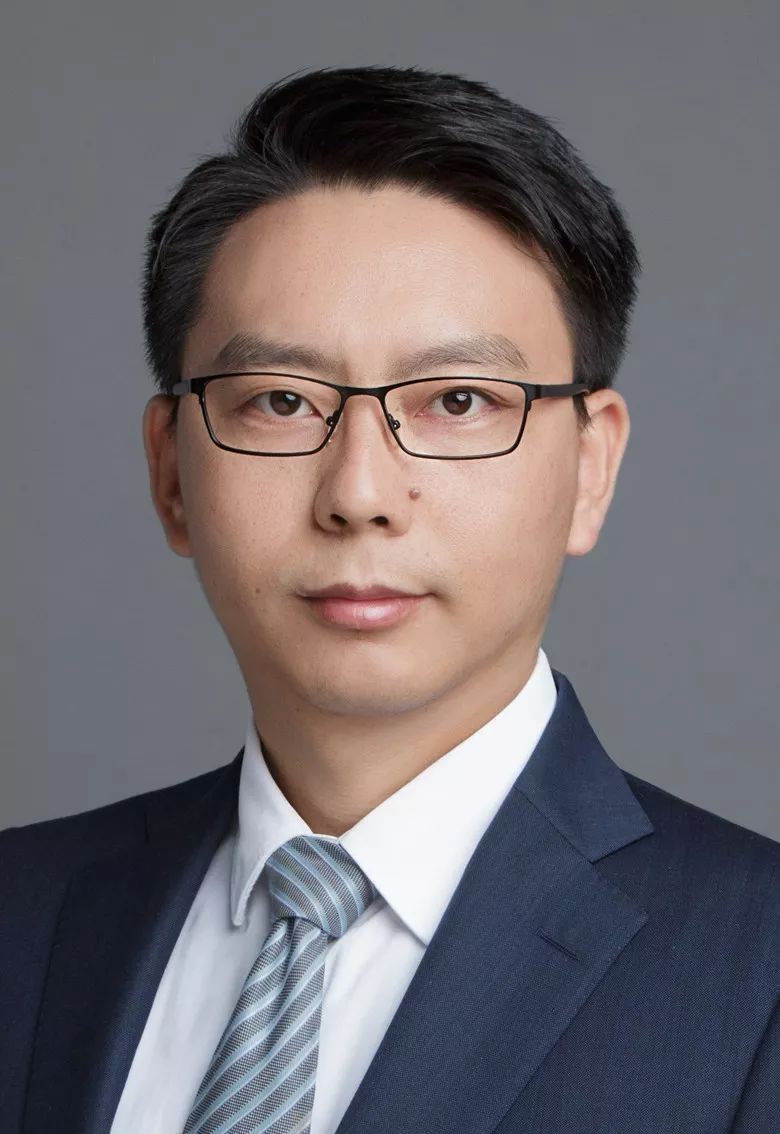,width=\textwidth}]
{Yanghua Xiao} is a professor of computer
science at Fudan University. He is the director of Knowledge Works Lab, Fudan University. He got his PHD degree in software theory from Fudan University, Shanghai, China, in 2009. He is one of young 973 scientists. His research interest includes big data management and mining, graph database, knowledge graph. He was a visiting professor of Human Genome Sequencing Center at Baylor College Medicine, and visiting researcher of Microsoft Research Asia.
\end{IEEEbiography}

\vspace{-40pt}

\begin{IEEEbiography}
[\psfig{figure=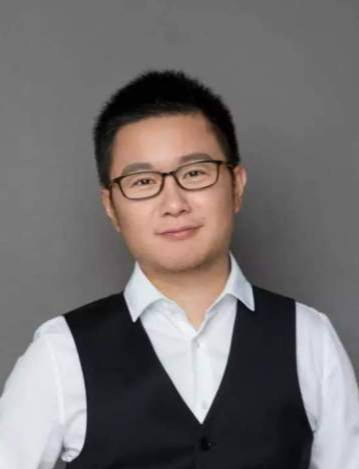,width=\textwidth}]
{Nicholas Jing Yuan} (Senior Member, IEEE) is currently a General Manager of AI Services, the Chief Scientist, and the Director of the Language and Speech Innovation Lab, Huawei Cloud. He has published more than 60 papers in top-tier conferences and journals, including several best paper awards such as SIGKDD (2016, 2018), ICDM (2013), and SIGSPATIAL (2010). His research work has been featured by influential media such as MIT Technology Review many times, and was reported directly to Bill Gates (Founder of Microsoft) by himself. He served regularly as program committee members in top tier conferences such as SIGKDD, WWW, ACL, and AAAI. 
He is a Senior Member of ACM and CCF.

\end{IEEEbiography}

\end{document}